\RequirePackage{fix-cm}
\documentclass[smallextended]{svjour3}       % onecolumn (second format)
\smartqed  % flush right qed marks, e.g. at end of proof
\usepackage{graphicx}
\usepackage{longtable}
\usepackage{cite}
\usepackage{amsmath,amssymb,amsfonts}
\usepackage{algorithmic}
\usepackage{textcomp}
\usepackage{xcolor}
\usepackage{mathptmx}
\usepackage[ruled,linesnumbered]{algorithm2e}
\usepackage{multirow}
\usepackage{bm}
\usepackage{booktabs}
\usepackage{enumitem}
\usepackage{tabularx}
\usepackage{supertabular}
\usepackage{natbib}
\usepackage[caption=false,font=normalsize,labelfont=sf,textfont=sf]{subfig}
\usepackage[labelfont=bf, labelsep=space, font=small]{caption}
\usepackage[toc,page]{appendix}
\usepackage[misc,geometry]{ifsym}

\usepackage[colorlinks = true,
            linkcolor = blue,
            urlcolor  = blue,
            citecolor = blue,
            anchorcolor = blue]{hyperref}
\DeclareGraphicsExtensions{.pdf,.png,.jpg}
\newcommand{\figref}[1]{Fig. \ref{#1}}
\newcommand{\tableref}[1]{Table \ref{#1}}
\usepackage{tikz}
\usetikzlibrary{tikzmark}
\usetikzlibrary{matrix}

\newcommand{\subf}[2]{%
  {\small\begin{tabular}[t]{@{}c@{}}
  \mbox{}\\[-\ht\strutbox]
  #1\\#2
  \end{tabular}}%
}

\begin{document}

\title{Fast, Accurate and Interpretable Time Series Classification Through Randomization}%\thanks{Grants or other notes
%about the article that should go on the front page should be
%placed here. General acknowledgments should be placed at the end of the article.}

% \subtitle{Do you have a subtitle?\\ If so, write it here}

%\titlerunning{Short form of title}        % if too long for running head

\author{Nestor Cabello       \and
        Elham Naghizade        \and
        Jianzhong Qi           \and
        Lars Kulik
}

% \authorrunning{Short form of author list} % if too long for running head

% \institute{F. Author \at
%               first address \\
%               Tel.: +123-45-678910\\
%               Fax: +123-45-678910\\
%               \email{fauthor@example.com}           %  \\
% %             \emph{Present address:} of F. Author  %  if needed
%           \and
%           S. Author \at
%               second address
% }

\institute{Nestor Cabello (\Letter) \at
        The University of Melbourne, Melbourne, Australia \\\email{ncabello@student.unimelb.edu.au} \\
          \and
          Elham Naghizade \at
          RMIT University, Melbourne, Australia \\\email{e.naghizade@rmit.edu.au} \\
            \and
            Jianzhong Qi \and Lars Kulik \at
            The University of Melbourne, Melbourne, Australia\\\email{\{jianzhong.qi, lkulik\}@unimelb.edu.au}
}

\date{Received: date / Accepted: date}
% The correct dates will be entered by the editor

\maketitle

\begin{abstract}
\emph{Time series classification} (TSC) aims to predict the class label of a given time series, which is critical to a rich set of application areas such as economics and  medicine. State-of-the-art TSC methods have mostly focused on classification accuracy and  efficiency, without considering the interpretability of their classifications, which is an important property required by modern applications such as \emph{appliance modeling} and legislation such as the \emph{European General Data Protection Regulation}. 
To address this gap, we propose a novel TSC method -- the \emph{Randomized-Supervised Time Series Forest} (r-STSF). r-STSF is highly efficient, achieves state-of-the-art classification accuracy and enables interpretability.  r-STSF takes an efficient interval-based approach to classify time series according to aggregate values of discriminatory sub-series (intervals). To achieve state-of-the-art accuracy, r-STSF builds an ensemble of randomized trees using the discriminatory sub-series. It uses four time series representations, nine aggregation functions and a supervised binary-inspired search combined with a feature ranking metric to identify highly discriminatory sub-series. The discriminatory sub-series enable interpretable classifications. Experiments on extensive datasets show that r-STSF achieves state-of-the-art accuracy while being orders of magnitude faster than most existing TSC methods. It is the only classifier from the state-of-the-art group that enables interpretability. Our findings also highlight that r-STSF is the best TSC method when classifying complex time series datasets.

% \PACS{PACS code1 \and PACS code2 \and more}
% \subclass{MSC code1 \and MSC code2 \and more}
\end{abstract}

\section{Introduction}
\label{sec:intro}

\begin{figure}[h]
\centering
\subfloat[]{
	\includegraphics[width=0.30\textwidth]{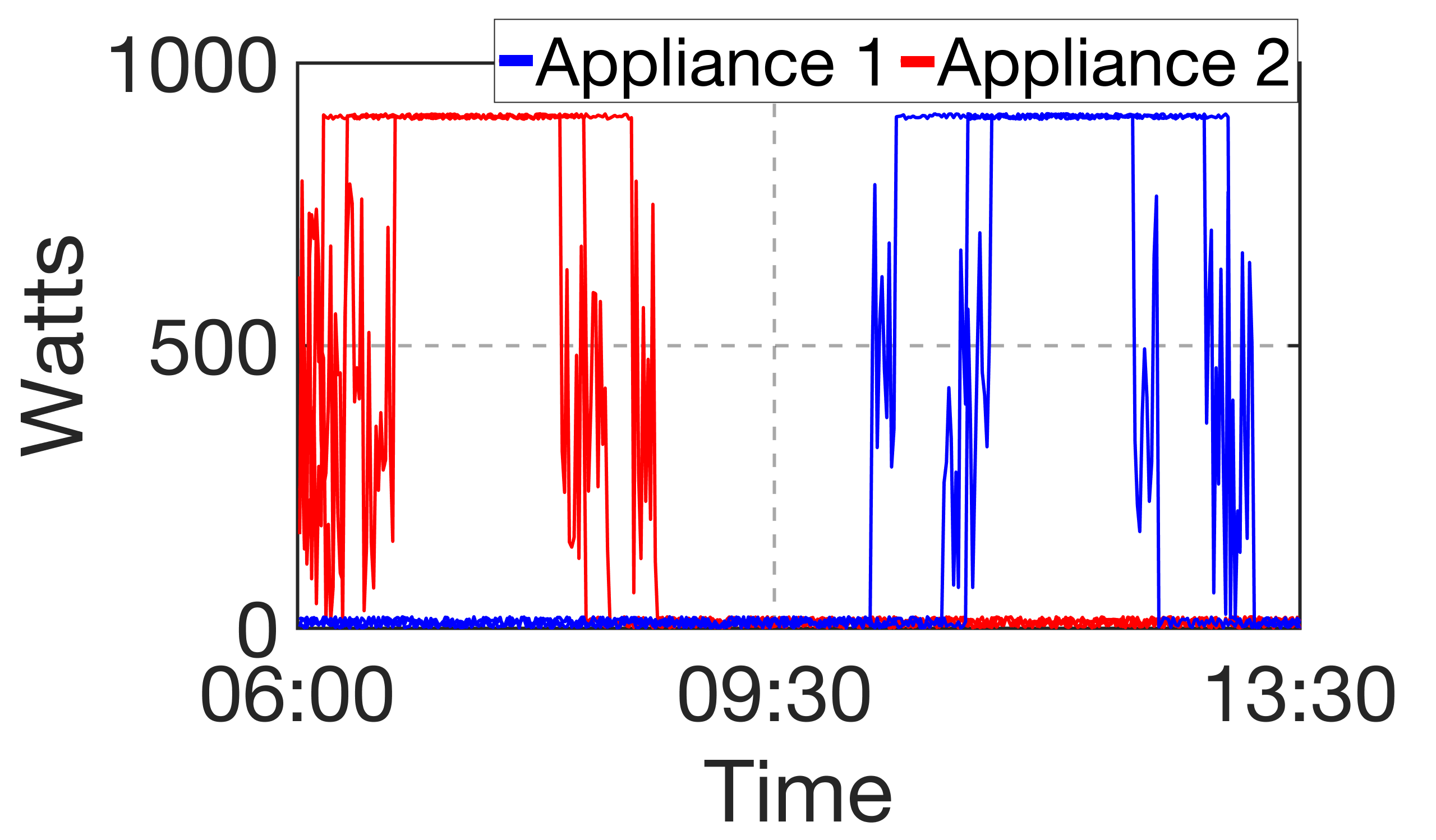}
	\hspace{5mm}
}
\subfloat[]{
	\includegraphics[width=0.30\textwidth]{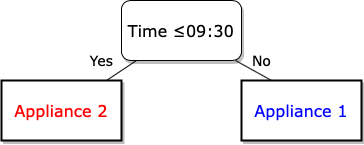}
}
\caption{\textcolor{black}{\textbf{(a)} Power consumption time series of two different types of appliances. \emph{Appliance 1} is represented with blue series and \emph{Appliance 2} with red series. \textbf{(b)} Tree-based (interpretable) classifier for this time series dataset. The type of appliance can be identified by its operation time interval. Series with high power consumption before 09:30 (i.e., between 06:30 and 09:30) are classified as \emph{Appliance 2}, whereas series after 09:30 as \emph{Appliance 1}}} \label{fig:AM-dummy}
\end{figure}

\emph{Time series classification} (TSC) aims to predict the class label of a given time series (or its \emph{feature-based representation}).
A time series is an ordered time-stamped sequence of observations from a variable of interest.
Various TSC methods have been proposed for a rich set of application areas such as economics (e.g., financial analysis~\citep{pattarin2004clustering}) and medicine (e.g., classification of electrocardiograms~\citep{karpagachelvi2012classification}).
Novel applications range from \emph{appliance modeling} (AM) to \emph{stress detection} (SD) and often collect rich data via sensors, resulting in large sets of time series with high update frequencies. They require fast and memory-efficient solutions. Besides, interpretability (i.e., ``the ability to explain or to present in understandable terms to a human"~\citep{gilpin2018explaining}) is also highly important. AM and SD require interpretable (readable) classifiers that use ``understandable" features rather than \emph{black-box} classifiers with complex features. For example, consider \figref{fig:AM-dummy}. A classifier for AM should report the time intervals that differentiate the time series from \emph{Appliance 1} and \emph{Appliance 2}, enabling householders to understand their energy consumption. Such explanations about automated decisions are required in 
legislation such as the \emph{European General Data Protection Regulation} (GDPR)~\citep{GDPR}. 
There is no consensus in the research community what interpretability means, but we consider an intuitive approach that regards a classifier as interpretable if it shows the factors that influence a prediction and if such factors have a physical meaning (e.g., height, cost, time).

State-of-the-art \textcolor{black}{(SOTA)} TSC methods such as the \emph{hierarchical vote collective of\\ transformation-based ensembles} (HIVE-COTE)~\citep{lines2018time} and \emph{time series combination of heterogeneous and integrated embedding forest} (TS-CHIEF)~\citep{shifaz2020ts} are highly accurate, but inefficient in terms of time and memory. HIVE-COTE has a bi-quadratic time complexity in the length of the time series, whereas in TS-CHIEF this complexity is quadratic. Another two TSC methods, InceptionTime~\citep{fawaz2020inceptiontime} and the \emph{RandOm Convolutional KErnel Transform} (ROCKET)~\citep{dempster2020rocket} have recently claimed SOTA classification accuracy. While InceptionTime is faster than HIVE-COTE which is largely due to the use of a GPU, it is still slower than ROCKET~\citep{middlehurst2020usage}. ROCKET is efficient empirically, but its time complexity is still quadratic in the number of extracted features or in the number of time series, whichever number is smaller.  
It may not scale well when classifying large datasets with very long series, as noted in its original proposal. All of these methods extract complex features without physical meaning from many different time series representations/transformations. They cannot provide interpretable classifications~\citep{le2019interpretable}. While recent approaches aim to provide fast and memory-efficient solutions with high accuracy, interpretable classifications remain under-addressed.

\begin{figure}[h]
    \centering
    \includegraphics[width=0.45\textwidth]{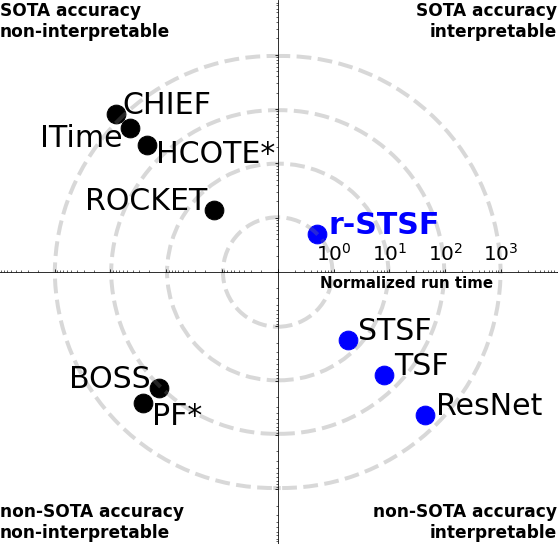}
    \caption{\textcolor{black}{Total run time (i.e., training plus testing time) of TSC methods when classifying 45 benchmark time series datasets normalized to r-STSF, i.e., r-STSF = 1. Each subsequent dashed circle represent an increase in the run time by a factor of 10. 
    Interpretable TSC methods are in blue color; non-interpretable ones in black color. \emph{r-STSF is the fastest approach and is the only TSC method that achieves both, SOTA accuracy and interpretability}. * For methods such as HCOTE and PF their running time is expected to be significantly larger as detailed in Section~\ref{subsec:running_time}}}
    \label{fig:runningTime_sotaAccu_interp}
\end{figure}

In this paper, we propose a novel TSC method: the \emph{Randomized-Supervised Time Series Forest} (r-STSF). \textcolor{black}{To the best of our knowledge, r-STSF is the only TSC method that achieves SOTA classification accuracy and allows for interpretability (cf.~\figref{fig:runningTime_sotaAccu_interp}). Moreover, it is extremely fast and memory-efficient. As shown in \figref{fig:runningTime_sotaAccu_interp}, r-STSF is the fastest TSC method. The SOTA TSC methods such as ROCKET, HIVE-COTE (HCOTE), InceptionTime (ITime) and TS-CHIEF (CHIEF) do not allow for interpretable classifications. Intrepretable TSC methods such as TSF, STSF, and ResNet do not achieve SOTA accuracy. Other competitive TSC methods such as BOSS and PF neither achieve SOTA accuracy nor allow for interpretability.}

\begin{figure}[h]
    \centering
    \includegraphics[width=0.50\textwidth]{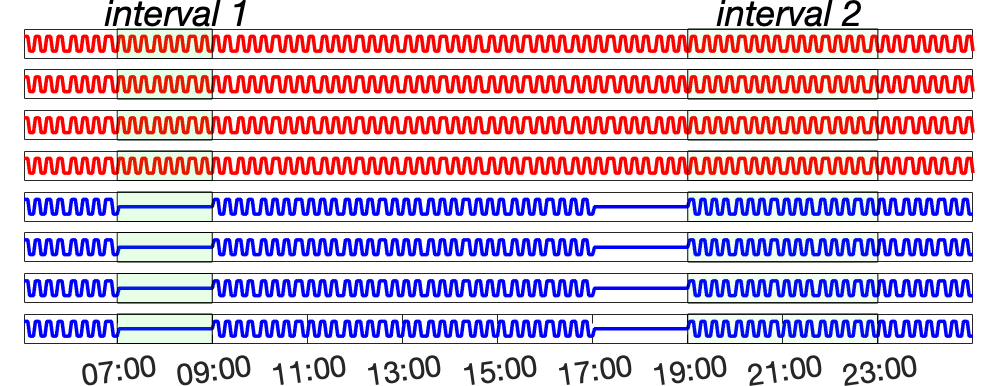}
    \caption{Discriminatory \textit{interval 1} versus non-discriminatory \textit{interval 2}}
    \label{fig:intervals}
\end{figure}

r-STSF is a tree-based ensemble classifier whose trees are grown using features derived from summary statistics over a number of randomly selected sub-series. Classifiers based on this paradigm are known as interval-based TSC methods; \emph{time series forest} (TSF)~\citep{deng2013time}, \emph{time series bag-of-features} (TSBF)~\citep{baydogan2013bag}, \emph{learned pattern similarity} (LPS)~\citep{baydogan2016time} and the \emph{supervised time series forest} (STSF)~\citep{cabello2020stsf} are typical examples. \textcolor{black}{Interval-based methods classify time series according to \emph{discriminatory phase-dependent intervals} (discriminatory intervals for short hereafter), i.e., sub-series located at the same time regions over all time series that maximize class separability}. As \figref{fig:intervals} shows, \textit{interval 1} is an example of a discriminatory interval. This interval differentiates the blue time series from the red ones. In contrast, \textit{interval 2} is non-discriminatory because it cannot separate the red and the blue time series.

Interval-based methods are fast and memory-efficient, while  their tree-based structure jointly with simple but meaningful features (the \emph{interval features}) enables interpretable classifications. The interval features are derived from simple transformations (i.e., summary statistics such as mean or standard deviation) over a number of sub-series and their meaning is easy to understand. The interactions between these features are modeled by binary trees, which are interpretable~\citep{breiman2001random}.

A well known interval-based method, TSF, has been shown to be one of the fastest TSC methods~\citep{bagnall2017great}. TSF grows an ensemble of trees for classification, and a different set of intervals features is computed at node level (on each of the trees from the ensemble). \textcolor{black}{TSF uses the \emph{temporal importance curve} to highlight the location of discriminatory intervals, enabling interpretability to the classification task. To compute the temporal importance curve, it is necessary to retrieve the entropy gain from each interval feature (i.e., each node of each tree from the ensemble). Thus, intervals that contribute the most to split the tree nodes are highlighted with a steep curve (i.e., high entropy gain) within their starting and ending time indices.}

Despite the advantages of TSF, most of its success (considering that its intervals are randomly selected) relies on a large number of intervals inspected when training a tree. When classifying large datasets with long series, TSF may become computationally expensive due to the larger number of intervals to be explored. 
Our previous work~\citep{cabello2020stsf} proposed a highly accurate interval-based classifier, STSF, which showed that a ``supervised" selection of intervals is significantly more efficient compared to one that is completely at random. STSF also trains an ensemble of trees, but it extracts a set of candidate discriminatory interval features for the training of each tree classifier instead of each tree node as done in TSF. 
Similar to TSF, STSF uses the interval features that split the tree nodes (i.e., discriminatory interval features) to provide interpretability to the classification results. STSF proposes the regions of interests (ROIs) to highlight the location of discriminatory intervals, which are the intersected regions of such intervals.

Other families of TSC methods have also considered interpretability. To the best of our knowledge, a shapelet-based classifier \emph{shapelet transform} (ST)~\citep{hills2014classification}, the deep-learning based methods \emph{residual networks} (ResNet) and \emph{fully convolutional networks} (FCN)~\citep{wang2017time}, and a SAX-based classifier mtSAX-SEQL+LR \citep{le2019interpretable} are the only classifiers (besides interval-based methods) able to provide interpretable classifications. However, we show in our experiments that they rank below SOTA TSC methods in terms of classification accuracy.  Intuitively, SOTA TSC methods rely on features in different transformations and complex representations, which are unavailable in the interpretable TSC methods mentioned above (including TSF and STSF).

To achieve high classification accuracy without relying on complex features, we propose a TSC method called \emph{r-STSF} that uses a novel perturbation scheme to grow an ensemble of uncorrelated trees. Uncorrelated trees reduce the variance of the ensemble and increase the classification accuracy~\citep{louppe2013understanding}. The perturbation scheme consists of (i)~random partitions of intervals when searching for discriminatory features and (ii)~an ensemble of randomized trees. To allow for each tree in the ensemble to use a different set of interval features for classification (i.e., ensemble of uncorrelated trees), r-STSF performs random partitions into the sub-series when searching for discriminatory interval features. In comparison, our previous STSF method uses fixed partitions (following a binary search-based scheme) to extract interval features. Fixed partitions are not suitable to create uncorrelated trees, because they may allow for the repetition of some interval features across different trees in the ensemble. We show that random instead of fixed partitions makes r-STSF significantly more accurate. To further decrease the variance of the ensemble, r-STSF adopts the extra-tree (ET) algorithm~\citep{geurts2006extremely}, which uses randomized trees. In such randomized trees, the cut-point of each feature is randomly selected when looking for the feature that provides the best split. These trees are different from those used in random forest (RF), where for each feature only the best cut-point (i.e., split with the lowest entropy) is selected. 
Although trees built by using RF are also considered randomized trees, they are only \textit{weakly randomized} because the cut point selection still relies on the lowest entropy. Methods based on ET, however, are \textit{strongly randomized} because the cut point selection is random. Thus, we will use the term \textit{randomized trees} exclusively for those trees that are built using the ET algorithm; all other type of trees will be considered non-randomized. We show that randomized trees instead of regular or non-randomized trees (as those based on RF) significantly increase the effectiveness of r-STSF.

Further, we introduce two new aggregation functions to form new features which are shown to improve the classification accuracy substantially: 
(i) the number of intersections of the sub-series with the mean axis of the entire time series, denoted as counts of mean-crossings in this paper, and (ii) the number of data points above the mean of a sub-series. These aggregation functions capture the shape of the time series and increase the classification accuracy more than generic statistics such as the mean or standard deviation.

To represent the temporal structure of time series data, we also include \emph{autoregressive representations} of the series in our feature extraction process. This representation models a time series with an autoregressive process, where a value of a time series is represented as a linear combination of (some of) its previous values. r-STSF computes autoregressive representations of the raw time series (via the autoregression coefficients), and it captures lagged relationships of time series data points by extracting discriminatory sub-series from such representations. Our experiments show that the extraction of features from the autoregressive representation increases the classification accuracy considerably (i.e., more than 5\%) for certain datasets.

r-STSF is very fast because it does not need to extract a set of candidate interval features for each tree classifier as STSF does. The training process of r-STSF is designed to require a small number (denoted by \emph{d}) of sets of candidate discriminatory interval features. 
r-STSF extracts $d$ sets of candidate interval features and merges all sets into a superset $\mathcal{F}$. Then, each of the $r$ trees in the ensemble is built by using a set of randomly selected interval features from $\mathcal{F}$. Our experiments show that by setting $d=0.1 \times r$, r-STSF achieves a similar classification accuracy (to computing a set of interval features for each tree) but being an order of magnitude faster.

An extensive experimental study on 85 time series datasets~\citep{UCRArchive} shows that r-STSF is statistically as accurate as SOTA classifiers. r-STSF is ranked as the sixth best classifier (according to the average ranks) among all fourteen TSC methods evaluated, being the most accurate interpretable TSC method and the only interpretable classifier that achieves classification accuracies competitive to SOTA methods. More importantly, r-STSF achieves the highest \emph{weighted average accuracy}~\citep{fernandez2014we} which considers the complexity of datasets (i.e., difficulty to achieve high classification accuracies). This shows that r-STSF is highly robust when classifying complex datasets. Since r-STSF is at least two orders of magnitude faster than most of the SOTA TSC methods and is the only SOTA classifier enabling interpretability, it is ideal for classifying large and complex datasets.

To summarize, in our previous work STSF~\citep{cabello2020stsf}, we show that computing interval features in a supervised manner from different time series representations by using robust statistics is a highly efficient and accurate interval-based classification strategy. In this article, we  extend STSF to r-STSF with substantial new contributions as follows:

\begin{itemize}[topsep=0pt]
    \item
    \textcolor{black}{We propose r-STSF, a novel TSC method that is highly accurate and efficient while offering interpretable classification results. r-STSF is  competitive to SOTA TSC methods in terms of average ranks and outperforms SOTA TSC methods in terms of weighted average accuracy (Section~\ref{subsec:sota-comparison}). It is also the fastest TSC method (Section~\ref{subsec:running_time}), while its interval-based design offers interpretable classification results (Section~\ref{sec:roi}).}  
  
    \item 
    We propose a novel perturbation scheme to create an ensemble of uncorrelated trees which improves the classification accuracy of interval-based techniques. Our scheme employs (i) random partitions when assessing the discriminatory quality of sub-series (Section~\ref{subsec:intervalfeaturesextraction}) and (ii) randomized trees to build the ensemble of trees for classification (Section~\ref{subsec:randomizedtrees}).
    Our selection of sets of sub-series can decrease the correlation of trees in the ensemble without affecting the strength of each tree.
    
    \item We redesign r-STSF's training process (Section~\ref{subsec:rstsfVSstsf}) and show that by (i) computing a small number of sets of candidate discriminatory interval features, (ii) merging all sets into a superset $\mathcal{F}$, and (iii)  randomly selecting interval features from  $\mathcal{F}$ to build each tree classifier, we achieve a similar classification accuracy (to computing a set of interval features for each tree) but being an order of magnitude faster.

    \item We show that time-specific aggregation functions (e.g., slope) have a larger positive impact in the classification accuracy than generic statistics (e.g., mean). To capture a higher level of details regarding the shape of the sub-series, we propose two time-specific aggregation functions: (i) the counts of mean-crossings, and (ii) the counts of data points above the mean (Section~\ref{subsec:intervalfeaturesextraction}). 
 
    \item We propose to capture lagged relationships of the time series data points by using autoregressive representations (i.e., autoregressive coefficients) of the time series  (Section~\ref{subsec:representation}). Our experiments show that extracting interval features from autoregressive representations of the series can increase the classification accuracy.

    \item We extend significantly the discussion of interpretability on TSC (Section~\ref{sec:roi}), presenting different case studies and showing the utility of our approach when interpreting classification results on different time series representations.
   
   \item To validate r-STSF's generalizability to more TSC data and to provide reference results for future  studies, we also run experiments over an additional 43 benchmark time series datasets and compare r-STSF with SOTA TSC methods such as ROCKET~\citep{dempster2020rocket} and InceptionTime~\citep{fawaz2020inceptiontime} (Section~\ref{subsec:43dsets}). To show the contribution of each component of r-STSF to the classification accuracy, we conduct an ablation study on r-STSF (Section~\ref{subsec:sensitivity}).

\end{itemize}

\section{Related Work}
Time series classification (TSC) methods can be broadly categorised into \emph{instance-based}, \emph{feature-based}, and \emph{ensemble-based}. Instance-based methods such as 1-Nearest Neighbour classifier with \emph{Dynamic Time Warping} (1-NN DTW)~\citep{ratanamahatana2005three} classify a time series according to the similarity of its ordered data points to those of time series with known class labels, where localized distortions of the data points are allowed. \textcolor{black}{Although it may be possible to use the warping path from the warping matrix of a testing instance to highlight the parts of the time series that are important to the classifier's decision, these methods have not considered interpretability explicitly.} 
Besides, these methods tend to be less accurate than SOTA TSC methods~\citep{bagnall2017great}, computationally inefficient for long series (i.e., quadratic to the length of the series), and noise sensitive if using raw series (i.e., without smoothing)~\citep{gorecki2013using}. In contrast, feature-based methods do not used the raw series directly. They represent the original time series with a set of derived properties, i.e., features, which leads to high-level representations. \emph{Shapelet-based}, \emph{dictionary-based}, and \emph{interval-based} methods are typical examples. 

\emph{Shapelet-based} methods~\citep{rakthanmanon2011fast, hills2014classification,grabocka2014learning} use shapelets, i.e, sub-series that are representative of class membership, and time series are classified according to their similarity to the (discriminatory) shapelets. \textcolor{black}{By conducting a post-transform clustering of the shapelets, these methods may provide some level of interpretability.} \emph{Shapelet Transform} (ST)~\citep{hills2014classification} is the SOTA shapelet-based method \citep{bagnall2017great}. \textcolor{black}{It achieves a competitive classification accuracy with a bi-quadratic time complexity over the length of the series, which makes it impractical for long series.}

\emph{Dictionary-based} methods~\citep{lin2012rotation,schafer2015boss} use the relative frequency of discriminatory sub-series for classification. \textcolor{black}{These methods transform each sub-series into a symbolic representation such as the \emph{Symbolic Fourier Approximation} (SFA)~\citep{schafer2012sfa} or the \emph{Symbolic Aggregate Approximation} (SAX)~\citep{lin2003symbolic} and employ a bag-of-patterns (BOP) model~\citep{lin2012rotation} to transform the symbols into a histogram representation}. The \emph{bag of SFA symbols} (BOSS)~\citep{schafer2015boss} is the best dictionary-based method according to~\cite{bagnall2017great}. \textcolor{black}{Although more recent dictionary-based classifiers such as the \emph{Word ExtrAction for time SEries cLassification} (WEASEL)~\citep{schafer2017fast}, \emph{Spatial BOSS} (S-BOSS)~\citep{large2019time}, and \emph{Contractable BOSS} (cBOSS)~\citep{middlehurst2019scalable} are more accurate than BOSS, their difference in classification accuracy is not significant~\citep{middlehurst2020temporal}, while WEASEL is even more memory-intensive than BOSS~\citep{le2019interpretable}. This makes BOSS a more practical TSC method.}
BOSS computes the number of times that a discriminatory sub-series (represented as a symbol) appears in the time series. This provides more accurate classifications if the  discriminatory sub-series appears in different classes, i.e., represents more than one class. In BOSS, the size of the alphabet grows exponentially to the word length, which makes this method memory intensive for long series. \textcolor{black}{Besides, it cannot provide interpretability to its results.} 

A recent TSC method, mtSAX-SEQL+LR~\citep{le2019interpretable} also transforms the time series into a symbolic representation, but it uses a \emph{Sequential Learner} (SEQL)~\citep{ifrim2011bounded} for feature selection and a \emph{logistic regression} (LR) classifier for classification. The LR classifier learns a linear classification model which is essentially a set of symbols and their coefficients. The symbols can be mapped back to their original location in the time series. Since the coefficients can be interpreted as the discriminatory power of the symbols, mtSAX-SEQL+LR uses such coefficients to enable interpretability into their classification results. mtSAX-SEQL+LR has quasi-quadratic time complexity over the length of the series. It is more efficient than BOSS and WEASEL (quadratic time over the length of the series). However, it is not competitive to SOTA TSC methods in terms of accuracy.

\emph{Interval-based} methods~\citep{deng2013time,baydogan2013bag,baydogan2016time} are well known to be highly efficient TSC methods~\citep{bagnall2017great}. These methods explore sets of sub-series to extract discriminatory intervals. The idea is that time series from the same class tend to have intervals with similar characteristics. \emph{Time series forest} (TSF) \citep{deng2013time} relies on random searches to reduce the high-dimensional interval space (i.e., quadratic to the length of the time series) and to explore intervals of different lengths. TSF uses three statistical measurements (mean, standard deviation, and slope) and tree-based ensembles to capture discriminatory intervals. TSF is fast and memory efficient~\citep{bagnall2017great}, and the discriminatory intervals \textcolor{black}{can be identified by using the \emph{temporal importance curve}~\citep{deng2013time}, which enables interpretable classifications.  The \emph{random interval spectral ensemble} (RISE)~\citep{lines2018time} was introduced  to capture frequency-domain features. RISE works similar to TSF, but it extracts spectral features over each random interval instead of statistical measurements as in TSF. Both TSF and RISE are highly efficient but less accurate than other feature-based TSC methods. More recent interval-based TSC methods such as our previous \emph{supervised time series forest} (STSF)~\citep{cabello2020stsf} and the \emph{canonical interval forest} (CIF)~\citep{middlehurst2020canonical} have improved  considerably the classification accuracy, and are competitive to highly accurate shapelet and dictionary-based classifiers. STSF selects its intervals in a supervised manner rather than completely at random as TSF does, and it uses a set of seven summary statistics to compute the interval features. Besides, STSF computes interval features not only from the original (i.e. raw) time series but also from additional time series representations such as the periodogram and first-order difference representations. CIF is similar to TSF but uses twenty-two descriptive statistics proposed by~\cite{lubba2019catch22} and not just the mean, standard deviation, and slope. Both STSF and CIF can provide interpretable classifications but are less accurate than SOTA TSC methods.} 

\emph{Ensemble-based} methods use ensembles of individual TSC methods. Three representatives are the \emph{elastic ensemble} (EE)~\citep{lines2015time}, the \emph{hierarchical vote collective of transformation-based ensembles} (HIVE-COTE)~\citep{lines2018time} and the recently proposed \emph{temporal dictionary ensemble} (TDE), which combines features from four dictionary-based TSC methods such as BOSS, WEASEL, S-BOSS, and cBOSS. TDE is significantly more accurate than all other dictionary-based TSC methods, but it is memory intensive. As shown in its original proposal, TDE is three times more expensive than BOSS in memory costs. For the ElectricDevices dataset~\citep{UCRArchive} which has 8926 training instances, but relatively short time series length (96 data points), TDE requires 10GB of memory~\citep{middlehurst2020temporal}. For large datasets with very long series such as RightWhaleCalls dataset~\citep{UCRArchive} which has 10934 training instances and series length of 4000, TDE would required approximately 300GB of memory (estimated according to BOSS results on RightWhaleCalls dataset presented in~\cite{middlehurst2019scalable}).
Hence, TDE is impractical to classify large time series datasets. EE and HIVE-COTE, are highly accurate but costly. They may become too expensive on long series. EE is based on 11 instance-based TSC methods, hence it has a quadratic time complexity over the length of the time series. HIVE-COTE uses EE, ST, BOSS, TSF, and RISE for classification and employs a modular hierarchical structure to allow a single probabilistic prediction from each classifier. 
HIVE-COTE is the SOTA TSC method in terms of accuracy but is impractical for long series (bi-quadratic time complexity over the length of the time series). \textcolor{black}{To make HIVE-COTE faster and more scalable, the recently proposed HIVE-COTE v1.0~\citep{middlehurst2020usage} makes modifications such as dropping the elastic ensemble (EE) and setting running time limits. In terms of accuracy, HIVE-COTE v1.0 is statistically similar to HIVE-COTE, while it is significantly faster in practice. In theory, HIVE-COTE v1.0 is still bounded to the time complexity of ST which is bi-quadratic to the length of the series. Two recent variations of HIVE-COTE, HC-TDE~\citep{middlehurst2020temporal} and HC-CIF~\citep{middlehurst2020canonical}, claim being significantly more accurate than HIVE-COTE. HC-TDE is built upon HIVE-COTE v1.0 and replaces BOSS with TDE. HC-CIF is built upon  HIVE-COTE and replaces TSF with CIF. Both HC-TDE and HC-CIF include ST, hence their time complexities are still bi-quadratic to the length of the series, which limit their applicability over long series.}

\vspace{-5mm}
\begin{table}[h]
\centering
\caption{\textcolor{black}{Summary of representative TSC methods: The time and space complexities show that these methods are not suitable for large datasets with long time series. None of the SOTA TSC methods allow for interpretable classifications. None of the interpretable TSC methods achieve SOTA classification accuracy}. Notation: $n$: number of time series; $m$: length of a series; $\kappa$: number of shapelets; $\alpha$: alphabet size; $\iota$: word length; $r$: number of trees; $c$: number of classes; $C_{e}$: number of candidate splits; $D_{t}$: number of dictionary-based transformations; $K$: number of kernels; $v$: number of extracted features.  For space, in TS-CHIEF's training time, we have only included the complexity of the similarity-based splitting. *Indicates that the information is not explicitly stated in the associated paper}

\label{table:related_work}
\setlength{\tabcolsep}{5pt}
\begin{tabular}{@{}lcccc@{}}
\toprule
\multirow{2}{*}{\textbf{\begin{tabular}[c]{@{}c@{}}Approach\end{tabular}}} & \multirow{2}{*}{\textbf{\begin{tabular}[c]{@{}c@{}}Training\\ time\end{tabular}}} & \multirow{2}{*}{\textbf{\begin{tabular}[c]{@{}c@{}}Training\\ memory cost\end{tabular}}} &
\multirow{2}{*}{\textbf{\begin{tabular}[c]{@{}c@{}}SOTA\\ accuracy\end{tabular}}} &
\multirow{2}{*}{\textbf{\begin{tabular}[c]{@{}c@{}}Interpretability\\ \end{tabular}}} \\
 &  &  &  &  \\ \midrule

\textbf{TSF} & $\mathcal{O}(r \cdot n \cdot \log\,n \cdot m)$ & $\mathcal{O}(rm)$ & No & \textcolor{blue}{Yes} \\
\textbf{ST} & $\mathcal{O}(n^{2}m^{4})$ & $\mathcal{O}(\kappa n)$ & No & \textcolor{blue}{Yes} \\
\textbf{FCN} & * & * & No & \textcolor{blue}{Yes} \\
\textbf{ResNet} & * & * & No & \textcolor{blue}{Yes}\\
\textbf{mtSAX-SEQL+LR} & $\mathcal{O}(n \cdot m^{3/2}\log\,m)$ & * & No & \textcolor{blue}{Yes}\\
\textbf{STSF} & $\mathcal{O}(r \cdot n \cdot \log\,n \cdot \log\,m)$ & $\mathcal{O}(rm)$ & No & \textcolor{blue}{Yes}\\
\textbf{CIF} & * & * & No & \textcolor{blue}{Yes}\\

\textbf{1NN-DTW} & $\mathcal{O}(n^{2}m^{2})$ & $\mathcal{O}(m^{2})$ & No & No \\
\textbf{BOSS} & $\mathcal{O}(n^{2}m^{2})$ & $\mathcal{O}(n\alpha^{\iota})$ & No & No  \\
\textbf{EE} & $\mathcal{O}(n^{2}m^{2})$ & $\mathcal{O}(m^{2})$ & No & No \\
\\
\multirow{2}{*}{\textbf{PF}} & \multirow{2}{*}{\textbf{\begin{tabular}[c]{@{}c@{}}$\mathcal{O}(r \cdot n \cdot \log\,n\cdot$ \\ $C_{e} \cdot c \cdot m^{2})$\end{tabular}}} & \multirow{2}{*}{*} & \multirow{2}{*}{\textcolor{black}{No}} & \multirow{2}{*}{No}
\\
\\
\\
\textbf{TDE} & * & * & No & No \\
\textbf{HIVE-COTE} & $\mathcal{O}(n^{2}m^{4})$ & $\mathcal{O}(\kappa nm^{2})$ & \textcolor{blue}{Yes} & No\\
\\
\multirow{2}{*}{\textbf{TS-CHIEF}} & \multirow{2}{*}{\textbf{\begin{tabular}[c]{@{}c@{}}$\mathcal{O}(r \cdot n \cdot \log\,n\cdot$ \\ $C_{e} \cdot c \cdot m^{2})$\end{tabular}}} & \multirow{2}{*}{\textbf{\begin{tabular}[c]{@{}c@{}}$\mathcal{O}(n \cdot m + r \cdot n \cdot c$ \\$+ D_{t} \cdot n \cdot m)$\end{tabular}}} & \multirow{2}{*}{\textcolor{blue}{Yes}} & \multirow{2}{*}{No}
\\ 
\\
\\
\textbf{\textcolor{black}{InceptionTime}} & * & * & \textcolor{blue}{Yes} & No
\\
\\
\multirow{2}{*}{\textbf{\textcolor{black}{ROCKET}}} & \multirow{2}{*}{\textbf{\begin{tabular}[c]{@{}c@{}}$\mathcal{O}(Knm + n^{2}v)$ or \\ $\mathcal{O}(Knm + nv^{2})$\end{tabular}}} & \multirow{2}{*}{*} & \multirow{2}{*}{\textcolor{blue}{Yes}} & \multirow{2}{*}{No}\\\\
\bottomrule
\end{tabular}
\end{table}
\vspace{-5mm}

Given the high complexity of HIVE-COTE, recent approaches such as \emph{proximity forest} (PF)~\citep{lucas2019proximity} and \emph{time series combination of heterogeneous and integrated embedding forest} (TS-CHIEF)~\citep{shifaz2020ts} focus on accurate and scalable classifications. PF builds an ensemble of proximity trees for classification. Proximity trees use elastic distance measures as the splitting criteria. While PF is more scalable (quasi-linear time complexity over the number of time series) than HIVE-COTE, it is much less accurate. TS-CHIEF builds on PF and incorporates BOSS and RISE features as splitting criteria. TS-CHIEF is statistically similar to HIVE-COTE in classification accuracy, but more scalable (similar to PF). PF and TS-CHIEF are quadratic to the series length, which make them costly for long series.

\emph{Deep learning classifiers} such as the \emph{fully convolutional networks} (FCN) and \emph{residual networks} (ResNet) obtain competitive  accuracy~\citep{wang2017time} and enable interpretations on the model decisions using the \emph{class activation map} (CAM)~\citep{zhou2016learning} to highlight relevant sub-series. However, they are also computationally expensive for long series and require a GPU. InceptionTime~\citep{fawaz2020inceptiontime} is the best deep learning-based classifier, and it is competitive to SOTA TSC methods. It improves ResNet's accuracy by using an ensemble of five different Inception networks which are randomly initialized. InceptionTime is faster than HIVE-COTE as it uses GPU parallelization; however, it is very slow when running without GPU~\citep{middlehurst2020usage} and cannot provide interpretability.

The recently proposed \emph{RandOm Convolutional KErnel Transform} (ROCKET) \citep{dempster2020rocket} is optimized for speed. ROCKET transforms the series into a feature-based representation by using random convolutional kernels, and it uses such features to train a linear classifier. ROCKET achieves SOTA classification accuracy, but the (complex) features extracted by ROCKET do not allow for result interpretation.  Besides, ROCKET reports a time complexity quadratic to the number of time series or to the number of extracted features (depending on which is smaller). Large datasets with very long series may affect ROCKET's scalability.

\tableref{table:related_work} summarizes the representative TSC methods. Most current TSC methods have time complexities that are quadratic or bi-quadratic to the length of the time series, which makes them less practical on long time series. Besides, to the best of our knowledge, with the exception of TSF, ST, FCN, ResNet, \textcolor{black}{mtSAX-SEQL+LR, STSF and CIF} no existing TSC methods provide insights about the classifier decision.

\section{Preliminaries}
\label{sec:preliminaries}
We take an interval-based approach to classify time series and identify discriminatory features. 
The basic idea is to extract sub-series for which aggregates (e.g., mean and standard deviation) are computed and used as features. To allow for interpretable classifications, we focus on \emph{phase-dependent intervals}, i.e., discriminatory features located at the same time regions over all time series in a given dataset.

\textbf{Interval feature.}
\textit{Given a set of time series $X$ = $\{ \bm{x^{1}}, \bm{x^{2}},\bm{x^{3}},...,\bm{x^{n}} \}$, where $\bm{x^{i}}$ = $\{x_{1}^i,x_{2}^i,...,x_{m}^i\}$, an aggregation function $a(\cdot)$, and an interval ($s$,$e$), an interval feature \boldsymbol{$f$} = $a(X,s,e)$ is a vector of length $n$, defined as follows:
}
\begin{center}
    \boldsymbol{$f$} = $\{a(\boldsymbol{x}^1,s,e),a(\boldsymbol{x}^2,s,e),a(\boldsymbol{x}^3,s,e),\ldots,a(\boldsymbol{x}^n,s,e)\}$
\end{center}
\textit{where $a(\boldsymbol{x}^i,s,e) = a(\{x_{s}^i,x_{s+1}^i,...,x_{e-1}^i,x_{e}^i\})_{ 1 \leqslant s \leqslant e \leqslant m}$}.
% \textit{where} 
% \begin{center}
% $f(\boldsymbol{x}^i,t_{s},t_{e}) = f(\{x_{t_{s}}^i,x_{t_{s+1}}^i,...,x_{t_{e-1}}^i,x_{t_{e}}^i\})_{ 1 \leqslant t_{s} \leqslant t_{e} \leqslant m} $ 
% \end{center}
% \vspace{5mm}
% For example,
With $s$=2, $e$=4, $a$=\mbox{mean}, an interval feature $\bm{f}=\mbox{mean}(X,2,4)$ is represented with the dashed rectangle as:
\\

\begin{center}
$X$ = $\begin{pmatrix}
\tikzmarknode{}{x_{1}^1} & \tikzmarknode{}{x_{2}^1} & \tikzmarknode{}{x_{3}^1} & \tikzmarknode{}{x_{4}^1} & \tikzmarknode{}{x_{5}^1} & \tikzmarknode{}{x_{6}^1} & \tikzmarknode{}{\ldots}  & \tikzmarknode{}{x_{m-1}^1} & \tikzmarknode{}{x_{m}^1}\\
\addlinespace

\tikzmarknode{}{x_{1}^2} & \tikzmarknode{}{x_{2}^2} & \tikzmarknode{}{x_{3}^2} & \tikzmarknode{}{x_{4}^2} & \tikzmarknode{}{x_{5}^2} & \tikzmarknode{}{x_{6}^2} & \tikzmarknode{}{\ldots}  & \tikzmarknode{}{x_{m-1}^2} &
\tikzmarknode{}{x_{m}^2}\\

\tikzmarknode{}{\vdots} & \tikzmarknode{}{\vdots} & \tikzmarknode{}{\vdots} & \tikzmarknode{}{\vdots} & \tikzmarknode{}{\vdots} & \tikzmarknode{}{\vdots} & \tikzmarknode{}{\vdots}  & \tikzmarknode{}{\vdots} & \tikzmarknode{}{\vdots}\\

\tikzmarknode{}{x_{1}^n} & \tikzmarknode{}{x_{2}^n} & \tikzmarknode{}{x_{3}^n} & \tikzmarknode{}{x_{4}^n} & \tikzmarknode{}{x_{5}^n} & \tikzmarknode{}{x_{6}^n} & \tikzmarknode{}{\ldots}  & \tikzmarknode{}{x_{m-1}^n} & \tikzmarknode{}{x_{m}^n}
\end{pmatrix}$
\end{center}

\begin{tikzpicture}[overlay,remember picture]

\draw[red,thick,dashed] (4.0,-0.01) rectangle (5.33,2.5);

\draw (4.67,2.2) ellipse (0.63 and 0.35);
\draw[->,shorten <=2pt] (5.2,2.25) -- (5.7,2.7) node[right]{mean$(\{x_{2}^1,x_{3}^1,x_{4}^1\})$};

\draw (4.67,0.45) ellipse (0.63 and 0.35);
\draw[->,shorten <=2pt] (5.2,0.32) --  (5.7,-0.01) node[right]{mean$(\{x_{2}^n,x_{3}^n,x_{4}^n\})$};
\end{tikzpicture}
\\

\textbf{Problem statement.}
Consider a set of $n$ univariate time series $X$ = $\{ \bm{x^{1}}, \bm{x^{2}},...,\bm{x^{n}} \}$, where each time series $\bm{x^{i}}$ = $\{x_{1}^i,x_{2}^i,...,x_{m}^i\}$ has $m$ ordered real-valued observations, sampled at equally-spaced time intervals. Each time series $\bm{x^{i}}$ is also associated with a class label $y^{i}$. We aim to find the set of interval features that yield the highest time series class prediction accuracy.
Finding such a  set of interval features is NP-hard. For a time series of length $m$, there are $\mathcal{O}(m^{2})$ different intervals, and hence $\mathcal{O}(2^{m^2})$ subsets of intervals. 
For a large $m$, it is prohibitively expensive to explore all subsets. We present an efficient heuristic to avoid the exhaustive search while retaining a high classification accuracy\\

\tableref{table:symbols} summarizes the symbols frequently used in this paper.

\begin{table}[h]
\centering
\caption{Frequently used symbols and their meanings}
\label{table:symbols}
\setlength{\tabcolsep}{5pt}
\begin{tabular}{@{}ll@{}}
\toprule
\textbf{\begin{tabular}[c]{@{}c@{}}Symbol \end{tabular}} & 
\textbf{\begin{tabular}[c]{@{}c@{}}Meaning \end{tabular}} 
\\ \midrule

$UCR_{85}$  & the 85 benchmark time series datasets from the UCR repository \\
$n$ & number of time series instances \\
$m$ & time series length \\
$X$ & time series set of size $n \times m$ \\
$y$ & class labels vector of size $n \times 1$ \\
$c$ & number of class labels \\
$\bm{x}^{i}$ & time series instance \\
$y^{i}$ & class label \\
$\bm{f}$ & interval feature of size $n \times 1$ \\
$a(\cdot)$ & aggregation function \\
$\intercal$ & tree classifier \\
$r$ & number of tree classifier in the ensemble \\
$X_{O}, X_{P}, X_{D}, X_{G}$ & original (raw), periodogram, derivative and autoregressive representations of $X$, respectively \\
% $X_{O}$ & raw (original) time series from $X$ \\
% $X_{P}$ & periodogram (frequency domain) representation of series from $X$ \\
% $X_{D}$ & derivative representation of series from $X$ \\
% $X_{G}$ & autoregressive representation of series from $X$ \\
$F_{O}, F_{P}, F_{D}, F_{G}$ & candidate discriminatory interval features extracted from $X_{O}, X_{P}, X_{D}$ and $X_{G}$, respectively  \\
$d$ & number of sets of candidate discriminatory interval features \\
$\mathcal{F}$ & set of candidate discriminatory interval features (union of $F_{O}, F_{P}, F_{D}, F_{G}$)  \\
$\mathcal{F}^{*}$ & set of discriminatory interval features \\ 
$i, j, k$ & indices to iterate across time series instances, time series data points, class labels, etc. \\

\bottomrule
\end{tabular}
\end{table}

\section{Our Approach}
\label{sec:approach}
\begin{figure}[h]
    \centering
    \includegraphics[scale=0.20]{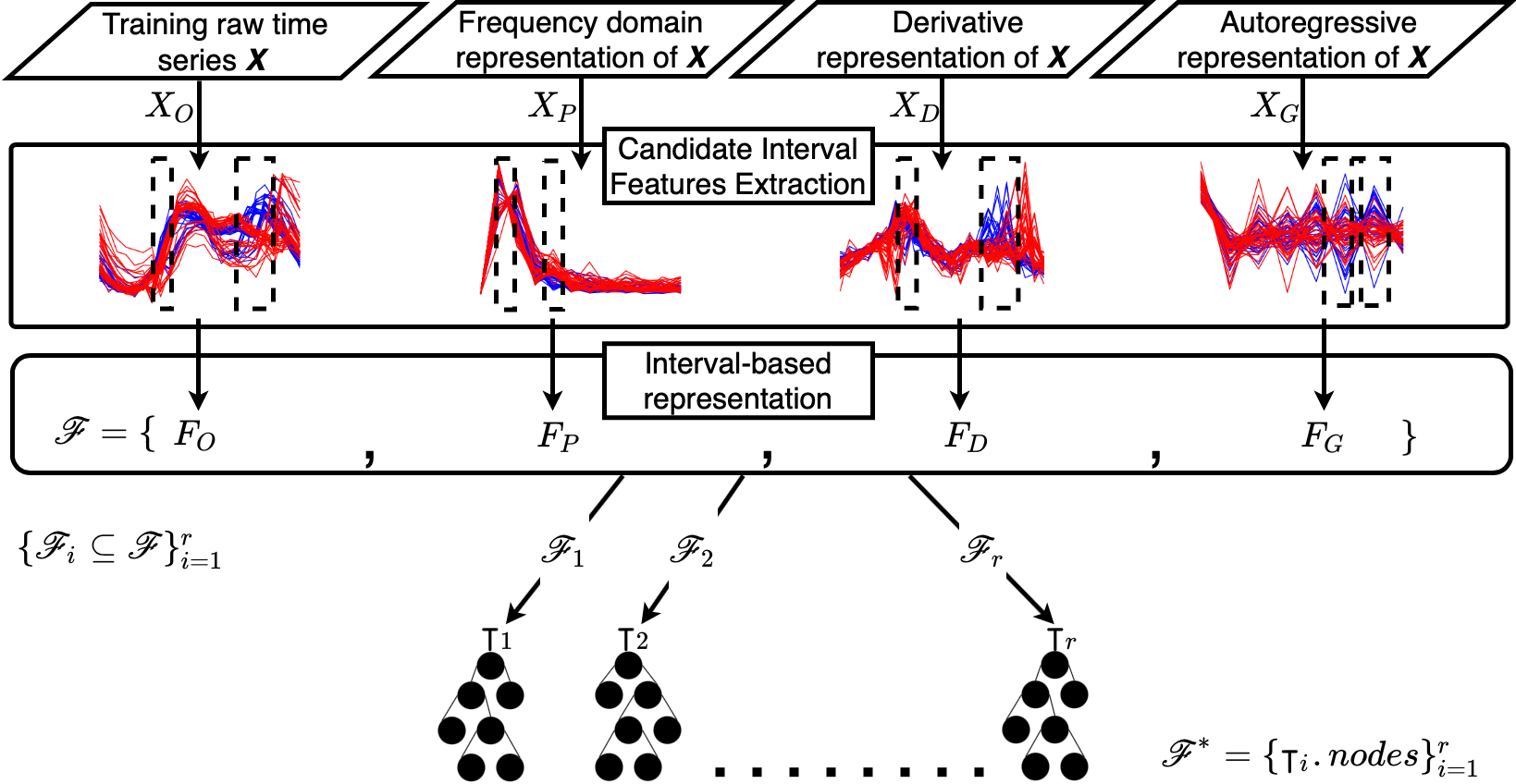}

    \caption{\textcolor{black}{Overview of r-STSF when training an ensemble of $r$ randomized trees. Sets of candidate discriminatory interval features $F_{O}, F_{P}, F_{D}, F_{G}$ are selected from the time series representations $X_{O}, X_{P}, X_{D}, X_{G}$, respectively. All sets of candidate discriminatory interval features are merged into a single superset $\mathcal{F}$. Each randomized tree is built by using a number of randomly selected interval features from $\mathcal{F}$. It is worth noting that each tree node selects its own group of interval features (from $\mathcal{F}$) for the split. The set of discriminatory intervals $\mathcal{F}^{*}$ from the nodes of the randomized trees offers interpretability to the classification outcome}
    } 
    \label{fig:overview}
\end{figure}

Our proposed TSC algorithm, \emph{r-STSF}, takes a stochastic optimization approach to select a set of interval features with a high discriminating power (i.e., \emph{candidate discriminatory interval features}) from the high dimensional interval feature space (i.e., all $\mathcal{O}(m^2)$ possible interval features). For a time series of length $m$, we reduce the interval feature space size to $\mathcal{O}(\log\,m)$. We search for the best interval features subset $\mathcal{F}^*$
(i.e., \emph{discriminatory interval features}) through an ensemble of binary trees, which has a time complexity of $\mathcal{O}(r \cdot n \cdot \log\,n \cdot \log\,m)$, where $r$ is the total number of trees in the ensemble and $n$ is the number of time series instances. \textcolor{black}{The trees in the ensemble are built in a randomized manner following the extra-trees algorithm} (see Section~\ref{subsec:randomizedtrees}) to reduce the variance of the ensemble and improve the classification accuracy. 
\figref{fig:overview} shows an overview of r-STSF when training the ensemble of \textcolor{black}{randomized} binary trees. 

For a given time series training set $X$, r-STSF does not only uses its original (raw) representation, i.e., $X_O$, but derives its periodogram, i.e., $X_P$, derivative, i.e., $X_D$, and autoregressive representation, i.e., $X_G$. For each representation, r-STSF selects a group of candidate discriminatory interval features $F_{O},\ F_{P},\ F_{D}$, and $F_{G}$, respectively.
\textcolor{black}{Next, a union set $\mathcal{F}$ is formed by merging all candidate discriminatory interval features. It is worth noting that the process of extracting candidate interval features is repeated for $d$ times, where $d$ is a system constant parameter. In other words, $d$ sets of candidate discriminatory interval features are extracted. In our empirical study, we find that a small value, e.g., $d = 0.1 \times r$,  is sufficient to yield a high TSC accuracy. This means that, for an ensemble of 500 trees, we just extract 50 sets of candidate discriminatory intervals. Lastly, each randomized tree is built by using a number of randomly selected interval features from $\mathcal{F}$. We set this number to the square root of the size of $\mathcal{F}$ following other tree-based ensemble approaches such as \emph{random forest} (RF)~\citep{breiman2001random}. A tree classifier, due to its intrinsic feature selection capability, enables the selection of a set of discriminatory intervals $\mathcal{F}^*$ with which TSC is performed. Features from $\mathcal{F}^*$ enable the interpretability of the TSC task}, \textcolor{black}{which will be discussed in Section~\ref{sec:roi}}.

\subsection{Time series representation}\label{subsec:representation}
We first discuss time series representations to extract intervals features. We use intervals from original (i.e., the time domain as described in Section 3), periodogram (i.e., the frequency domain), derivative, \textcolor{black}{and autoregressive} representations. We focus on the latter \textcolor{black}{three} representations.

\begin{figure}[h]
\centering
\begin{tabular}{c c}

\subf{\includegraphics[scale=0.1]{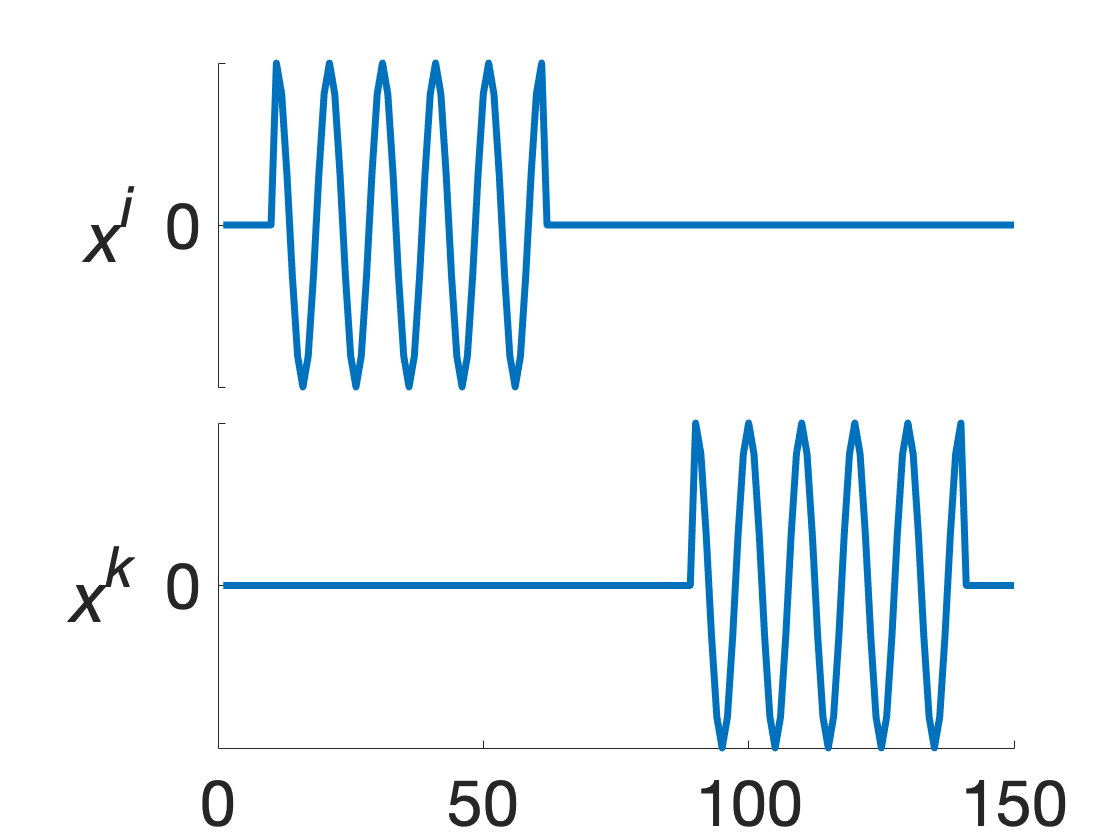}}
     {(a)}

\subf{\includegraphics[scale=0.1]{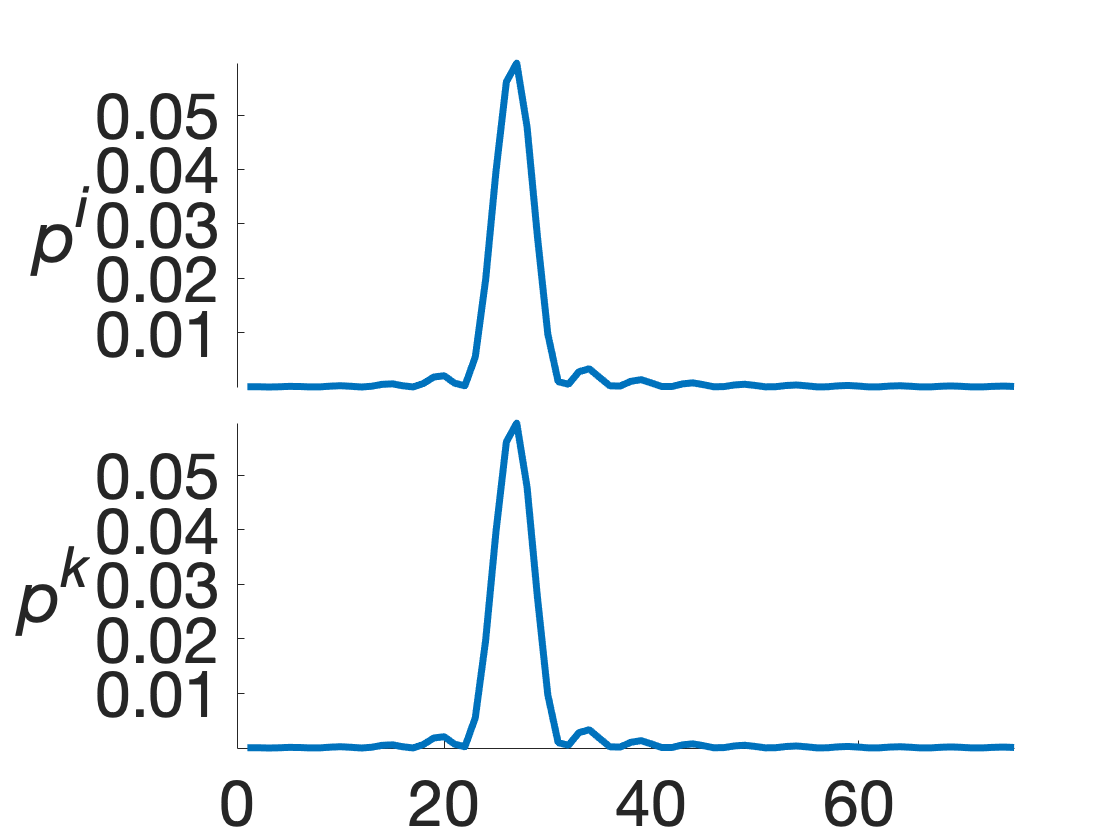}}
     {(b)}

\end{tabular}
\caption{\textbf{(a)} Time series $x^i$ and $x^k$, with a same class label (i.e., $y^i=y^k$) present a similar sub-series but located at different time regions (i.e., phase-independent). \textbf{(b)}~The periodogram representation $p^i$ and $p^k$ of series $x^i$ and $x^k$, respectively. Our algorithm searches for discriminatory phase-dependent interval features, which in (a) are challenging to identify. 
The periodogram representation provides more flexibility as it considers the frequency of the discriminatory sub-series (ignoring its location in time) and thus helps  identify discriminatory sub-series even when they appear at non-identical location 
in time across different time series}
\label{fig:periodogram}
\end{figure}

\textbf{Periodogram representation:}
Several TSC methods~\citep{bagnall2012transformation,lines2018time} use the \emph{periodogram representation} when examining time series similarity in the frequency domain. We adopt this and exploit the periodogram representation of each time series derived from the \emph{discrete Fourier transform} (DFT). 
The DFT decomposes a real-valued time series $\bm{x^{i}}$ into a linear combination of sinusoidal functions with amplitudes $o$ and $q$, and phase $\omega$:

\begin{equation}
\label{eq:fft}
\bm{x_{t}} = \sum\nolimits_{j=1}^{m} ( o_{j}\cos({2 \pi \omega_{j} t}) + q_{j}\sin({2 \pi \omega_{j} t}) )
\end{equation}
The periodogram $\bm{x^{i}_{P}}$ = $\{p_{1}^i,p_{2}^i,...,p_{m}^i\}$ of  series $\bm{x^{i}}$ is represented using the set of amplitudes $\{(o_{1},q_{1}),(o_{2},q_{2}),...,(o_{m},q_{m})\}$ from Equation~\eqref{eq:fft}, i.e., $p_{j}^{i} = \sqrt{o_{j}^{2} + q_{j}^{2}}$.
A property of the DFT of a real-valued series is that it is symmetric, i.e., $(o_{j},q_{j}) = (o_{m-j-1},q_{m-j-1})$. 
Thus, we can shrink the size of the periodogram by half, resulting in  $\bm{p^{i}}$ = $\{p_{1}^i,p_{2}^i,...,p_{m/2}^i\}$. For long series, this reduces the computation cost substantially in assessing the discriminatory power of the interval features. A side benefit of this representation is that it helps to \emph{indirectly} detect \emph{phase-independent discriminatory intervals,} i.e., discriminatory features located at different time regions of the original series. 
Take~\figref{fig:periodogram} as an example. r-STSF assesses the similarity of a group of series based on the discriminatory power of the extracted phase-dependent interval features. Hence, time series $x^i$ and $x^k$, with a same class label (i.e., $y^i=y^k$) but time-shifted discriminatory interval, are more likely to be identified as similar by using the periodogram representation of the series (i.e., $p^i$ and $p^k$) rather than their original (time domain) representation.

\textbf{Derivative representation:} Using a (first-order) derivative representation of a time series rather than the original time series improves the classification accuracy \citep{keogh2001derivative,gorecki2013using} as it provides trend information.
Given a time series $\bm{x^{i}}$, its  derivative representation is $\bm{x_{D}^{i}} = \{x_{2}^i-x_{1}^i,x_{3}^i-x_{2}^i,...,x_{m}^i-x_{m-1}^i\}$.

\textbf{Autoregressive representation:} An autoregressive model, as shown in Eq. \eqref{eq:autoreg}, assumes that the current observation of a time series, $x_{t}$, can be explained as a linear combination of its past $l$ observations, model coefficients $\{\beta_{i}\}_{i=1}^{l}$, a constant $b$, and the error term $\epsilon_{t}$. 

\begin{equation}
\label{eq:autoreg}
x_{t} = b + \sum\nolimits_{k=1}^{l} \beta_{k}x_{t-k} + \epsilon_{t}
\end{equation}

If a time series dataset satisfies this assumption, \emph{the time series can be modeled as an autoregressive process which can be used for the classification task, since series from different classes may have different autoregressive models}. The model (autoregressive) coefficients $\{\beta_{k}\}_{k=1}^{l}$ form an autoregressive representation. Given a time series $\bm{x^{i}}$, its autoregressive representation is $\bm{x_{G}^{i}} = \{\beta_{1}^i, \beta_{2}^i,...,\beta_{l-1}^i, \beta_{l}^i\}$.
Three well-known methods to estimate such coefficients are the \emph{ordinary least-squares} (OLS), \emph{Yule-Walker}, and \emph{Burg's method}. We use Burg's method as it usually yields better model fitting than the Yule-Walker approach, and it is computationally efficient~\citep{proakis2014, brockwell2002introduction}, while OLS has a time complexity quadratic to the length of the time series. Moreover, we set the lag order  $l = 12(m/100)^{1/4}$, where $m$ is the series length, as suggested by~\cite{schwert2002tests}.

The idea of using autoregressive coefficients for classification has been previously adopted in TSC methods~\citep{lines2018time}. Nonetheless, unlike these approaches, we do not use the autoregressive coefficients directly as features. Instead, we build an autoregressive representation (i.e., series of autoregressive coefficients), from which interval features are derived by using a set of aggregation functions. Our insight is that whilst the individual coefficients may not be discriminatory, the summary representation of a group of them is likely to retain more discriminatory information.

\subsection{Extraction of Candidate Discriminatory Interval Features}
\label{subsec:intervalfeaturesextraction}
Our process to extract candidate discriminatory interval features is summarized in Algorithms~\ref{algo:candidate_interval_feats} and~\ref{algo:binarysearch}. This process is similar to that in STSF. The key difference is that, unlike STSF, r-STSF does not partition each interval into halves to assess the \textit{quality} of sub-intervals.  Instead, r-STSF creates partitions with random cut points (Lines 4 and 5 in Algorithm~\ref{algo:binarysearch}).
This strategy plays a significant role in further boosting the TSC accuracy (see Section~\ref{subsec:perturbation-scheme}), since it allows to explore a larger number sub-intervals and hence generates less correlated trees.

\begin{algorithm}
\footnotesize
\caption{GetIntervalFeatures}
\label{algo:candidate_interval_feats}
\KwIn{ $X$: set  of $n$ time series of length $m$; $y$: class label vector; $A$: set of aggregation functions; $fr$: feature ranking metric.}

$\mathcal{F} \leftarrow \emptyset$\;

\For{$each $ time series representation}{
    $X \text{ changes according to the corresponding representation}$\;
    \For{$each \ a \in A$}{
        $u \leftarrow \mbox{GetRandomCutPoint}(m)$\;
        $X_{L} \leftarrow X(1:u)$; $X_{R} \leftarrow X(u+1:m)$\;
        $F_{L} \leftarrow  \mbox{SupervisedIntervalSearch}(X_{L},y,a,fr,\emptyset)$\; $F_{R} \leftarrow  \mbox{SupervisedIntervalSearch}(X_{R},y,a,fr,\emptyset)$\;
        $\mathcal{F} \leftarrow \mathcal{F} \cup \{F_{L} + F_{R}\}$\;
    }
}
\Return $\mathcal{F}$\;
\end{algorithm}
\normalsize

\begin{algorithm}
\footnotesize
\caption{SupervisedIntervalSearch}
\label{algo:binarysearch}
\KwIn{ $X'$: set of $n$ time series of length $m'$;
$y$: class label vector; $a$: aggregation function; $fr$: feature ranking metric; $F$: subset of candidate discriminatory intervals.}
\eIf{$m' < 2$} {
\Return $F$\;
}{
\color{black}
 $u \leftarrow \mbox{GetRandomCutPoint}(m')$\;
  $f_{L} \leftarrow a(X',1, u)$; 
  $f_{R} \leftarrow a(X',u, m)$\;

 $score_{L} \leftarrow fr(f_{L},y)$; $score_{R} \leftarrow fr(f_{R},y)$\;
\eIf{$score_{L} >= score_{R}$}{
 $F \leftarrow F \cup \{ f_{L} \}$\;
 $\mbox{SupervisedIntervalSearch}(X'(1:u),y,a,fr,F)$\;
}{
 $F \leftarrow F \cup \{ f_{R} \}$\;
 $\mbox{SupervisedIntervalSearch}(X'(u:m),y,a,fr,F)$\;
}
}
\Return $F$\;
\end{algorithm}

{\bf Feature ranking metric:}
We assess the quality of sub-intervals using Fisher Score ~\citep{duda2012pattern} as the feature ranking metric. The Fisher score of an interval feature indicates how well the  feature separates a class of time series from the other classes. A higher Fisher score suggests a more discriminatory feature. For a given time series subset of size $n\times m'$, an interval feature $\bm{f}$ of size $n\times 1$ is extracted (by applying $a(\cdot)$ on each time series row-wise). Interval feature $\bm{f}$ and a vector of class labels $\boldsymbol{y} \in \{1,2,...,c\}^{n}$ are used to compute the Fisher score to obtain the discriminatory quality of $\bm{f}$ as follows:
\begin{equation}
\label{eq:fisherscore}
\mbox{FisherScore}(\bm{f},\bm{y}) = \big(\sum\nolimits_{k=1}^{c} n_{k}(\mu^{\bm{f}}_{k} - \mu^{\bm{f}})^2 \big) / \big(\sum\nolimits_{k=1}^{c} n_{k}(\sigma^{\bm{f}}_{k})^2\big)
\end{equation}
\normalsize
Here, $\mu^{\bm{f}}$ is the overall mean of the elements in 
$\bm{f}$; $\mu^{\bm{f}}_{k}$ and $\sigma^{\bm{f}}_{k}$ 
are the mean and standard deviation of the elements in $\bm{f}$ labelled with the $k$-th class; and $n_{k}$ is the number of time series labelled with the $k$-th class. 

Other feature ranking metric such as Fast Correlation Based Filter (FCBF)~\citep{yu2003feature} or ReliefF~\citep{robnik2003theoretical} could also be used. We use Fisher score for its fast computation. Computing Fisher scores does need to search for the best split when assessing the relevance of features as in FCBF, or search for the nearest instances as in ReliefF.

{\bf Aggregation functions:}
Standard interval-based classifiers such as TSF use \emph{mean}, \emph{standard deviation} (\emph{std}), and \emph{slope}  (i.e., slope of the least-squares regression line) aggregation functions to obtain a representative value for the sub-series. Others interval-based methods such as STSF expand the set of aggregation functions by adding more robust statistics such as the \emph{median} and the \emph{interquartile range} (\emph{iqr}). STSF also incorporates the \emph{minimum} (\emph{min}) and \emph{maximum} (\emph{max}) statistics  as they can potentially detect discriminatory extreme values. According to our experiments, the \emph{slope} is the aggregation function that contributes the most to the classification accuracy of STSF. This aggregation function captures the shape of the series, and it is the only one in the set of statistics used by STSF that can  do so. \emph{To reinforce capturing a higher level of details regarding the shape of the sub-series}, we further propose two aggregation functions: (i) \emph{counts of mean-crossings} (\emph{cmc}) (i.e., the number of intersections of the sub-series with the mean axis of the entire time series), and (ii) \emph{counts of values above the mean} (\emph{cam}) (i.e., the number of data points above the mean of a sub-series). Hence, r-STSF uses nine aggregations to compute the interval features: \emph{cmc}, \emph{cam} and the seven aggregation proposed in STSF.

\subsection{Classification with Randomized Trees}
\label{subsec:randomizedtrees}
Our algorithm creates \emph{randomized binary trees} for classification (summarized in Algorithm~\ref{algo:tree_generation}). In such trees, the cut-point of each feature is randomly selected when looking for the feature that provides the best split (i.e., best random split).
This perturbation strategy was originally proposed by~\cite{geurts2006extremely}
for the \emph{extra-trees} (ET) algorithm. 
The idea is to create an ensemble of uncorrelated trees (i.e., different tree models), which decreases the estimation variance (i.e., variability of the predictions) and hence decreases the classification error. 
We use this strategy to decrease the similarities among the trees in our ensemble. We build our trees node-by-node recursively starting from the root node. At each node, we do not inspect all possible cut-points for each feature (to split that node). Instead, we \emph{randomly} select one value $cp$ as the cut-point (for each feature) (Line 7). This cut-point splits the learning sample $LS$ and its class label vector $y$ into two sets each. The rows of the learning sample and label vector where the values of a feature are less than or equal to its corresponding cut-point are sent to $LS_{L}$ and $y_{L}$, respectively (Line 8). The remaining rows of $LS$ and $y$ are sent to $LS_{R} $ and $y_{R}$, respectively (Line 9). Then, we compute the \emph{information gain} (IG, Eq.~\eqref{eq:ig}) for each split (Line 10) and use the split with the maximum gain (Lines 11 to 15) to expand our tree (Lines 17 and 18). If the node cannot be further split, i.e., all the samples have the same class label, the node becomes a leaf or a terminal node (Lines 1 to 4).

\begin{algorithm}
\footnotesize
\caption{CreateRandomTree}
\label{algo:tree_generation}
\KwIn{ $LS$: learning sample of size $n \times v$, $v$ interval features, each feature (or column) with $n$ values; $y$: class label vector of size $n \times 1$.}
 \If{\text{cannot further split LS}}{
     \text{Designate this node as a terminal node}\;
     \Return
     }
 $maxIG \leftarrow 0$\;
\For{$each \ f \in \text{columns of }LS$}{
    $cp \leftarrow \text{Pick a value from feature } f$\;
    $LS_{L} \leftarrow LS(f \leqslant cp)$;
    $y_{L}\leftarrow y(f \leqslant cp)$\;
    $LS_{R} \leftarrow LS(f > cp)$; 
    $y_{R}\leftarrow y(f > cp)$\;
    $currentIG \leftarrow 
    \mbox{IG}(y,y_{L},y_{R})$\;
    \If{$currentIG > maxIG$}{
     $maxIG \leftarrow currentIG$\;
     $best_{LS_{L}} \leftarrow LS_{L}$; $best_{LS_{R}} \leftarrow LS_{R}$\;
     $best_{y_{L}} \leftarrow y_{L}$; $best_{y_{R}} \leftarrow y_{R}$\;
     }
}
$\mbox{CreateRandomTree}(best_{LS_{L}},best_{y_{L}})$\;
$\mbox{CreateRandomTree}(best_{LS_{R}},best_{y_{R}})$\;
\end{algorithm}

For a given vector of class labels $\boldsymbol{y} = \{y^{1},y^{2},...,y^{i},...,y^{n}\}$ where $y^{i} \in \{1,2,...,c\}$, and its corresponding subsets after split, $\boldsymbol{y}_{L}$ and $\boldsymbol{y}_{R}$, the IG is given by:

\begin{equation}
    \label{eq:ig}
    \hbox{IG}(\boldsymbol{y},\boldsymbol{y}_{L},\boldsymbol{y}_{R}) =             \hbox{H}(\boldsymbol{y}) 
        - 
        \left[ 
            \frac{|\boldsymbol{y}_{L}|}{|\boldsymbol{y}|} \hbox{H}(\boldsymbol{y}_{L}) 
            + 
            \frac{|\boldsymbol{y}_{R}|}{|\boldsymbol{y}|} \hbox{H}(\boldsymbol{y}_{R}) 
        \right]
\end{equation}

Here, $\hbox{H}$ is the entropy (i.e., impurity measure), which is computed as
$\hbox{H}(\boldsymbol{y}_{*}) = - \sum_{k=1}^{c} \rho_{k} \log  \rho_{k}$, 
where $\rho_{k} = |\{y^{i}|y^{i}=k; y^{i} \in \boldsymbol{y_{*}}\}|/|\boldsymbol{y}_{*}|$ is the relative frequency of class label $y^{i}$ in $\boldsymbol{y_{*}}$.

\subsection{r-STSF vs. STSF}
\label{subsec:rstsfVSstsf}

r-STSF extends from STSF. \tableref{table:rstsfVSstsf} summarizes the main differences between r-STSF and STSF. The additional time series representation and aggregation functions and our novel perturbation scheme (i.e., supervised search through random partitions of sub-series and randomized binary trees to select discriminatory interval features) considerably improve the classification accuracy of r-STSF over its predecessor STSF (see Section~\ref{subsec:sota-comparison}). Further, r-STSF's approach to train its tree-based ensemble is significantly more efficient than that of STSF. r-STSF requires an order of magnitude less computations when searching for candidate discriminatory interval features (see Section~\ref{subsec:running_time}). It is worth noting that on STSF each tree node uses the same group of features (given as input to their corresponding tree) when looking for the best split whereas on r-STSF each tree node uses a different group of randomly selected features from $\mathcal{F}$ when looking for the best random split.

\begin{table}[h]
\centering
\caption{Main differences between r-STSF and STSF}
\label{table:rstsfVSstsf}
\resizebox{\textwidth}{!}{%
\begin{tabular}{@{}lccccl@{}}
\toprule
\textbf{}                 & \multirow{2}{*}{\textbf{\begin{tabular}[c]{@{}c@{}}Time Series \\ Representations\end{tabular}}} & \multirow{2}{*}{\textbf{\begin{tabular}[c]{@{}c@{}}Aggregation \\ Functions\end{tabular}}} & \multirow{2}{*}{\textbf{\begin{tabular}[c]{@{}c@{}}Supervised \\ Search\end{tabular}}} & 
\multirow{2}{*}{\textbf{\begin{tabular}[c]{@{}c@{}}Tree Classifier ($\intercal$)\end{tabular}}} &
\multirow{2}{*}{\textbf{\begin{tabular}[l]{@{}l@{}}Train Ensemble of \emph{r} Trees\end{tabular}}}
\\
\\ \midrule
\textbf{STSF}                   & \multirow{3}{*}{\begin{tabular}[c]{@{}c@{}} raw series, \\ periodogram, \\ derivative \end{tabular}}
& 
\multirow{3}{*}{\begin{tabular}[c]{@{}c@{}} mean, std, \\ slope, median, \\ iqr, min, max  \end{tabular}}
 & 
 \multirow{2}{*}{\begin{tabular}[c]{@{}c@{}}  fixed partition \\of sub-series \end{tabular}}
 & 
 \multirow{2}{*}{\begin{tabular}[c]{@{}c@{}}  binary tree using \\
 best learned split \end{tabular}} 
 &
  \multirow{4}{*}{
    \begin{tabular}[c]{@{}l@{}}  
        \textbf{for} $i = 1$ \textbf{to} $r$ \textbf{do}\\
        \quad $feats \leftarrow \mbox{GetIntervalFeatures}()$\\
        \quad $\intercal_{i}.\hbox{build}(feats)$\\
        \textbf{end}
    \end{tabular}}
\\
\\
\\
\\
\\
\\
\\
\textbf{\textcolor{black}{r-STSF}} & 
\multirow{4}{*}{\begin{tabular}[c]{@{}c@{}} raw series, \\ periodogram, \\ derivative, \\ \emph{autoregressive} \end{tabular}}  
& 
\multirow{4}{*}{\begin{tabular}[c]{@{}c@{}} mean, std, \\ slope, median, \\ iqr, min, max, \\ \emph{cmc}, \emph{cam}  \end{tabular}}  
&
\multirow{2}{*}{\begin{tabular}[c]{@{}c@{}}  \emph{random} partition \\of sub-series \end{tabular}}
&
\multirow{3}{*}{\begin{tabular}[c]{@{}c@{}}  \emph{randomized}\\ binary tree using \\ best \emph{random} split \\ 
\end{tabular}}
& 
\multirow{8}{*}{
\begin{tabular}[c]{@{}l@{}}  
    $\mathcal{F} \leftarrow \emptyset$ \\
    \textbf{for} $j = 1$ \textbf{to} $0.1 \times r$ \textbf{do}\\
    \quad $\mathcal{F} \leftarrow \mathcal{F} \cup \mbox{GetIntervalFeatures}()$\\
    \textbf{end}\\
    \textbf{for} $i = 1$ \textbf{to} $r$ \textbf{do}\\
    \quad $feats \leftarrow \text{Pick} \sqrt{|\mathcal{F}|} \text{ features from } \mathcal{F}$ \\
    \quad $\intercal_{i}.\hbox{build}(feats)$ \\
    \textbf{end}
\end{tabular}}     
\\
\\
\\
\\
\\
\\
\\
\\
\bottomrule
\end{tabular}%
}
\end{table}

\subsection{Computation Complexity}
\label{subsec:complexity}

The training time of r-STSF depends on two processes: (i) extracting candidate interval features and (ii) building a tree ensemble with a subset of the extracted features.
 
\textbf{Candidate interval feature  extraction:} This process is summarized in Algorithm~\ref{algo:candidate_interval_feats}. The set of candidate interval features is found by a supervised search on each segment created after the initial random partition. The supervised search follows a binary-inspired search strategy (Algorithm~\ref{algo:binarysearch}). In a single run of Algorithm~\ref{algo:candidate_interval_feats}, the total number of candidate interval features extracted is $\mathcal{O}(z \cdot g\cdot \log\,m)$, where $z$ is the number of time series representations, $g$ is the number of aggregation functions, and $m$ is the time series length. Since $z$ and $g$ are constants, the total number of candidate interval features extracted in a single run of Algorithm~\ref{algo:candidate_interval_feats} is $\mathcal{O}(\log\,m)$.

For an ensemble with $r$ trees, our experiments suggest that r-STSF requires  $0.1 \times r$ runs of Algorithm~\ref{algo:candidate_interval_feats} for classification accuracy optimization. r-STSF sets the number of runs of Algorithm~\ref{algo:candidate_interval_feats} to the constant parameter $d = 50$. Hence, the total number of interval features computed to build an ensemble with $r$ trees is $\mathcal{O}(d \cdot \log\,m)$. 
Overall, the total number of interval features computed is $\mathcal{O}(\log\,m)$. The time complexity for extracting such features from $n$ (training) time series is $\mathcal{O}(n \cdot \log\,m)$.

\textbf{Tree-based ensemble construction:} The time complexity to train a single tree is $\mathcal{O}(n \cdot h \cdot v)$, where $n$ is the number of training instances (i.e., time series), $v$ is the number of interval features for node splitting, and $h$ is the depth of the tree. For a balanced tree, $h$ is in $\mathcal{O}(\log\,n)$. Also, as detailed above, the number of candidate interval features, $v$, is $\mathcal{O}(\log\,m)$. Hence, the time complexity of training a single tree is $\mathcal{O}(n \cdot \log\,n \cdot \log\,m )$. To compute $r$ trees, the time complexity is $\mathcal{O}(r \cdot n \cdot \log\,n \cdot \log\,m )$.

\textbf{Computation complexity of r-STSF:} The total training time for r-STSF is thus  $\mathcal{O}(n \cdot \log\,m) + \mathcal{O}(r \cdot n \cdot \log\,n \cdot \log\,m )  = \mathcal{O}(r \cdot n \cdot \log\,n \cdot \log\,m )$. The testing (or prediction) time complexity for r-STSF is $\mathcal{O}(r \cdot n \cdot \log\,n)$.
Similar to other interval-based approaches, our space complexity is $\mathcal{O}(r \cdot m)$.

\section{Evaluation}
\label{sec:evaluation}

We evaluate the classification effectiveness (i.e., accuracy) and computational efficiency of r-STSF in this section.
To assess effectiveness, we use two metrics: the well-known \emph{average ranking} and a metric known as \emph{weighted average accuracy} that weights classifier accuracy based on the difficulty level of the datasets (detailed in Section~\ref{subsec:waa}). 
We compare r-STSF against interpretable, SOTA, \textcolor{black}{commonly used~\citep{UCRArchive}}, and recent TSC methods in Section~\ref{subsec:effectiveness}.
To assess efficiency, we measure the algorithm's training and testing times. 
We report the overall comparison results against the competitors in Section~\ref{subsec:running_time}. 
Further, to validate r-STSF's generalizability to new time series, we show its effectiveness over an additional set of time series datasets in Section~\ref{subsec:43dsets}.

By default, r-STSF is trained with 500 trees and \textcolor{black}{uses the four time series representations and the nine aggregation functions summarized in~\tableref{table:rstsfVSstsf}.}
r-STSF is implemented in Python. All experiments are run on a macOS 10.14.3 system with 16GB RAM, dual core CPU (i5, 2.3GHz) using a single thread. We provide the experimental results and source code at \url{https://github.com/stevcabello/r-STSF}.

\subsection{Evaluation Metrics}
\label{subsec:waa}
To assess the effectiveness of a new time series classifier, it is a common practice in the TSC community to compare the average ranking of the new classifier against existing classifiers over multiple datasets. The average ranking considers all the evaluation datasets to be equally important. If the majority of the evaluation datasets are easy-to-classify (e.g., classification accuracy over 90\%), the best-ranked classifier may not be the most effective when dealing with more challenging data. Hence,~\cite{fernandez2014we} propose to weight each dataset according to  its complexity (i.e., classification difficulty). The best classifier will be the one with the highest \emph{weighted average accuracy}. For a given matrix of classification accuracy (in percentage), $Z$, of size $N_{d} \times N_{c}$, where each row represents the $i$-th dataset, and each column the $j$-th classifier, the weight of each dataset $w_{i}$ is computed as: $w_{i} = \frac{N_{d}(1-M_{i})}{N_{d}-\sum_{k=1}^{N_{d}}M_{k}}$, where $M_{i} = max_{j=1,...,N_{c}}\{Z_{ij}/100\}$ is the maximum classification accuracy achieved on dataset $i$ among the $N_c$ classifiers. 
The weighted average accuracy for the $j$-th  classifier is thus $\frac{1}{N_{d}}\sum_{i=1}^{N_{d}} w_{i} Z_{ij}$, where $j=1,...,N_{c}$.

We use both average ranking and weighted average accuracy to assess the effectiveness of r-STSF and the competitors. For a visual comparison according to the average rank, we use the critical difference diagram~\citep{demvsar2006statistical} which is widely used by the TSC community to assess the statistical differences of the ranks when comparing multiple classifiers over multiple datasets. Bold lines show groups of statistically similar methods. A smaller rank indicates a better classifier. To visualize the differences in weighted average accuracy of multiple TSC methods, we use a bar plot. A higher weighted average accuracy indicates a better classifier. The $y$-axis limits of the bar plots change according to the weighted average accuracies of the classifiers to improve the quality of the illustrations.

\subsection{Classification Effectiveness}
\label{subsec:effectiveness}
r-STSF is tested on the well-known 85 benchmark datasets from the the UCR Time Series Classification Repository~\citep{UCRArchive}, i.e., $UCR_{85}$. We use the default train and test split of each dataset (as provided in the repository). To evaluate r-STSF, we compare it against other interpretable TSC methods such as TSF~\citep{deng2013time}, ST~\citep{hills2014classification},
ResNet~\citep{wang2017time},
RISE~\citep{lines2018time}, mtSAX-SEQL+LR~\citep{le2019interpretable}, STSF~\citep{cabello2020stsf} and the recently proposed CIF~\citep{middlehurst2020canonical}. We did not include the interpretable deep-learning method FCN since ResNet (which is also an interpretable deep-learning method) is well-known to outperform FCN~\citep{fawaz2019deep}. Moreover, we compare our approach against SOTA TSC methods such as HIVE-COTE~\citep{lines2018time} (including the two new variants HC-TDE~\citep{middlehurst2020temporal} and HC-CIF~\citep{middlehurst2020canonical}), TS-CHIEF~\citep{shifaz2020ts}, InceptionTime~\citep{fawaz2020inceptiontime}, ROCKET~\citep{dempster2020rocket}, two highly accurate and \textcolor{black}{commonly used} approaches PF~\citep{lucas2019proximity} and BOSS~\citep{schafer2015boss}, and two recently proposed TSC methods TDE~\cite{middlehurst2020temporal} and CIF~\citep{middlehurst2020canonical}. In this last comparison, we also include the interval-based baseline TSF, the competitive and interpretable deep learning approach ResNet, and r-STSF's predecessor, STSF to enable further discussions on the advantages of r-STSF when compared to other interval-based and interpretable methods. Accuracy results from each TSC method are obtained from its original paper. In cases where results from a dataset are missing, we use the results reported in the UCR repository. \textcolor{black}{Missing results on TDE are completed with results from another dictionary-based TSC method (and a component of TDE) such as BOSS. Missing results on HIVE-COTE variants such as HC-TDE and HC-CIF are completed with results from HIVE-COTE.}
\textcolor{black}{~\tableref{table:comparison}, Appendix~\ref{appendix:85dsets_appendix}
shows the classification accuracies (as reported) from the majority of competitors TSC methods mentioned in this section. We did not include the classification accuracies of every competitor approach due to space limitations, however, they are included in our accompanying website.}

\subsubsection{Comparing with Interpretable TSC Methods}
\label{subsec:interpretable-comparison}
\begin{figure}[h]
\centering
\subfloat[\label{fig:interpretableclassifiers_85dsets_ar}]{
	\includegraphics[scale=0.15]{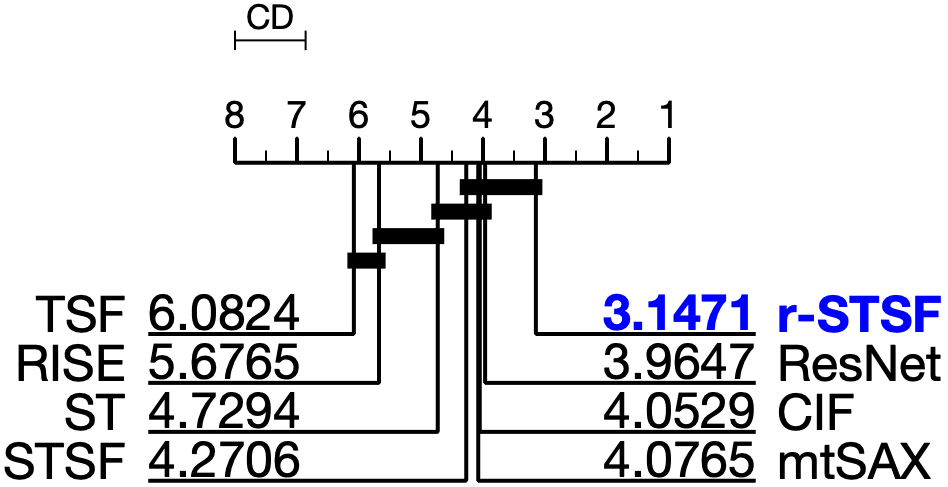}
}\hspace{5mm}
\subfloat[\label{fig:interpretableclassifiers_85dsets_waa}]{
	\includegraphics[scale=0.060]{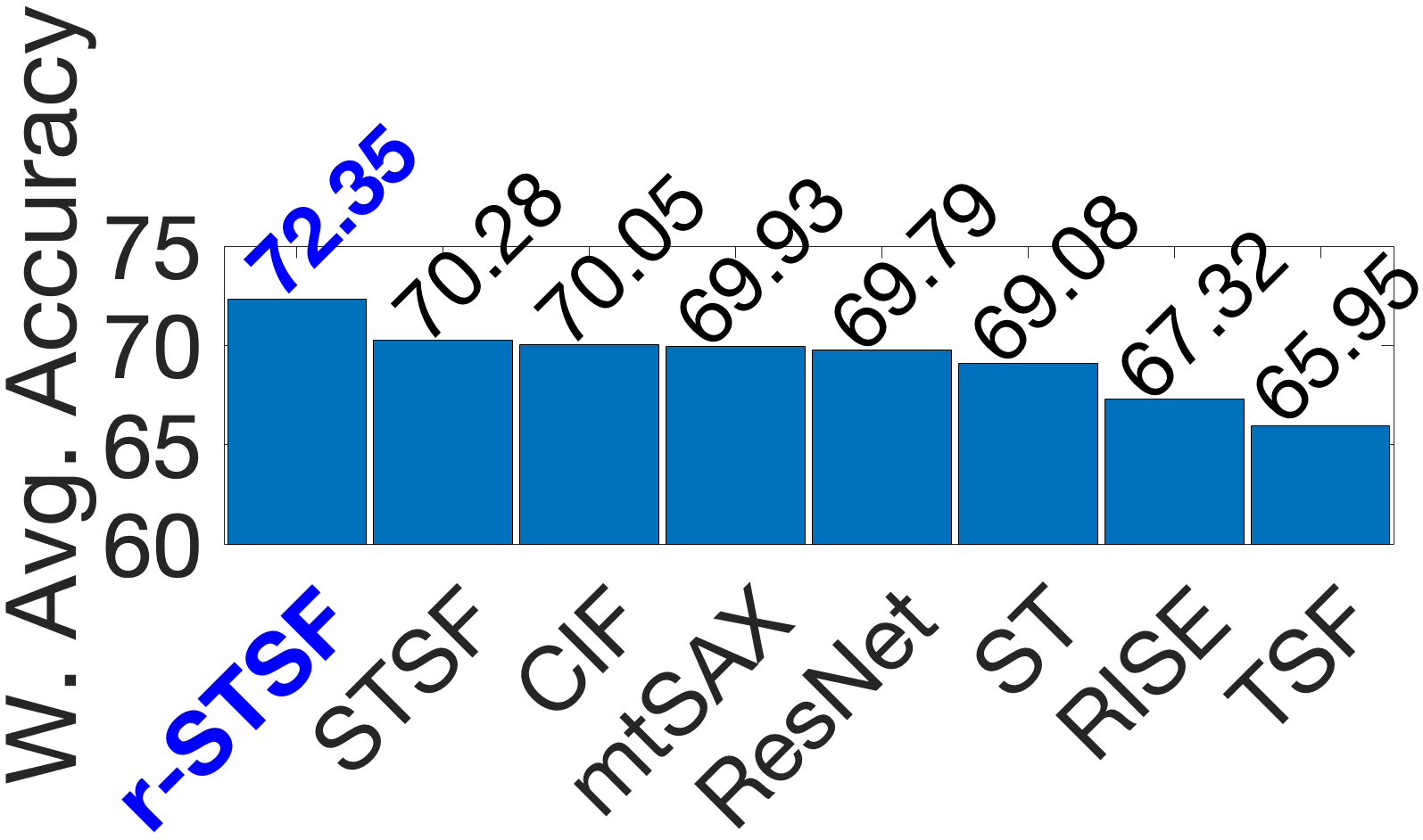}
}
\caption{\textcolor{black}{\textbf{(a)} Critical difference diagram of average ranks, and \textbf{(b)} Weighted average accuracy of r-STSF and other interpretable TSC methods on $UCR_{85}$. r-STSF has the best average rank and highest weighted average accuracy}}
\label{fig:ar_waa_interpretableclassifiers_85dsets}
\end{figure}

Interpretability has remained under-addressed due to the difficulties in achieving both interpretable and highly accurate  classification results. Intuitively, SOTA TSC methods rely on features in different transformations and complex representations. Such features are not available for TSC methods that aim for interpretability. As shown in~\figref{fig:ar_waa_interpretableclassifiers_85dsets}, r-STSF achieves the best average rank (\figref{fig:interpretableclassifiers_85dsets_ar}) and the highest weighted average accuracy (\figref{fig:interpretableclassifiers_85dsets_waa}) among the interpretable TSC methods. Besides, r-STSF is the fastest method, which will be shown later in Section~\ref{subsec:running_time}.

\begin{figure}[h]
\centering
\subfloat[\label{fig:sotaclassifiers_85dsets_ar}]{
	\includegraphics[scale=0.145]{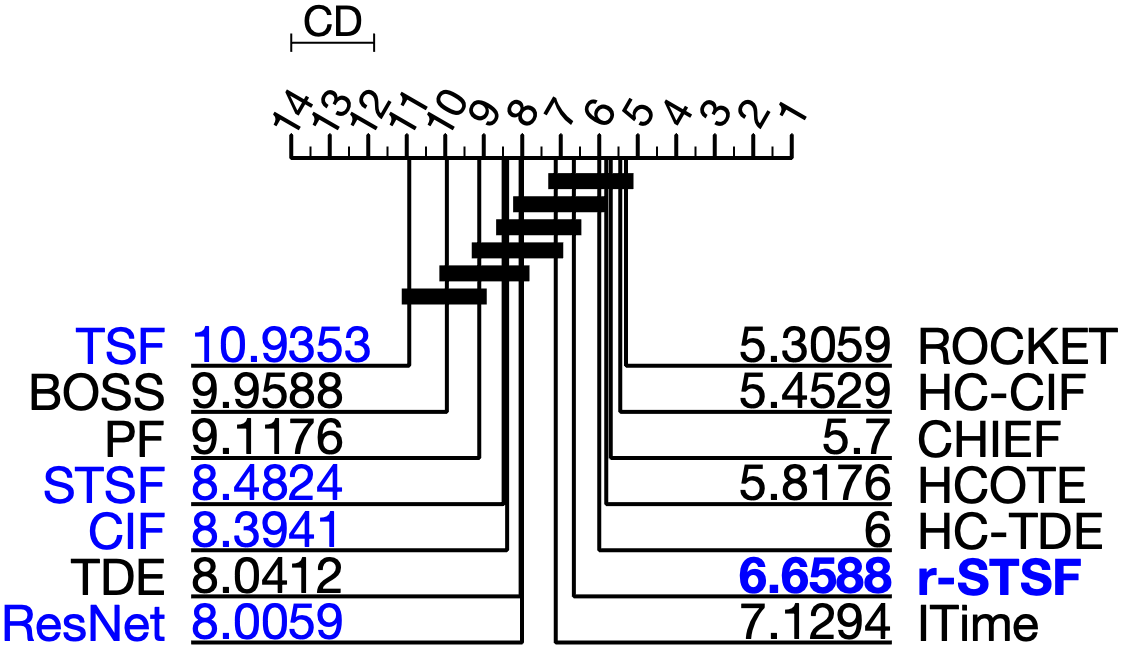}
}\hspace{5mm}
\subfloat[\label{fig:sotaclassifiers_85dsets_waa}]{
	\includegraphics[scale=0.07]{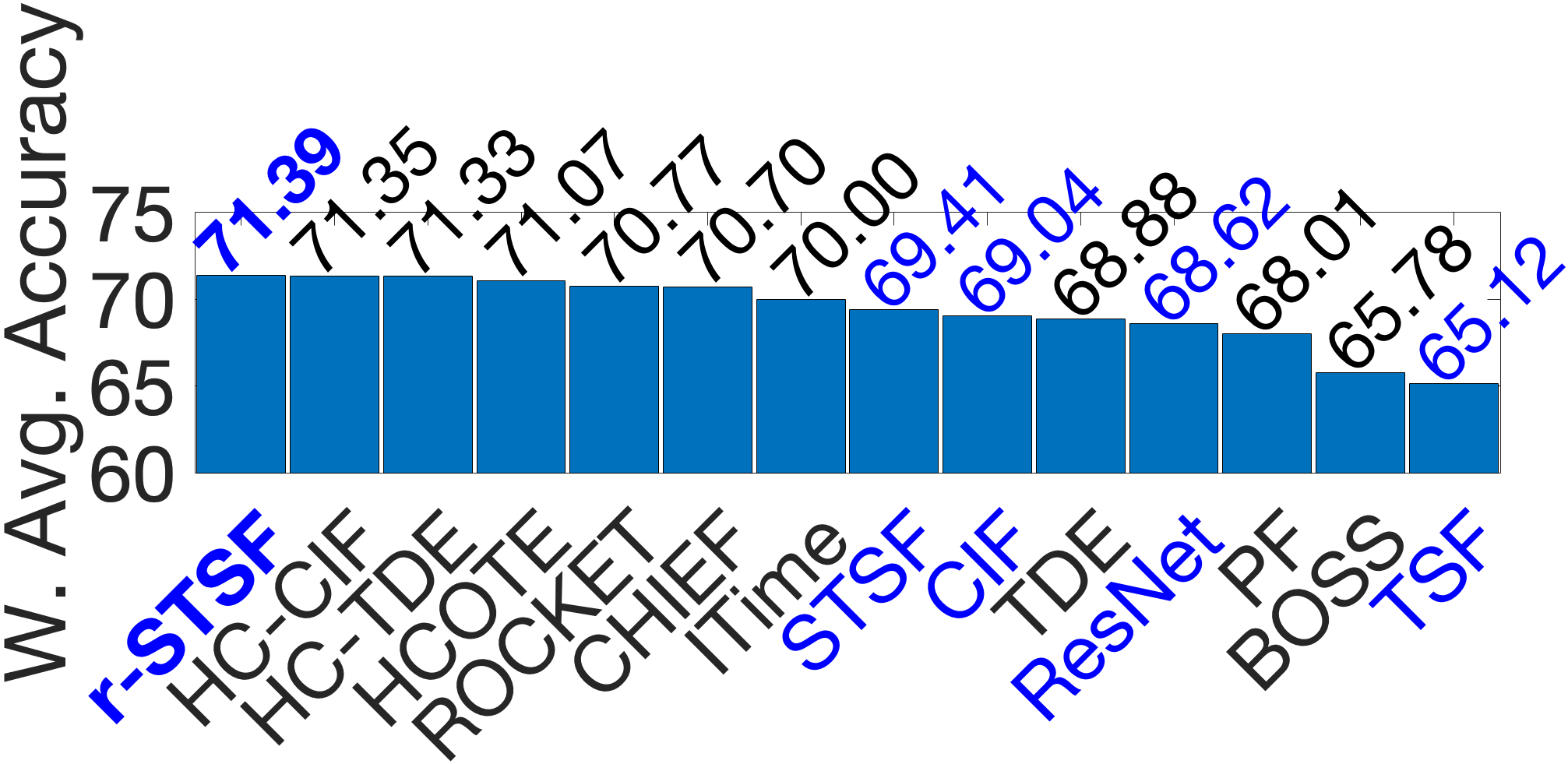}
}
\caption{\textcolor{black}{\textbf{(a)} Critical difference diagram of average ranks, and \textbf{(b)} Weighted average accuracy of r-STSF and competitors on $UCR_{85}$. Interpretable TSC methods are in blue. Non-interpretable ones in black. r-STSF is the only interpretable classifier in the group of SOTA TSC methods and achieves the highest weighted average accuracy}}
\label{fig:ar_waa_sotaclassifiers_85dsets}
\end{figure}

\subsubsection{Comparing with SOTA and commonly used TSC Methods}
\label{subsec:sota-comparison}

r-STSF is the only interpretable TSC method that achieves competitive classification accuracy comparable to SOTA methods. \textcolor{black}{As \figref{fig:sotaclassifiers_85dsets_ar} shows, despite having a larger average rank than SOTA TSC methods,} r-STSF is \emph{not} significantly different from them. More importantly, r-STSF achieves the highest weighted average accuracy (\figref{fig:sotaclassifiers_85dsets_waa}) which shows its robustness to classify more difficult datasets. r-STSF outperforms STSF according to both evaluation metrics. This validates the importance of our substantial new contributions to achieve higher classification accuracies.
\color{black}

\subsubsection{Comparing TSC Methods with Data from Different Application Domains}
\label{subsec:TSdomain-comparison}
\begin{figure}[htb]
    \centering
    \includegraphics[scale=0.065]{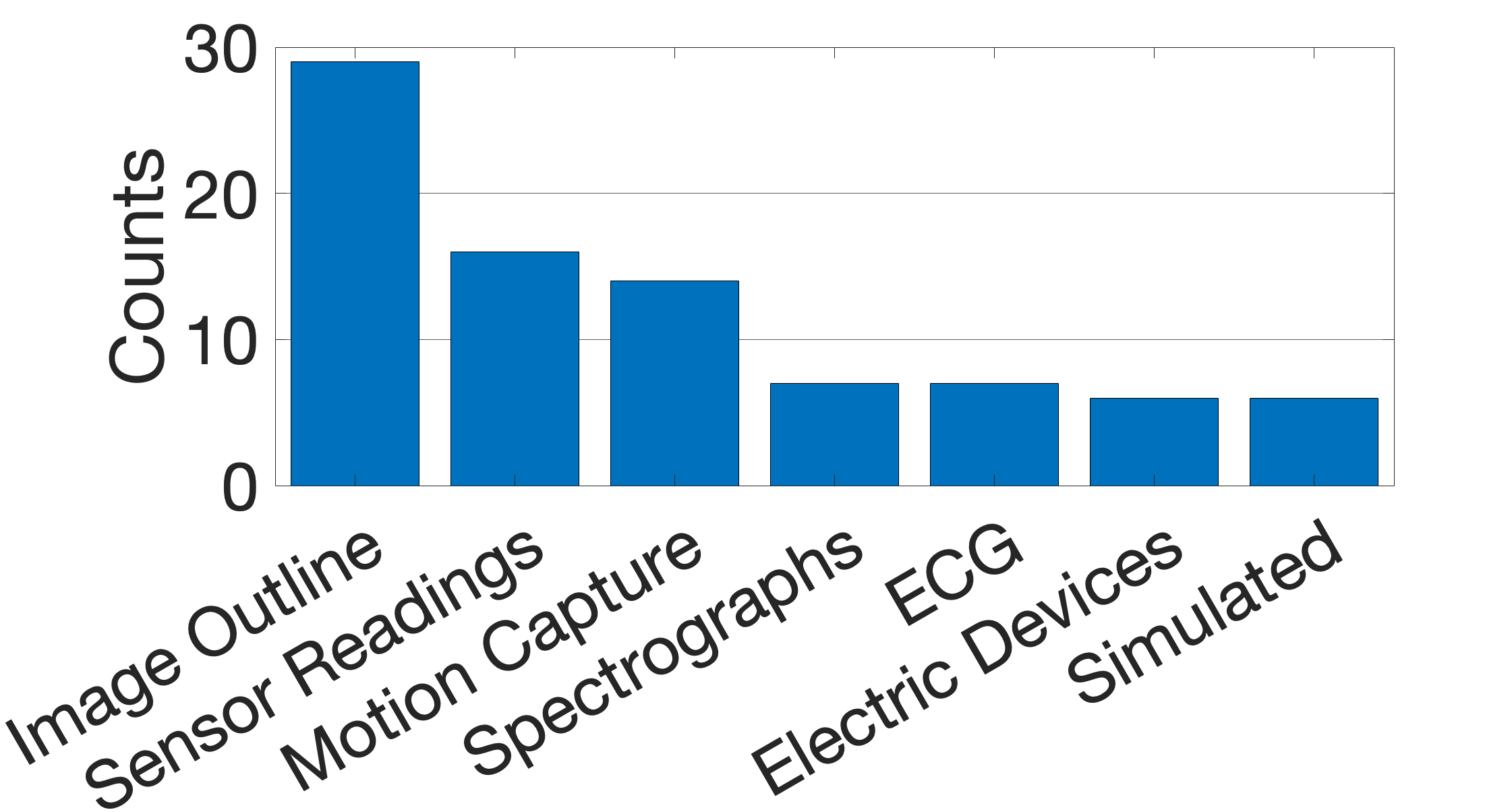}
    \caption{\textcolor{black}{Summary of dataset domains of $UCR_{85}$. There is an over representation of data from image outline classification problems}
    } 
    \label{fig:85dsets_types_summary}
\end{figure}

It has been a common practice in the TSC community to use 
$UCR_{85}$ for empirical studies.  
These datasets come from a variety of domains, with an emphasis on time series from the \emph{image outline} domain, as shown in~\figref{fig:85dsets_types_summary}. 

\begin{figure}[]
\centering
\subfloat[\label{fig:85dsets_image} Image Outline]{
	\includegraphics[scale=0.125]{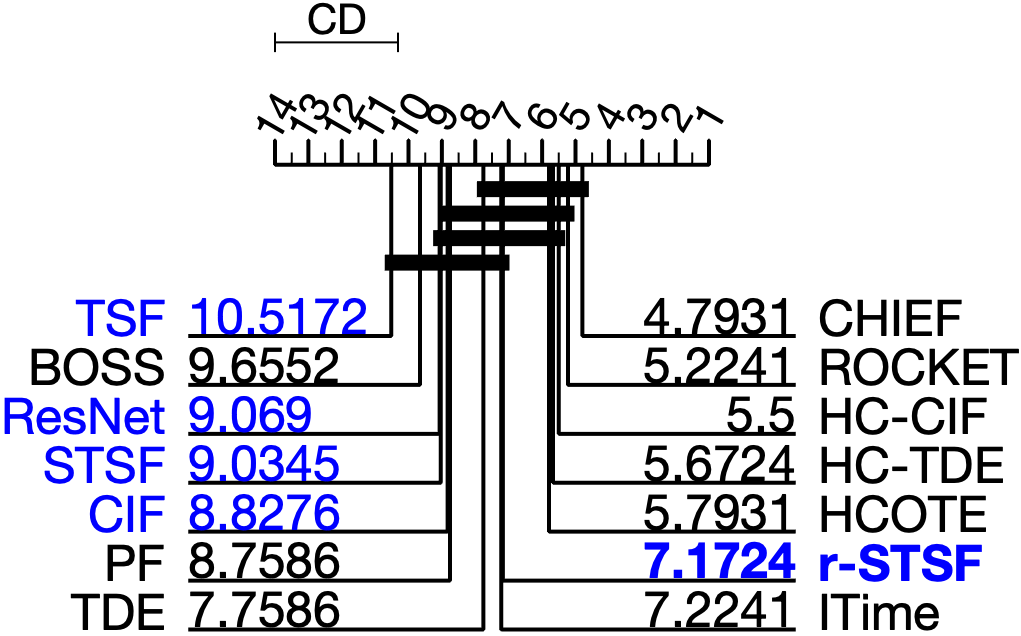}
}\hspace{10mm}
\subfloat[\label{fig:85dsets_sensor} Sensor Readings]{
	\includegraphics[scale=0.125]{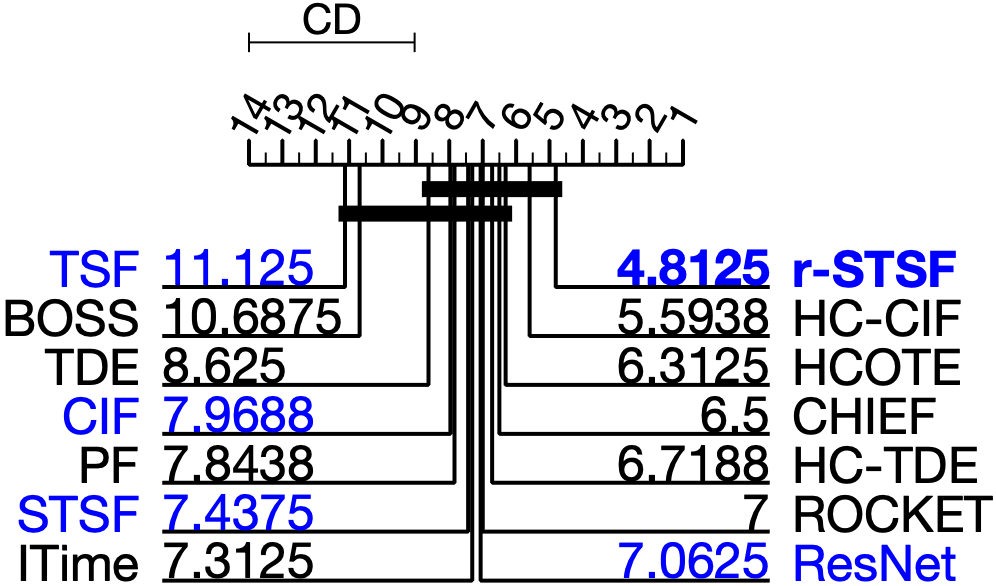}
}\\
\subfloat[\label{fig:85dsets_motion} Motion Capture]{
	\includegraphics[scale=0.125]{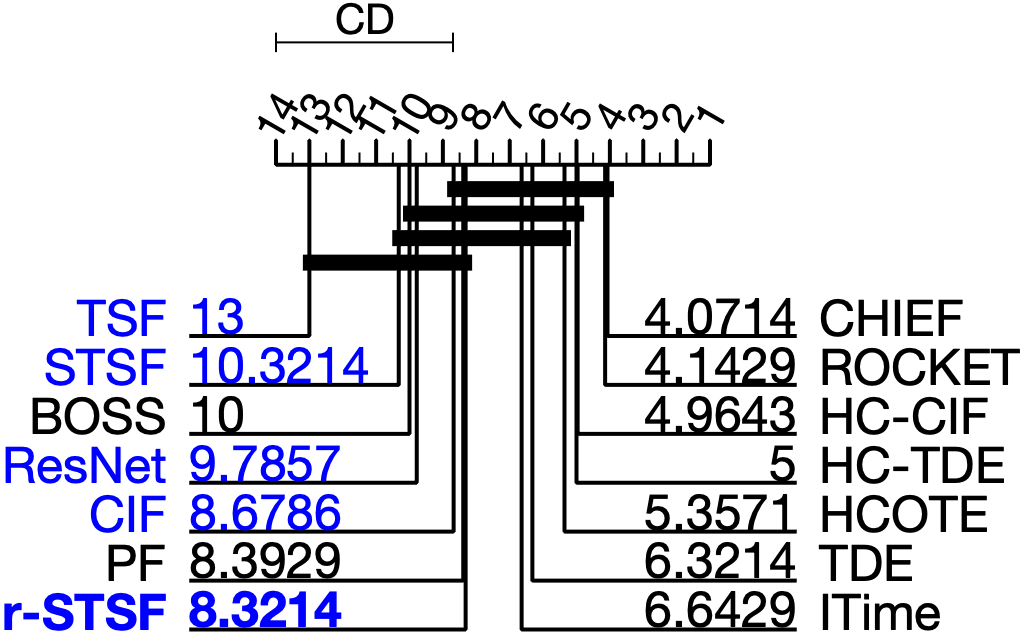}
}\hspace{10mm}
\subfloat[\label{fig:85dsets_spectro} Spectrographs]{
	\includegraphics[scale=0.125]{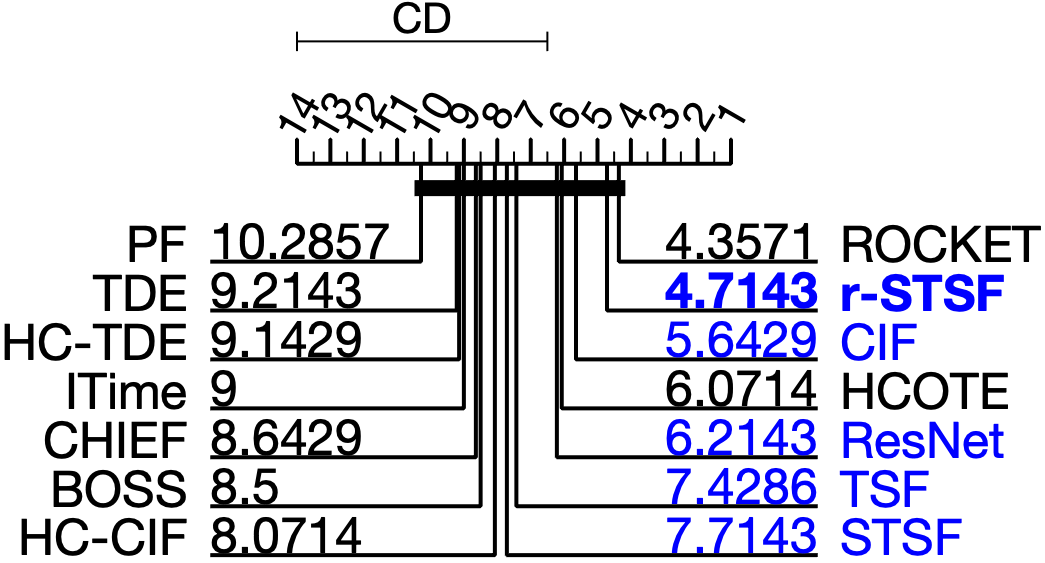}
}\\
\subfloat[\label{fig:85dsets_ECG} ECG]{
	\includegraphics[scale=0.125]{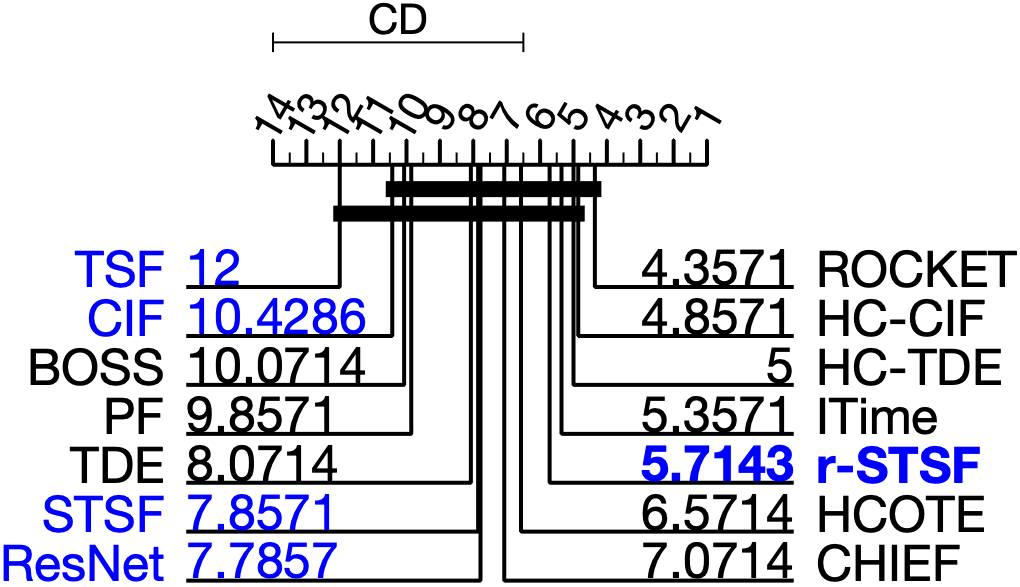}
}\hspace{10mm}
\subfloat[\label{fig:85dsets_devices} Electric Devices]{
	\includegraphics[scale=0.125]{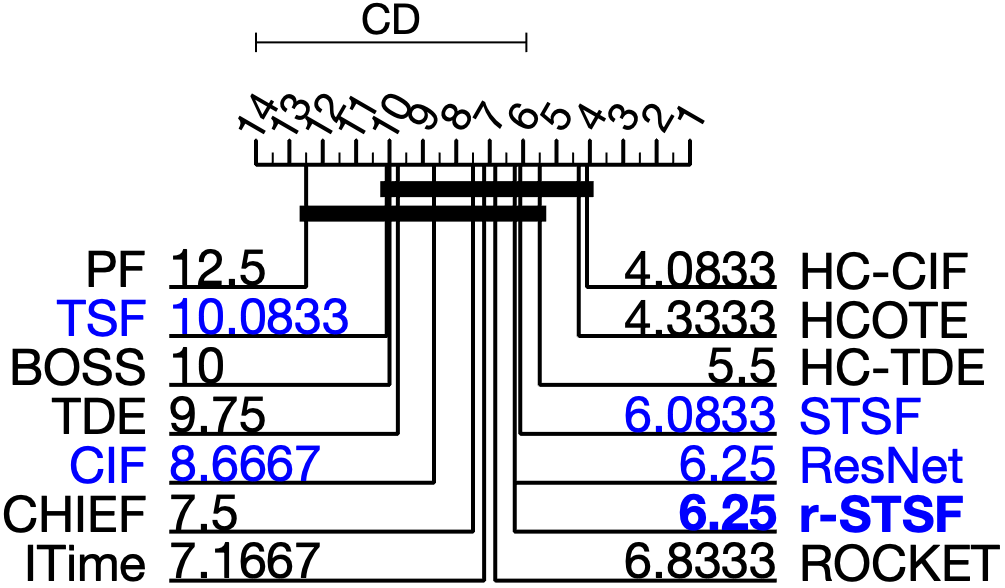}
}\\
\subfloat[\label{fig:85dsets_simulated} Simulated]{
	\includegraphics[scale=0.125]{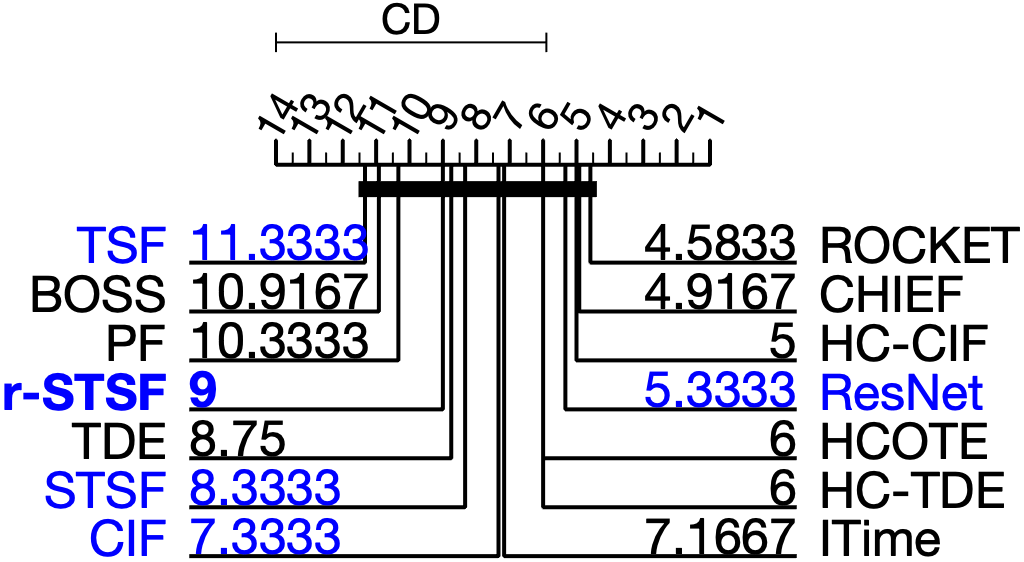}
}
\caption{\textcolor{black}{Critical difference diagram of average ranks per time series domain in $UCR_{85}$.
\textbf{(a)} Image Outline, \textbf{(b)} Sensor Readings, \textbf{(c)} Motion Capture, \textbf{(d)} Spectrographs, \textbf{(e)} ECG, \textbf{(f)} Electric Devices, \textbf{(g)} Simulated. Interpretable TSC methods are in color blue. Non-interpretable ones in color black}
}
\label{fig:85dsets_allDomains}
\end{figure}

We argue that using an unbalanced distribution of times series domains may not be appropriate to assess the overall effectiveness of different TSC methods. As~\figref{fig:85dsets_allDomains} shows,  the average ranks of different TSC method change when classifying datasets from different domains. The overall average ranks of the TSC methods on $UCR_{85}$ (\figref{fig:sotaclassifiers_85dsets_ar}) are highly influenced by  the classification accuracy in the over represented time series domain, i.e., \emph{image outline} (\figref{fig:85dsets_image}). The dominant or over represented domain could affect positively or negatively the \textcolor{black}{``perceived"} effectiveness of a classifier. For example, adding more time series datasets from \emph{sensor readings} or \emph{spectrograhps} domains could improve the average rank of r-STSF. Similarly, adding more datasets from \emph{ECG} or \emph{electric devices} domains could decrease the average performance of the SOTA method TS-CHIEF.

\begin{figure}[h]
\centering
\subfloat[\label{fig:waa_85dsets_image} Image Outline]{
	\includegraphics[scale=0.065]{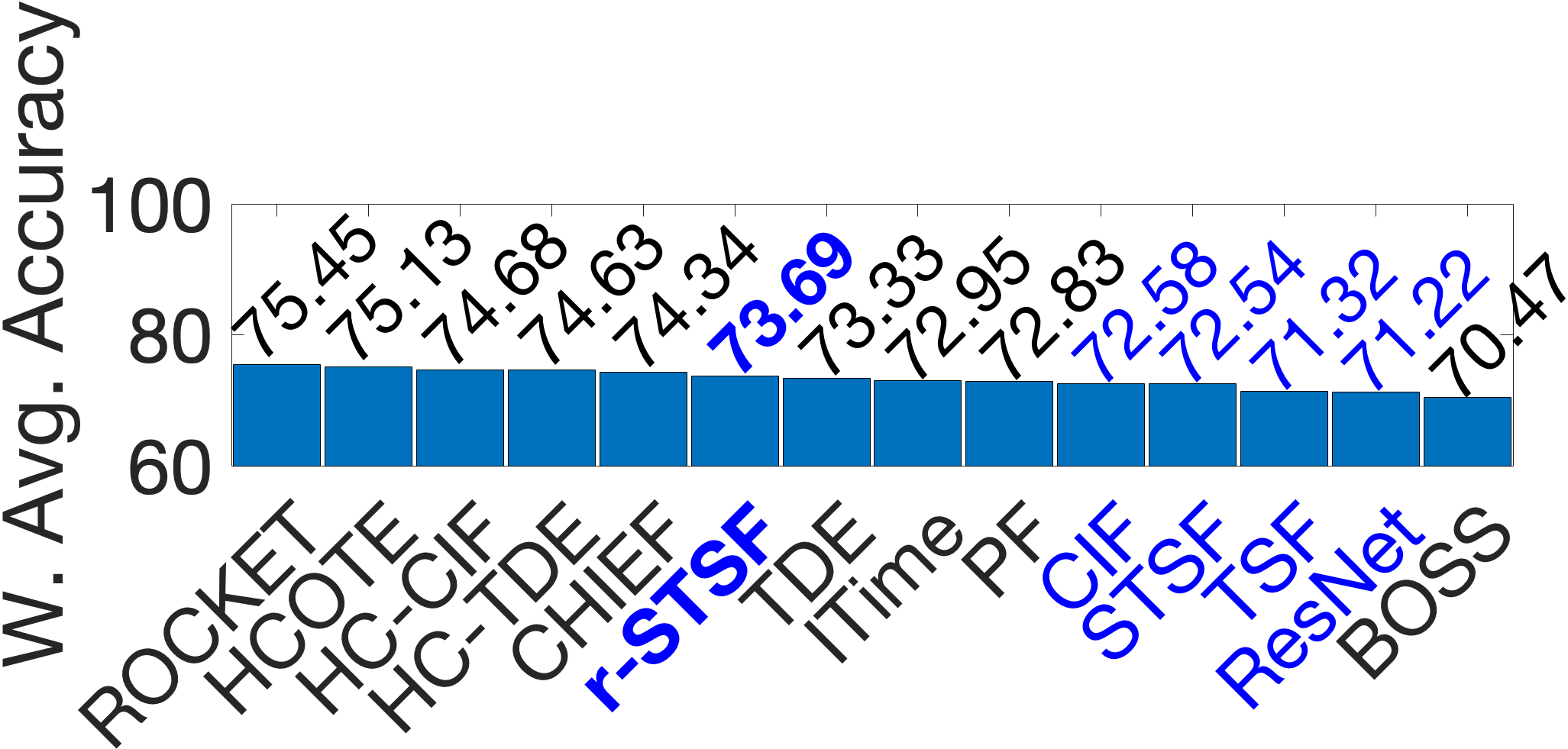}
}\hspace{5mm}
\subfloat[\label{fig:waa_85dsets_sensor} Sensor Readings]{
	\includegraphics[scale=0.065]{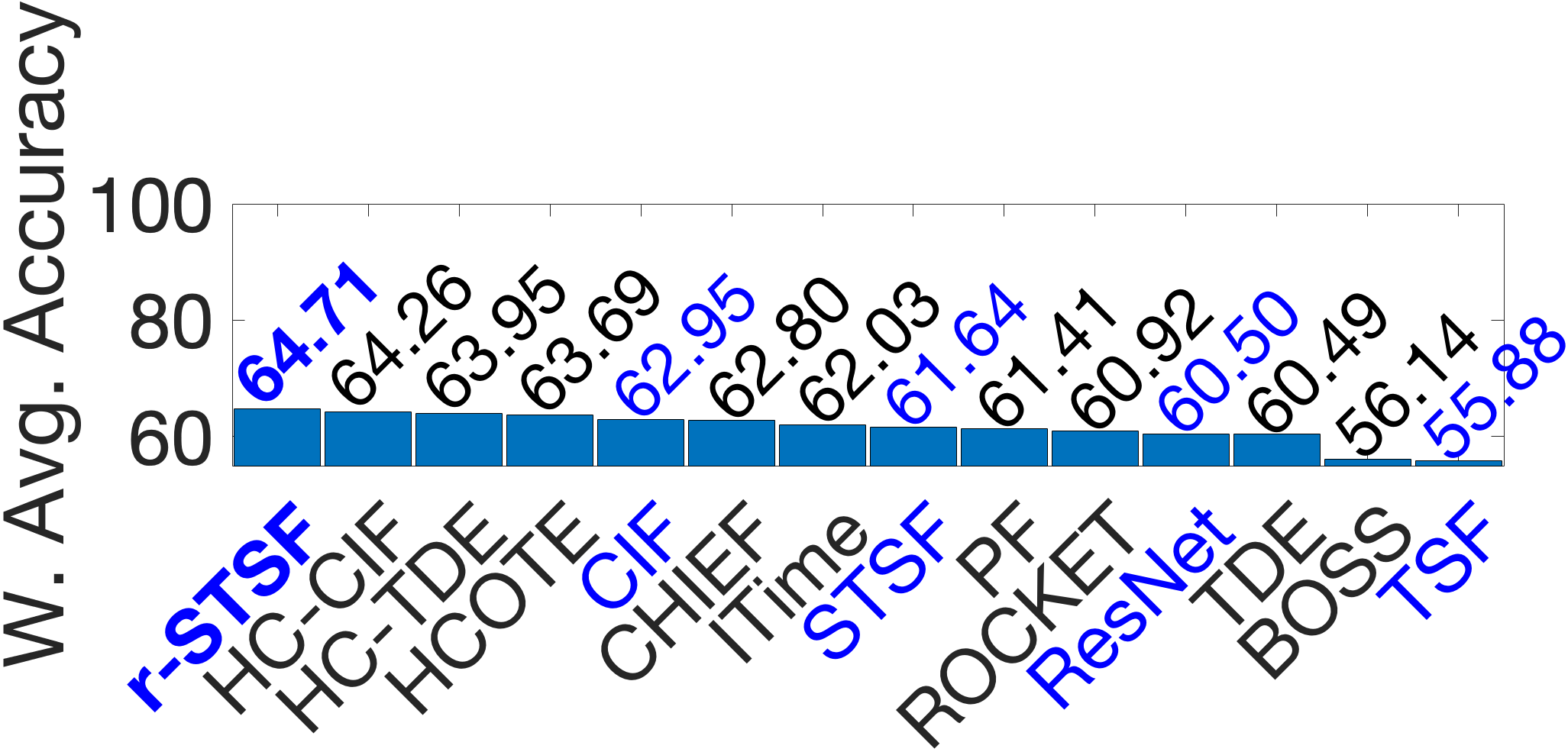}
}\\
\subfloat[\label{fig:waa_85dsets_motion} Motion Capture]{
	\includegraphics[scale=0.065]{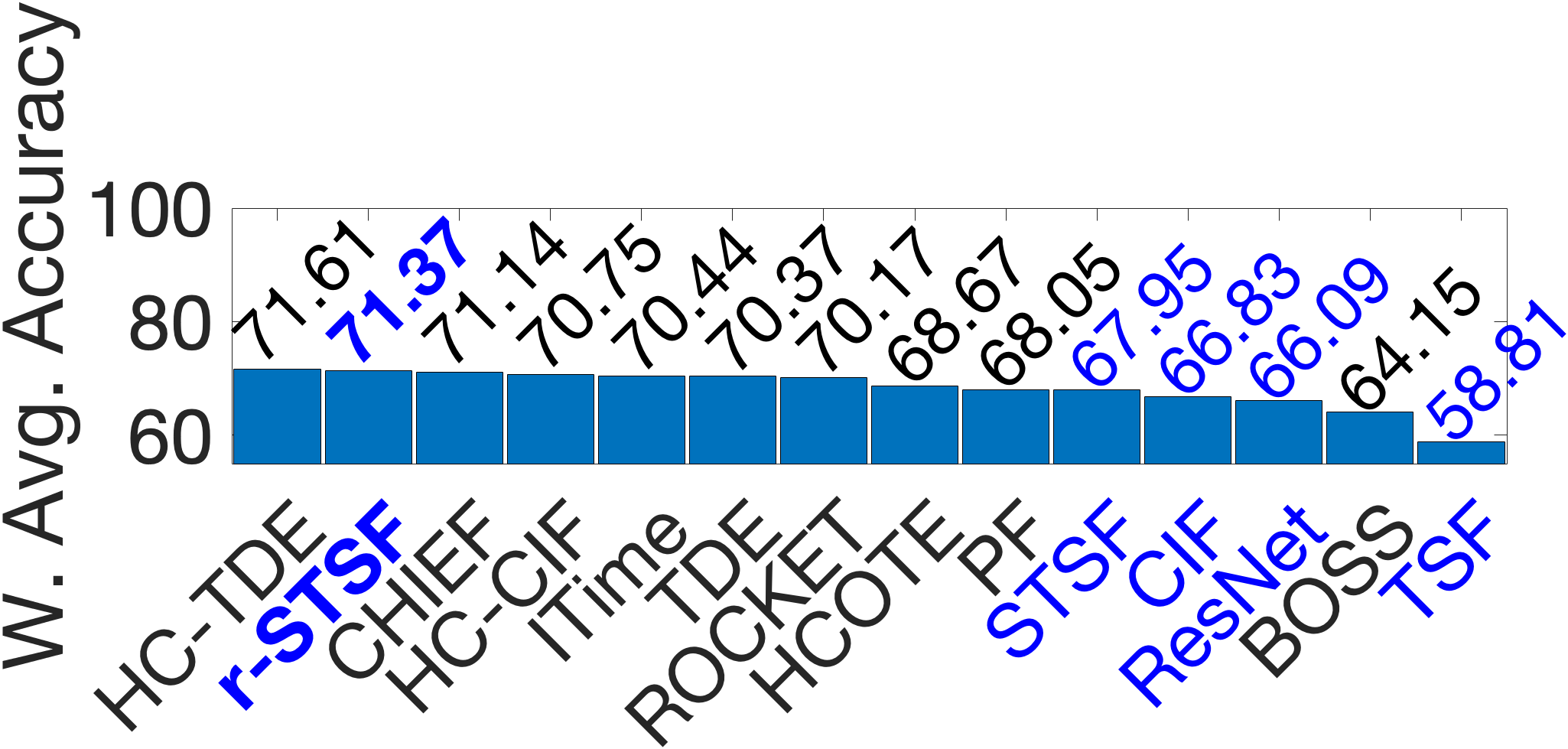}
}\hspace{5mm}
\subfloat[\label{fig:waa_85dsets_spectro} Spectrographs]{
	\includegraphics[scale=0.065]{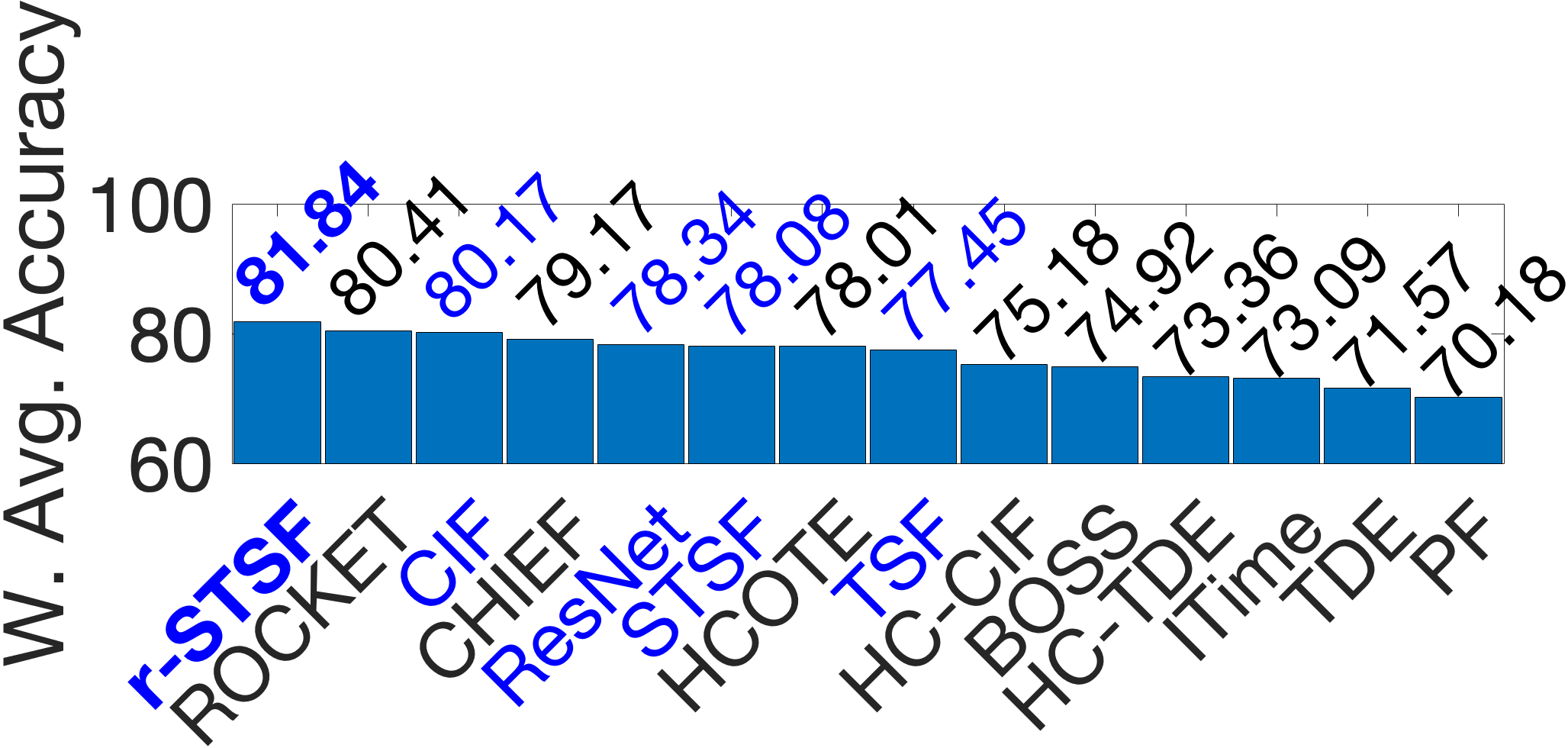}
}\\
\subfloat[\label{fig:waa_85dsets_ECG} ECG]{
	\includegraphics[scale=0.065]{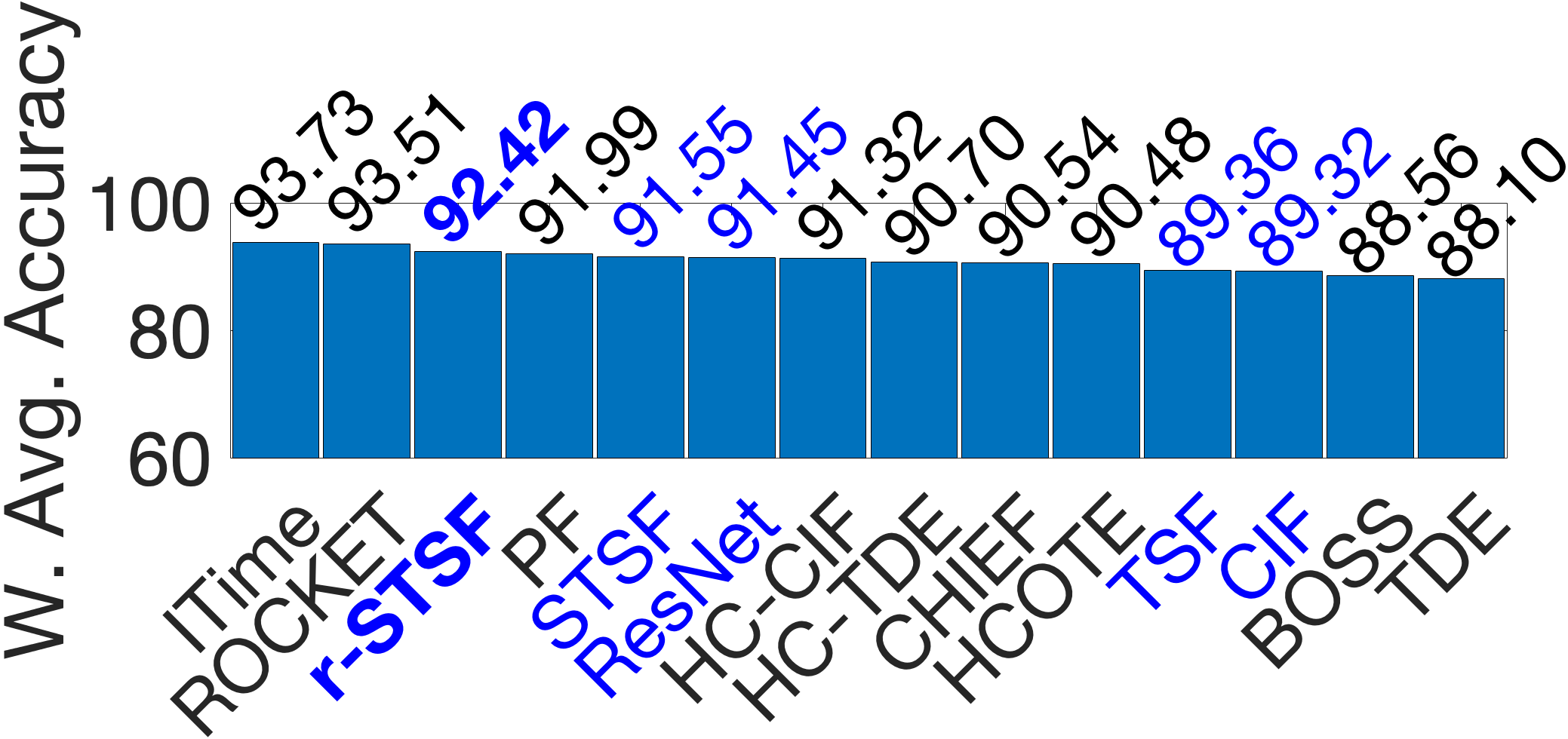}
}\hspace{5mm}
\subfloat[\label{fig:waa_85dsets_devices} Electric Devices]{
	\includegraphics[scale=0.065]{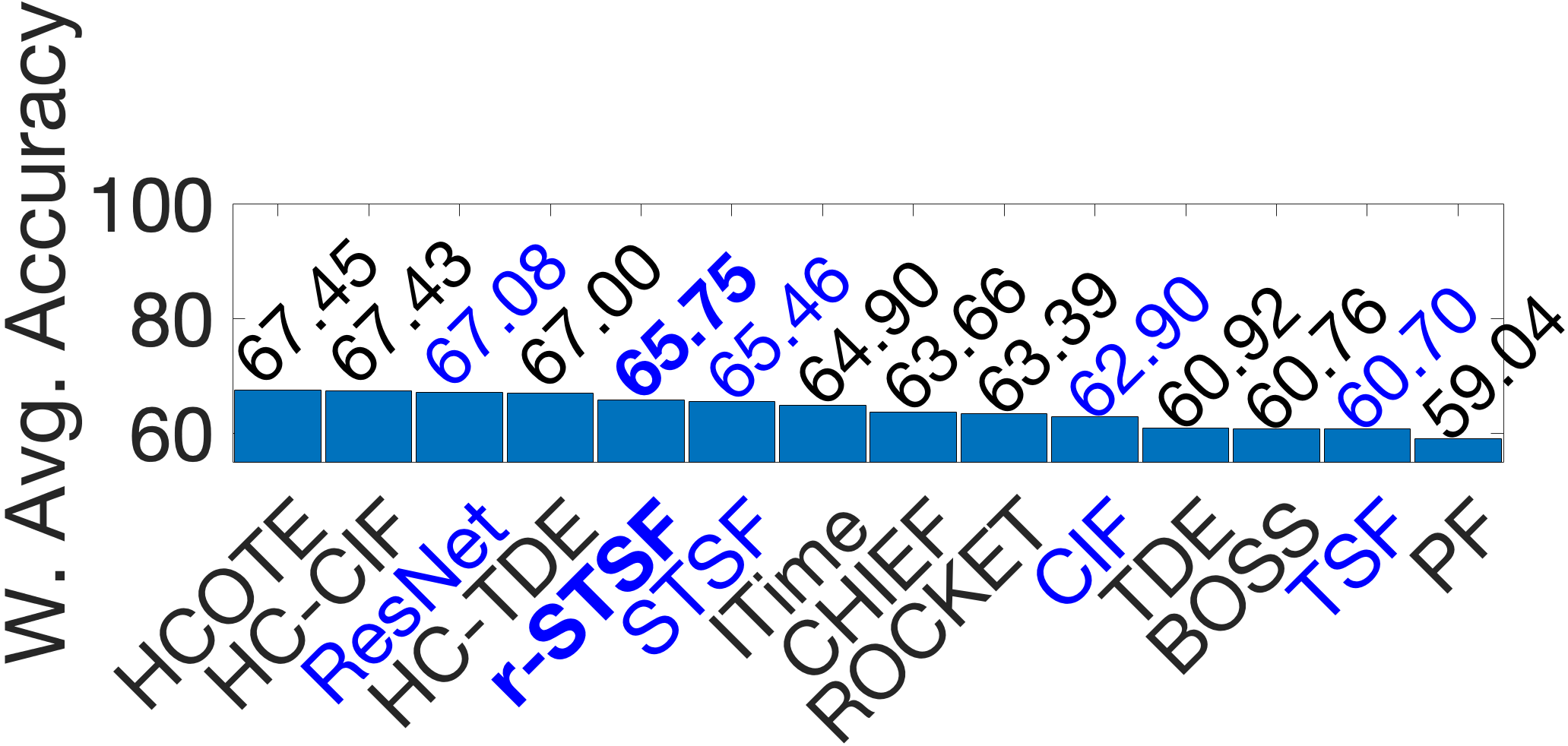}
}\\
\subfloat[\label{fig:waa_85dsets_simulated} Simulated]{
	\includegraphics[scale=0.065]{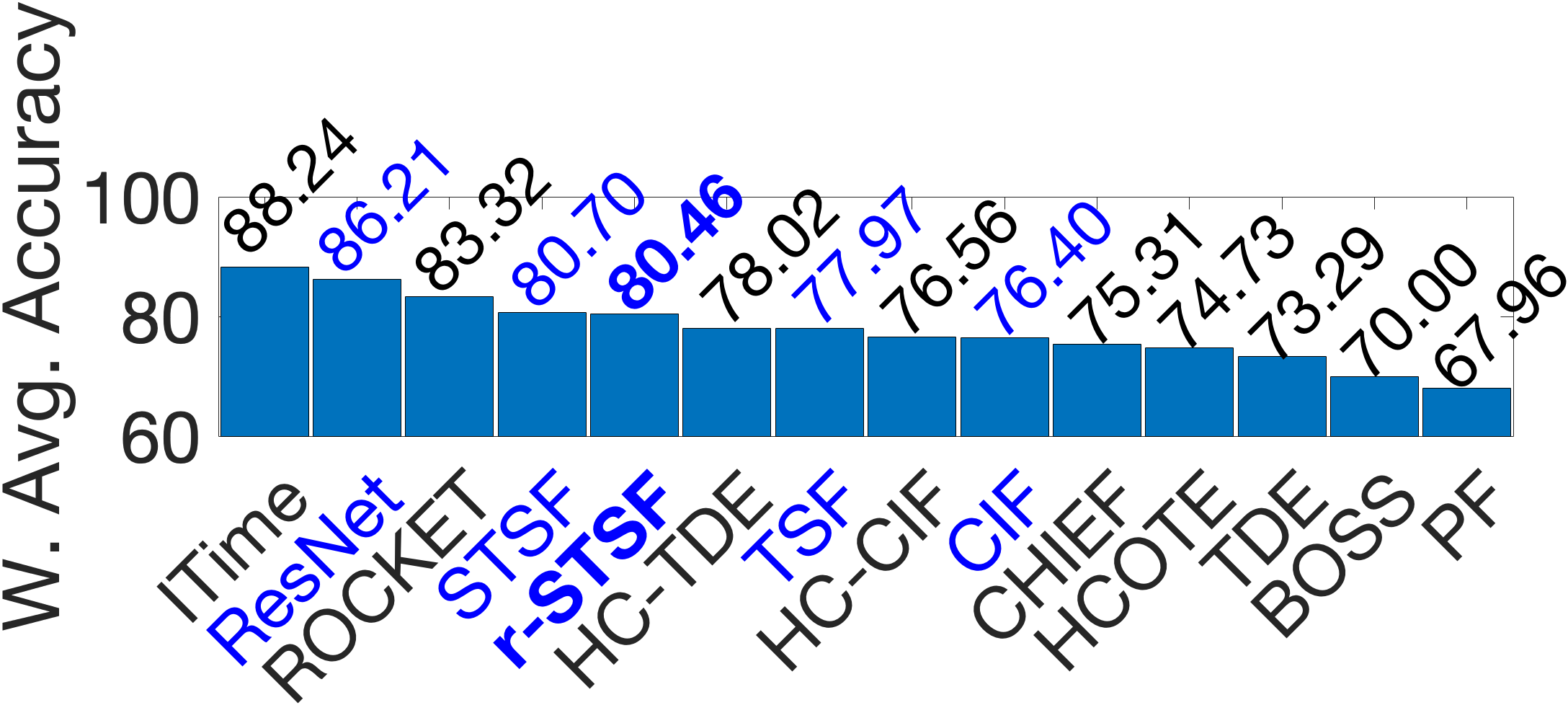}
}
\caption{\textcolor{black}{Weighted average accuracy per time series domain in $UCR_{85}$.
\textbf{(a)} Image Outline, \textbf{(b)} Sensor Readings, \textbf{(c)} Motion Capture, \textbf{(d)} Spectrographs, \textbf{(e)} ECG, \textbf{(f)} Electric Devices, \textbf{(g)} Simulated. Interpretable TSC methods are in color blue. Non-interpretable ones in color black}
}
\label{fig:waa_85dsets_allDomains}
\end{figure}

r-STSF is the best approach when classifying series from \emph{sensor readings} domain (see~\figref{fig:85dsets_sensor}), which is the second largest domain on $UCR_{85}$. The overall better performance of r-STSF on this domain is confirmed by the weighted average accuracy comparison (see~\figref{fig:waa_85dsets_sensor}). r-STSF, as well as other interval-based methods, focuses on \emph{phase-dependent} discriminatory intervals, i.e., discriminatory features located at the same time regions over all time series. 
The high accuracy of r-STSF when classifying \emph{sensor readings} series suggests that there are \textcolor{black}{relevant features embedded in phase-dependent intervals.} Nonetheless, these features cannot be found in the original (time-stamped) series but in additional representations of the time series. This is inferred from the rather poor performance (i.e., higher average ranks) of the interval-based methods TSF and CIF that only use the original series to extract discriminatory features. Further, the additional parameters and design criteria make r-STSF to outperform STSF on this type of domain.

In series from the \emph{spectrographs} domain, r-STSF also achieves a good performance. r-STSF is the second best classifier according to the average ranks (see~\figref{fig:85dsets_spectro}) and best classifier according to the weighted average accuracy (see~\figref{fig:waa_85dsets_spectro}). It is evident that on this type of domains there exists phase-dependent discriminatory features. This is the only domain where interval-based methods outperform most of the other SOTA classifiers. \textcolor{black}{Series derived from the \emph{spectrographs} domain show the spectrum of frequencies of a signal as it varies with time, i.e., show the time at which the signal's frequency changes. Thus, if series exhibit changes of their frequency at different times, interval-based TSC methods can detect those times and provide accurate classification accuracies.}

According to~\figref{fig:85dsets_allDomains} r-STSF usually ranks between the 1st and 6th place, with the exception of the \emph{simulated} and \emph{motion capture} domains where r-STSF's effectiveness decreases. The \emph{motion capture} datasets are prone to contain \emph{phase-independent} discriminatory intervals, which pose an additional challenge for r-STSF (and other interval-based methods). This reflects on the higher average rank of r-STSF when classifying \emph{motion capture} series (see~\figref{fig:85dsets_motion}). Nonetheless, despite the challenges for r-STSF when classifying this type of series, if datasets (from \emph{motion capture} domain) are complex (i.e., difficult-to-classify), r-STSF outperforms most of the TSC methods (see~\figref{fig:waa_85dsets_motion}). 

Among $UCR_{85}$, only six are representative of the \emph{simulated} domain, which are CBF, ChlorineConcentration (ChlCon), Mallat, ShapeletSim, SyntheticControl (SynthCon), and TwoPatterns. As shown in~\tableref{table:comparison}, Appendix~\ref{appendix:85dsets_appendix}, most of classifiers (including r-STSF) achieve accuracies above 95\% for the majority of the \emph{simulated} datasets. Consider the case of CBF dataset where the accuracy of most of the classifiers is above 99\%, and r-STSF ranks in 12th place (among the fourteen evaluated TSC methods, but 9th if considering the ten classifiers shown in~\tableref{table:comparison}, Appendix~\ref{appendix:85dsets_appendix}) despite having an accuracy of 99.17\%. In this kind of scenarios, the ranks amplify the differences among TSC methods and may not reflect \textcolor{black}{their actual} performance difference. Even in this extreme case, r-STSF is the fifth best TSC method according to the weighted average accuracy (\figref{fig:waa_85dsets_simulated}). 

It is worth to note that r-STSF outperforms STSF in the majority of domains. Only in datasets from \emph{electric devices} (see~\figref{fig:85dsets_devices} and \figref{fig:waa_85dsets_devices}) and \emph{simulated} (~\figref{fig:85dsets_simulated} and \figref{fig:waa_85dsets_simulated}) domains, STSF performs slightly better or is really close to the accuracy of r-STSF. This suggests that on these types of datasets/domains the number of relevant features is small. In such datasets randomized approaches are less likely to identify the relevant features~\citep{geurts2006extremely}. \textcolor{black}{An apriori estimation on the percentage of relevant (or irrelevant) features could be implemented to make r-STSF adaptive. Thus, when a dataset is expected to have a high percentage of relevant features r-STSF trains an ensemble of randomized binary trees for classification. Otherwise, r-STSF uses non-randomized binary trees for classification.}

\subsection{Running Time Evaluation}
\label{subsec:running_time}

\textbf{Training time:} We measure the training times of TSF, STSF, ResNet, BOSS, PF, TS-CHIEF (CHIEF), HIVE-COTE (HCOTE), InceptionTime (ITime), ROCKET, and r-STSF over 45 datasets (underlined in~\tableref{table:comparison}, Appendix~\ref{appendix:85dsets_appendix}). CIF, TDE, and the new HIVE-COTE variants HC-CIF and HC-TDE, were not included on this experiment. In the case of CIF, twenty-two features are computed from every extracted interval, making this approach more expensive than TSF which only computes three features for every interval (confirmed experimentally by the CIF paper). TDE is a memory intensive TSC method (it uses three times more memory than BOSS as stated by TDE's authors). We ran out of memory when classifying datasets such as StarLightCurves~\citep{UCRArchive} that has 1000 training instances with series of length 1024. HC-CIF and HC-TDE are more expensive in time and memory, respectively, than HIVE-COTE.

\begin{figure}[h]
\centering

\subfloat[\label{fig:SOTA-trntimes-comparison_ar}]{
	\includegraphics[scale=0.045]{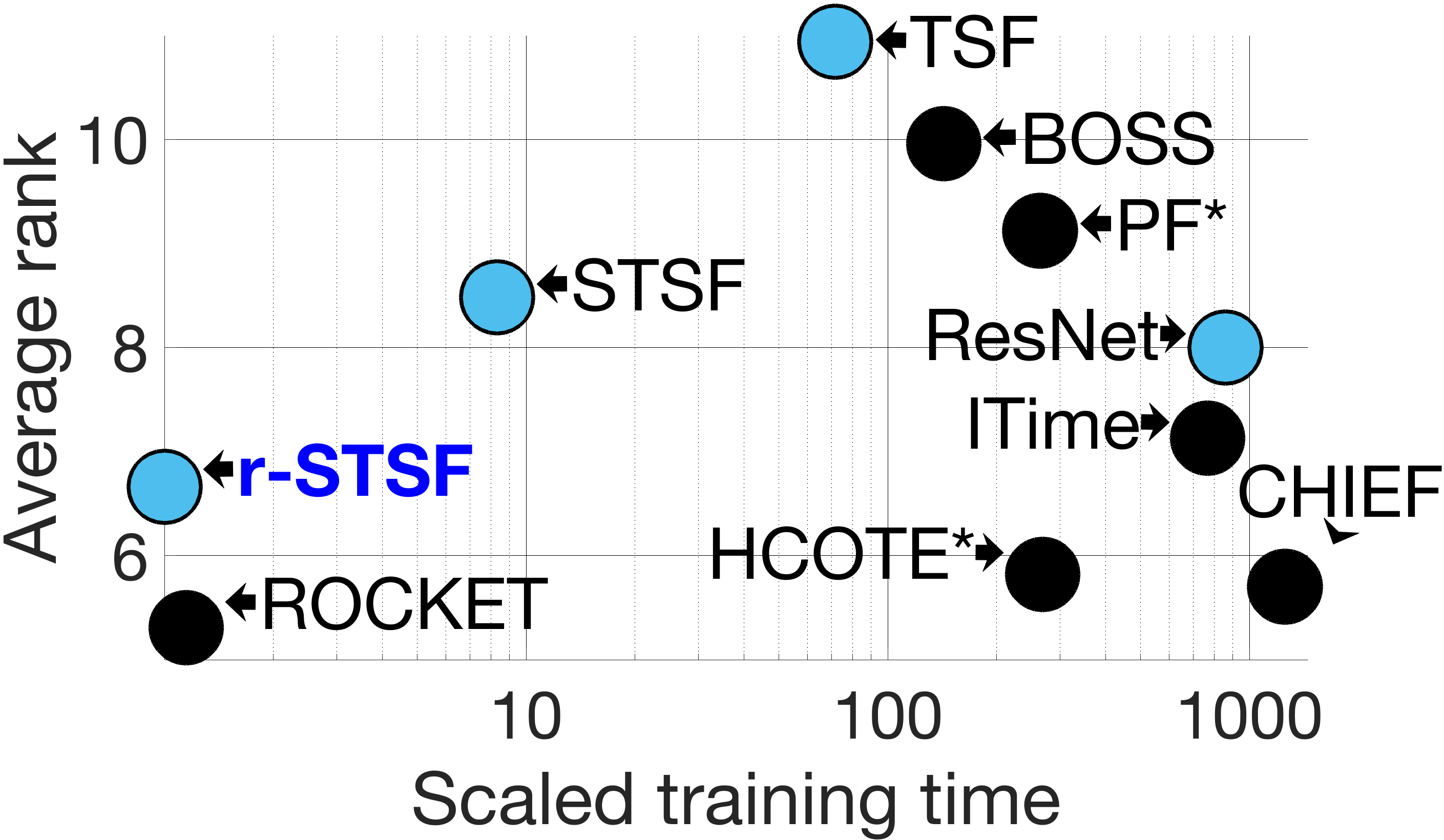}
}\hspace{5mm}
\subfloat[\label{fig:SOTA-trntimes-comparison_waa}]{
	\includegraphics[scale=0.045]{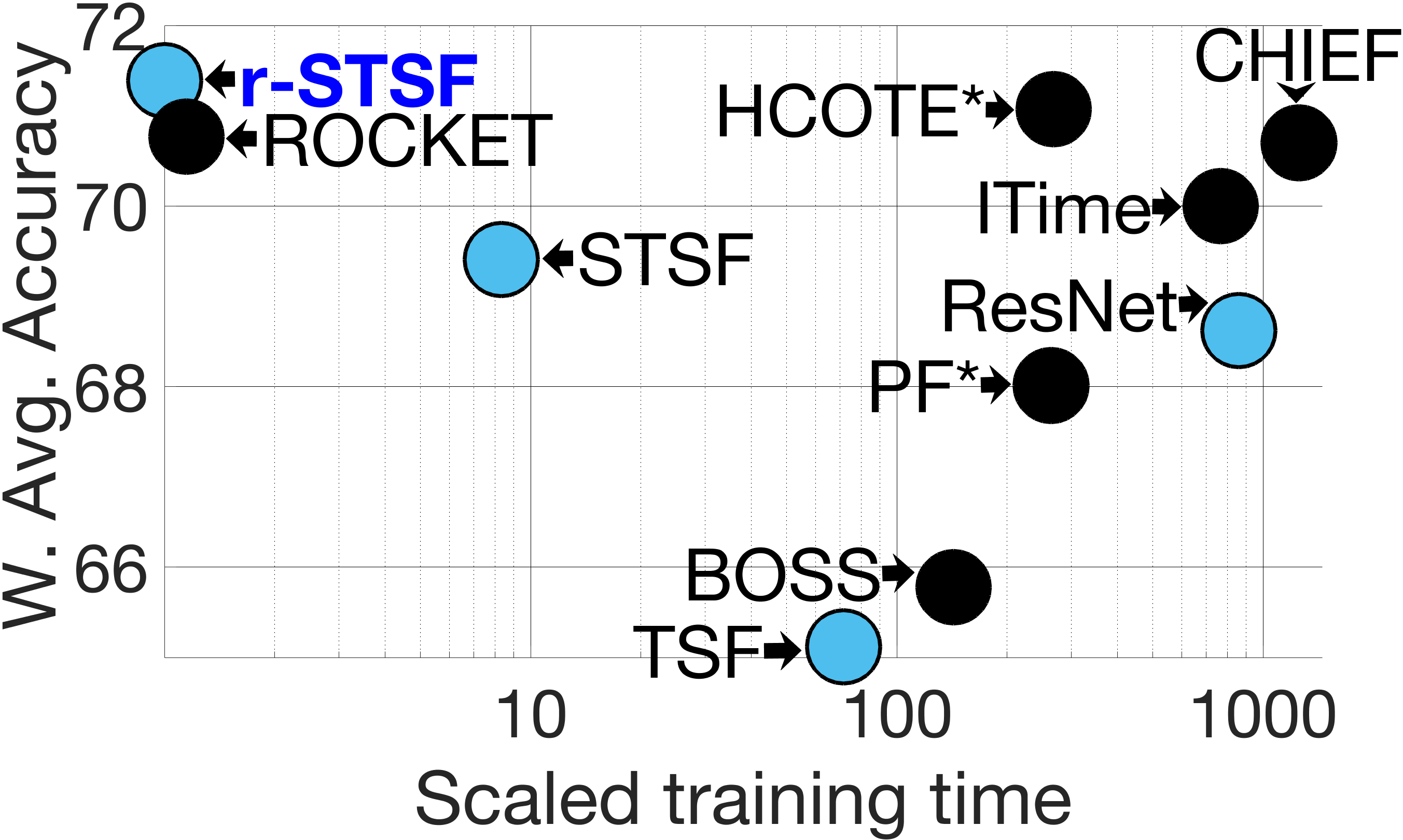}
}
\caption{\textcolor{black}{\textbf{(a)} Average ranks and \textbf{(b)} Weighted average accuracy vs. scaled training times of 
STSF, TSF, BOSS, ResNet, HIVE-COTE, PF, TS-CHIEF, InceptionTime, ROCKET, and r-STSF when training on $UCR_{45}$. All training times are scaled by the training time of r-STSF,  i.e., training time of r-STSF = 1. Interpretable TSC methods are colored in blue. Non-interpretable methods are colored in black. r-STSF is fast, scalable, and highly accurate, and it allows for interpretable classifications. According to the weighted average accuracy, r-STSF is the fastest at training and the most effective TSC method. * For methods such as HCOTE and PF their running time is expected to be significantly larger as detailed on this section}
}
\label{fig:SOTA-trntimes-comparison}
\end{figure}

We used the original 45 datasets of the UCR repository, $UCR_{45}$ (instead of $UCR_{85}$) due to the expensive computational time from the majority of the competitors. These TSC methods are implemented in different programming languages. TSF and STSF are implemented in Matlab; HIVE-COTE, PF, TS-CHIEF, and BOSS are implemented in Java; and ResNet, InceptionTime, ROCKET, and r-STSF are implemented in Python. While comparing running times of TSC methods implemented in different programming languages may seem biased, r-STSF is implemented in Python which is - as an interpreted language - slower than Java (used for the majority of baselines), and similar to Matlab (used for TSF and STSF). Parameters of each competitor are set as suggested in their original paper.
In~\figref{fig:SOTA-trntimes-comparison}, we present the average ranks (\figref{fig:SOTA-trntimes-comparison_ar}) and weighted average accuracy (\figref{fig:SOTA-trntimes-comparison_waa}) of each TSC method and its scaled training times. r-STSF is three orders of magnitude faster than TS-CHIEF, and two orders of magnitude faster than the majority of TSC methods, except for TSF, STSF, and ROCKET. r-STSF is an order of magnitude faster than TSF and STSF and has a similar training time to that of ROCKET.

PF uses only 100 trees instead of 500 as done in TSF, STSF, TS-CHIEF, and r-STSF. Since PF's training times scales linearly with the number of trees, using 500 trees will make PF's time five times larger than that shown in~\figref{fig:SOTA-trntimes-comparison}. However, PF's average rank is unlikely to change. As the PF paper suggests: ``It is unlikely that more trees would provide a very significant improvement, because the ratio of error-rates between 100 and 50 (trees) is already close to 1 (i.e., the errors are only slightly reduced)".
Due the expensive training times of HIVE-COTE, which is known to be very slow~\citep{shifaz2020ts,lucas2019proximity}, we only report the training times of HIVE-COTE when classifying 21 (out of 45) small to medium datasets. The results suggest that r-STSF is two orders of magnitude faster than HIVE-COTE. The addition of large datasets (i.e., with more than 1,000 instances) such as NonInvFetECGTho1, NonInvFetECGTho2, and StarLightCurves is expected to increase significantly the difference between r-STSF and HIVE-COTE. From the SOTA group, without considering HIVE-COTE (i.e., ROCKET, TS-CHIEF, r-STSF and InceptionTime), TS-CHIEF is the most expensive TSC method.  InceptionTime is highly accurate but two orders of magnitude slower than ROCKET and r-STSF, which are the fastest approaches. According to the weighted average accuracy, r-STSF is the fastest at training and the most effective TSC method (\figref{fig:SOTA-trntimes-comparison_waa}).

\begin{figure}[h]
\centering

\subfloat[\label{fig:SOTA-tsttimes-comparison_ar}]{
	\includegraphics[scale=0.045]{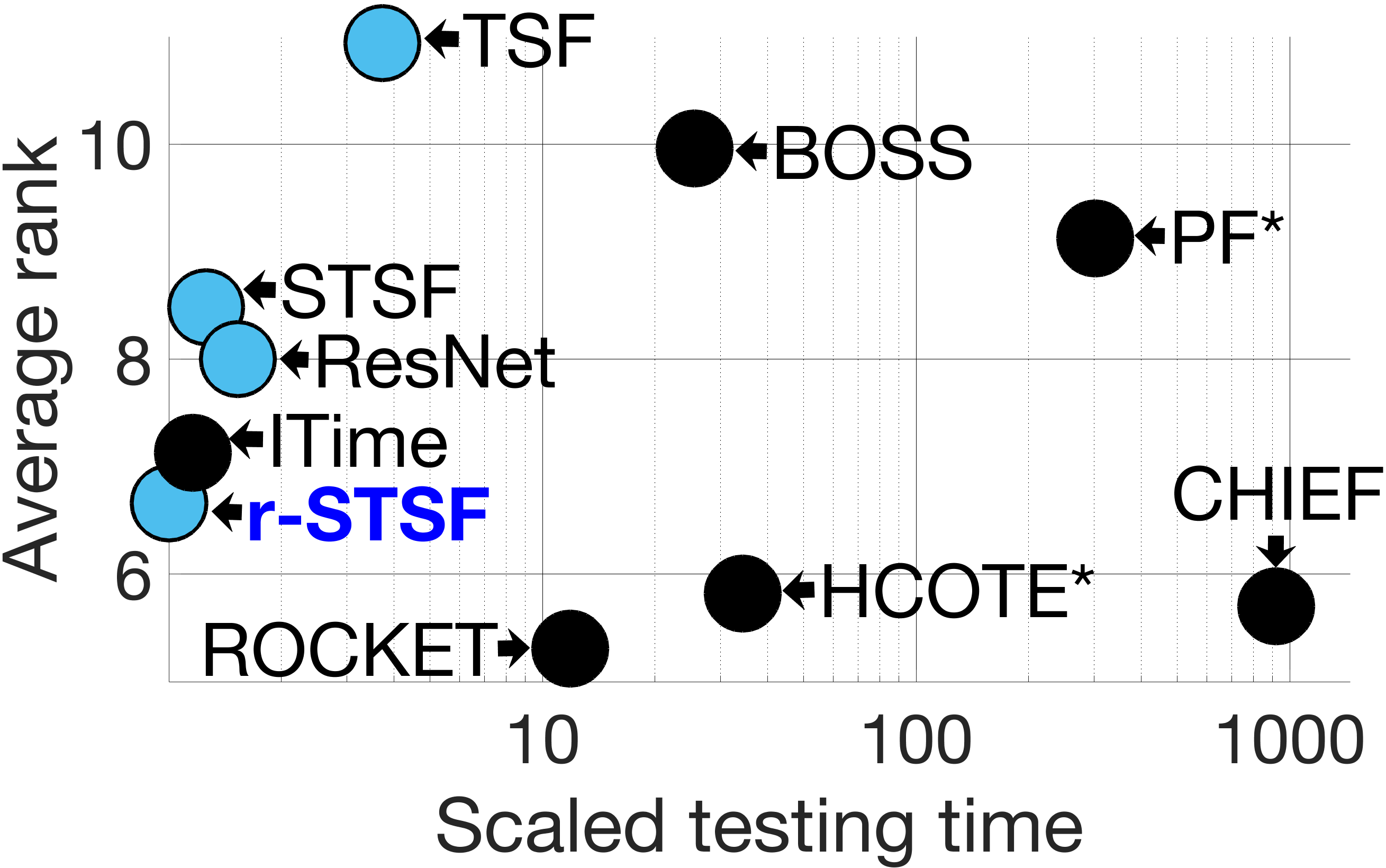}
}\hspace{5mm}
\subfloat[\label{fig:SOTA-tsttimes-comparison_waa}]{
	\includegraphics[scale=0.045]{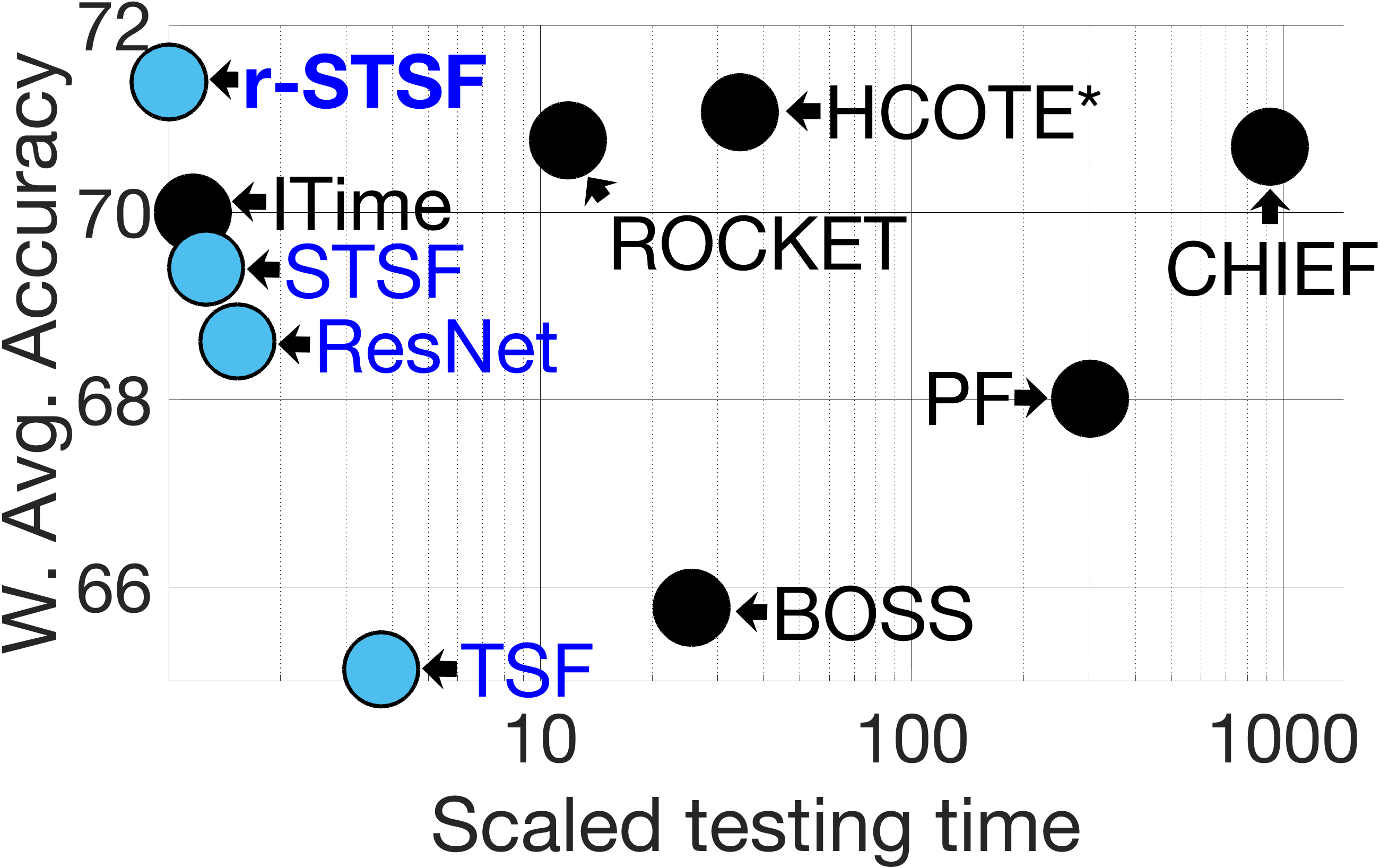}
}
\caption{\textcolor{black}{\textbf{(a)} Average ranks and \textbf{(b)} Weighted average accuracy vs. scaled testing times of STSF, TSF, BOSS, ResNet, HIVE-COTE, PF, TS-CHIEF, InceptionTime, ROCKET, and r-STSF when classifying on $UCR_{45}$. All testing times are scaled by the testing time of r-STSF,  i.e., testing time of r-STSF = 1. Interpretable TSC methods are colored in blue. Non-interpretable methods are colored in black. r-STSF is fast at testing and highly accurate. According to the weighted average accuracy, r-STSF is the fastest at testing and the most effective TSC method. * For methods such as HCOTE and PF their running time is expected to be significantly larger as detailed on this section}
}
\label{fig:SOTA-tsttimes-comparison}
\end{figure}

\noindent \textbf{Testing time:} In the majority of datasets from $UCR_{45}$, the number of testing instances is much larger than the number of training instances. This affects the performance of classifiers such as ROCKET, which is fast on training but may become slower at testing. ROCKET uses a high number of kernels (10,000) to transform the time series into a feature-based representation. The computation of each representation is related not only to the number of kernels but also to the characteristics of the datasets (i.e., number of instances and the length of the time series). For large datasets with very long series, ROCKET may become computationally expensive. This difference in the number of instances between training and testing sets is reflected in ROCKET's testing time, which is an order of magnitude slower than r-STSF and even than InceptionTime (\figref{fig:SOTA-tsttimes-comparison}).

The SOTA methods HIVE-COTE and TS-CHIEF are both expensive at testing. The testing time of HIVE-COTE on 21 out of 45 small to medium datasets is already an order of magnitude slower than r-STSF when tested on $UCR_{45}$. TS-CHIEF is two orders of magnitude slower than r-STSF. Similarly, PF is also two orders of magnitude slower than r-STSF at testing. These results suggest that from the SOTA group, r-STSF is the only approach which is fast at both training and testing. ROCKET is fast at training, but much slower at testing (specially if the testing datasets have a large number of instances). In contrast, InceptionTime is slow at training but much faster at testing. HIVE-COTE and TS-CHIEF are very expensive for both training and testing. According to the weighted average accuracy, r-STSF is the fastest at testing and the most effective TSC method (\figref{fig:SOTA-tsttimes-comparison_waa}).

\subsection{Generalizability on Additional Benchmark Datasets}
\label{subsec:43dsets}
\begin{figure}[h]
\centering
\subfloat[\label{fig:43dsets_ordered_avgRank_allClassifiers}]{
	\includegraphics[scale=0.14]{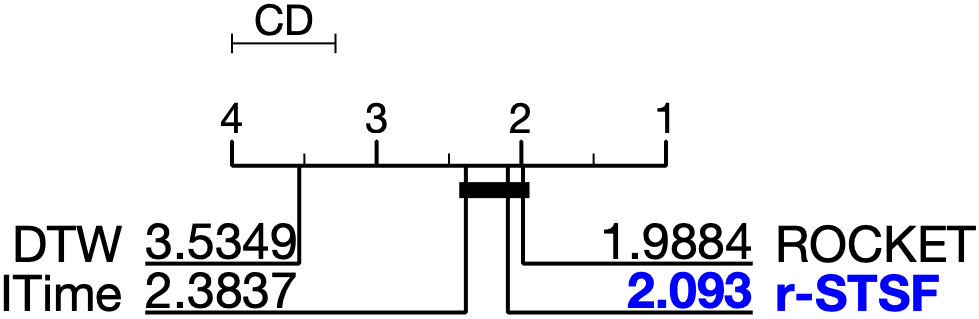}
}\hspace{10mm}
\subfloat[\label{fig:43dsets_ordered_WavgAccu_allClassifiers}]{
	\includegraphics[scale=0.055]{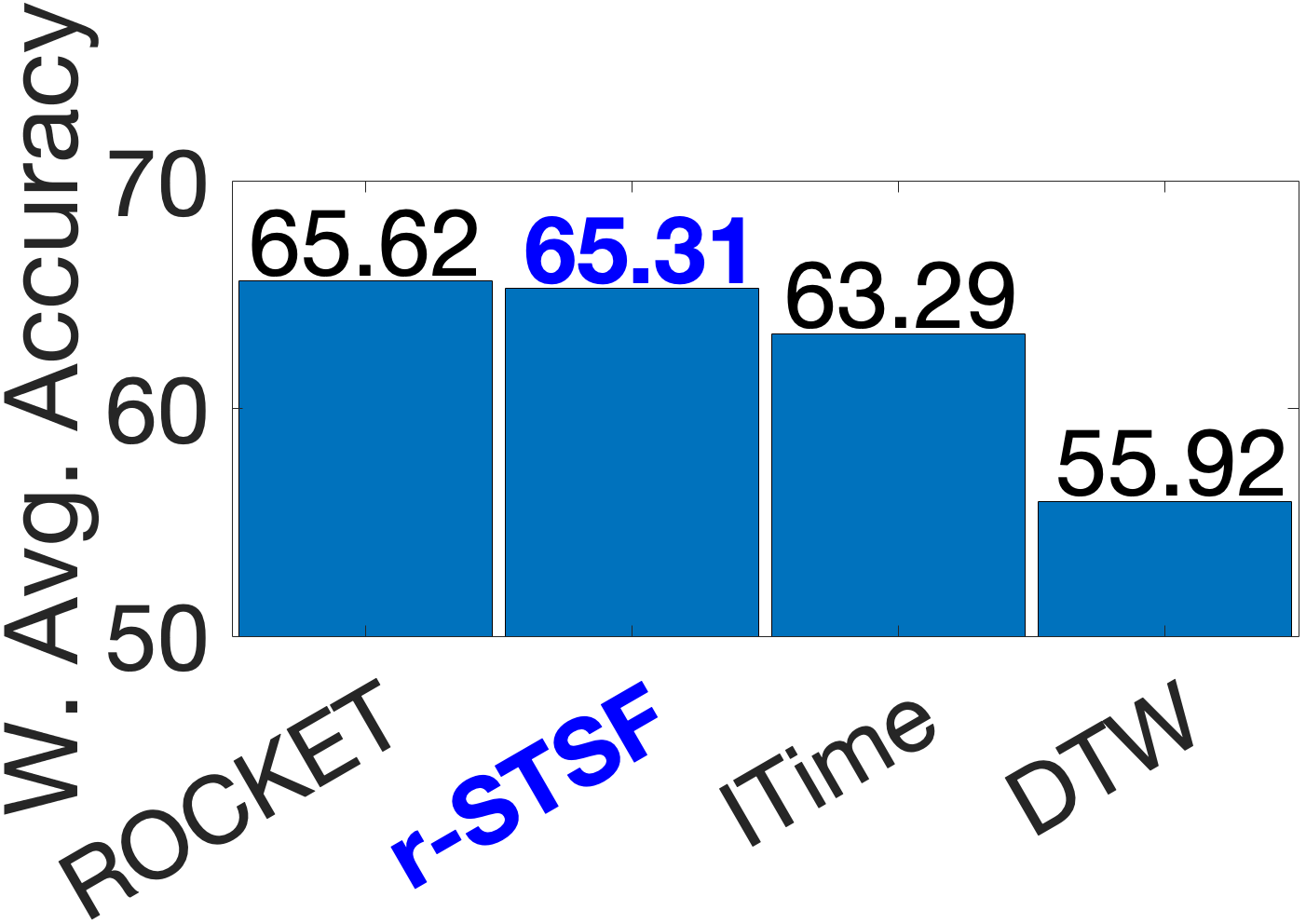}
}
\caption{\textcolor{black}{Comparison of TSC methods when assessed based on \textbf{(a)} average ranks and \textbf{(b)} weighted average accuracy. According to both metrics, r-STSF is the second best classifier when classifying on $UCR_{43}$}
}
\label{fig:43dsets_ordered_comparisonOfEvaluationMetrics}
\end{figure}

An additional set of 43 time series datasets was recently included in the UCR repository~\citep{UCRArchive}, $UCR_{43}$. We compare r-STSF with two very recent TSC methods: ROCKET and InceptionTime, and with a common baseline, 1NN-DTW, when classifying $UCR_{43}$ 
(\tableref{table:comparison2}). For r-STSF, the results are obtained  from 10 runs on each dataset. The results for ROCKET and InceptionTime are reported from the original papers. The results of 1NN-DTW are as reported by~\cite{dempster2020rocket}. There were no results reported for HIVE-COTE, TS-CHIEF, and new HIVE-COTE variants (i.e., HC-TDE and HC-CIF) on all of these new datasets. Hence, all these methods were not included in this subsection. As shown in~\figref{fig:43dsets_ordered_comparisonOfEvaluationMetrics}, when using $UCR_{43}$, ROCKET is the most effective classifier according to the average ranking and the weighted average accuracy. r-STSF follows closely after ROCKET, and is the second best classifier also according to both metrics. 
These results confirm the generalizability of r-STSF to new datasets, noting that it has to trade-off accuracy for interpretable classifications whereas ROCKET, with its complex features, does not provide interpretability. 
Besides, r-STSF achieves strong results despite the fact that 
approximately 25\% of additional datasets 
are related to problems involving ``gesture", which are prone to having \emph{phase-independent} discriminatory intervals and hence are more challenging for r-STSF. 

\begin{table*}[]
\centering
\caption{Average classification accuracy over 10 runs of r-STSF when classifying on $UCR_{43}$. Other methods are as reported in their original proposals or accompanying websites}
\label{table:comparison2}
% \resizebox{\textwidth}{!}{%
\begin{tabular}{@{}lcccc@{}}
\toprule
\textbf{Datasets}                 & \textbf{1NN-DTW} & \textbf{InceptionTime} & \textbf{ROCKET} & \textbf{r-STSF}   \\ \midrule
\textbf{ACSF1}                   & 62.00          & 89.60  & 87.80           & \textbf{90.40}  \\
\textbf{AllGestureWiimoteX}      & 71.71          & \textbf{77.23}  & 76.19           & 65.40         \\
\textbf{AllGestureWiimoteY}      & 73.00          & \textbf{81.34}  & 76.17           & 68.81          \\
\textbf{AllGestureWiimoteZ}      & 65.14          & \textbf{79.20}  & 74.91           & 65.91          \\
\textbf{BME}                     & 98.00          & 99.33           & \textbf{100.00} & 99.87 \\
\textbf{ChinaTown}               & 95.36          & 98.31           & 98.02           & \textbf{98.57}  \\
\textbf{Crop}                    & 71.17          & 75.10           & 75.02           & \textbf{77.70}  \\
\textbf{DodgerLoopDay}           & 58.75          & 15.00           & 57.63           & \textbf{63.38}  \\
\textbf{DodgerLoopGame}          & \textbf{92.75} & 85.36           & 87.25           & 85.65           \\
\textbf{DodgerLoopWeekend}        & \textbf{97.83}   & 96.96                  & 97.25           & \textbf{97.83}  \\
\textbf{EOGHorizontalSignal}     & 47.51          & 58.78           & \textbf{64.09}  & 57.16           \\
\textbf{EOGVerticalSignal}       & 47.51          & 46.41           & \textbf{54.23}  & 52.82           \\
\textbf{EthanolLevel}            & 28.20          & \textbf{80.36}  & 58.20           & 61.84           \\
\textbf{FreezerRegularTrain}     & 90.70          & 99.65           & 99.76           & \textbf{99.92}  \\
\textbf{FreezerSmallTrain}       & 67.58          & 86.57           & 95.19           & \textbf{98.14}  \\
\textbf{Fungi}                   & 82.26          & \textbf{100.00} & \textbf{100.00} & 87.53           \\
\textbf{GestureMidAirD1}         & 63.85          & 73.23           & \textbf{80.62}  & 68.85           \\
\textbf{GestureMidAirD2}         & 60.00          & \textbf{70.77}  & 68.31           & 62.08           \\
\textbf{GestureMidAirD3}         & 37.69          & 36.62           & \textbf{57.85}  & 49.46           \\
\textbf{GesturePebbleZ1}         & 82.56          & 92.21           & \textbf{96.63}  & 89.77           \\
\textbf{GesturePebbleZ2}         & 77.85          & 87.47           & \textbf{89.11}  & 88.80           \\
\textbf{GunPointAgeSpan}         & 96.52          & 98.73           & \textbf{99.68}  & 98.99           \\
\textbf{GunPointMaleVersusFemale} & 97.47            & 99.56                  & 99.78           & \textbf{100.00} \\
\textbf{GunPointOldVersusYoung}  & 96.51          & 96.19           & 99.05           & \textbf{100.00} \\
\textbf{HouseTwenty}             & 94.12          & \textbf{97.48}  & 96.39           & 91.93           \\
\textbf{InsectEPGRegularTrain}   & 82.73          & 99.84           & 99.96           & \textbf{100.00} \\
\textbf{InsectEPGSmallTrain}     & 69.48          & 94.14           & 98.15           & \textbf{100.00} \\
\textbf{MelbournePedestrian}     & 84.82          & 90.75           & 90.35           & \textbf{97.12}  \\
\textbf{MixedShapesRegularTrain} & 90.89          & 96.62           & \textbf{97.04}  & 94.90           \\
\textbf{MixedShapesSmallTrain}   & 83.26          & 91.16           & \textbf{93.86}  & 89.59           \\
\textbf{PLAID}                   & 83.61          & \textbf{93.74}  & 88.96           & 89.85           \\
\textbf{PickupGestureWiimoteZ}   & 66.00          & 74.40           & \textbf{81.00}  & 70.40           \\
\textbf{PigAirwayPressure}       & 9.62           & \textbf{53.17}  & 8.85            & 37.50           \\
\textbf{PigArtPressure}          & 19.71          & \textbf{99.33}  & 95.29           & 92.40           \\
\textbf{PigCVP}                  & 15.87          & \textbf{95.29}  & 93.27           & 68.32           \\
\textbf{PowerCons}               & 92.22          & 94.78           & 93.11           & \textbf{100.00} \\
\textbf{Rock}                    & 84.00          & 75.20           & \textbf{89.80}  & 75.80           \\
\textbf{SemgHandGenderCh2}       & 84.50          & 80.23           & 92.30           & \textbf{96.60}  \\
\textbf{SemgHandMovementCh2}     & 63.78          & 42.00           & 64.44           & \textbf{83.49}  \\
\textbf{SemgHandSubjectCh2}      & 80.00          & 78.71           & 88.36           & \textbf{88.93}  \\
\textbf{ShakeGestureWiimoteZ}    & 84.00          & \textbf{90.00}  & 89.20           & 85.60           \\
\textbf{SmoothSubspace}          & 94.67          & \textbf{98.13}  & 97.93           & 98.00           \\
\textbf{UMD}                     & 97.22          & 98.19           & 99.24           & \textbf{99.79}  \\
\textbf{}                        &                &                 &                 &                 \\ \bottomrule
\textbf{Average rank}            & 3.53         & 2.38         & \textbf{1.99}        & \textcolor{blue}{\textbf{2.09}}          \\
\textbf{Weighted Average accuracy}            & 55.92         & 63.29         & \textbf{65.62}        & \textcolor{blue}{\textbf{65.31}}          \\
\bottomrule
\end{tabular}%
% }
\end{table*}

\section{Sensitivity Analysis}

\label{subsec:sensitivity}
To provide a comprehensive analysis of r-STSF, we explore the impact of different parameters of r-STSF  on its classification accuracy. The parameters to consider are: (i) number of candidate discriminatory interval feature sets, (ii) time series representations, and (iii) aggregation functions. The number of trees in the ensemble is not considered in this analysis. It is set to 500 following the other tree-based TSC methods~\citep{deng2013time,shifaz2020ts}.
Moreover, we also present experimental results to support our claims regarding the improvements to the average classification accuracy of r-STSF when (i) using random partitions (instead of middle-point partitions) of intervals when searching for candidate discriminatory interval features, and (ii) building ensembles of randomized binary trees (i.e., \emph{extra-trees}) instead of non-randomized binary trees (i.e., \emph{random forest}). All the results presented in this section are based on the average classification accuracy over three runs of r-STSF when classifying on $UCR_{85}$. r-STSF uses by default (unless said otherwise) the four time series representations and the nine aggregation functions mentioned in Section~\ref{sec:approach}.

\subsection{Number of Candidate Discriminatory Interval  Feature Sets} 
\begin{figure}[h]
\centering
\subfloat[\label{fig:nCandDiscrIntvFeats_ar}]{
	\includegraphics[scale=0.15]{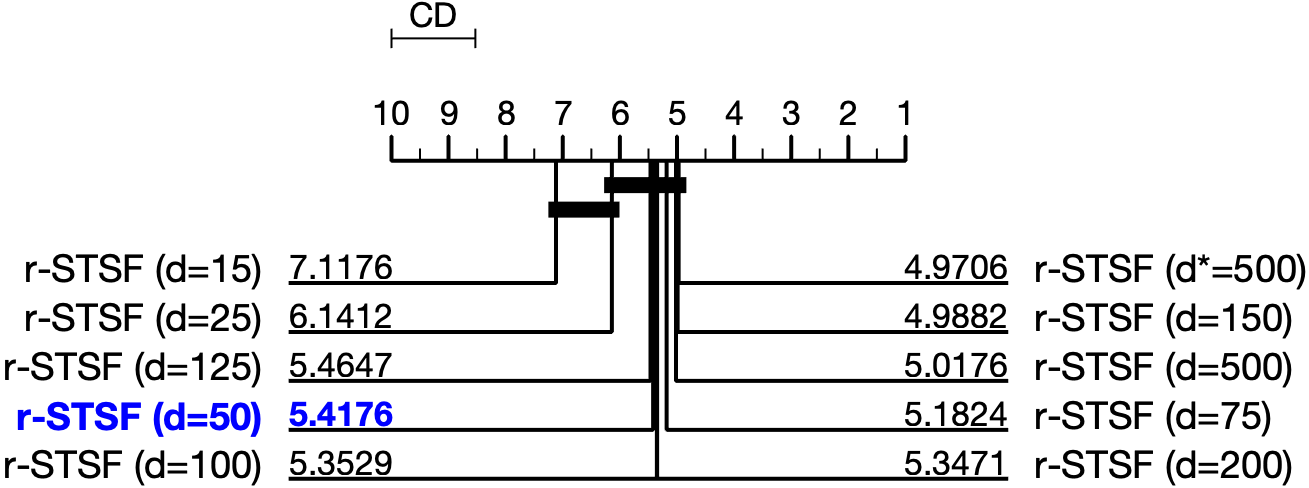}
}
\subfloat[\label{fig:nCandDiscIntvFeats_waa}]{
	\includegraphics[scale=0.060]{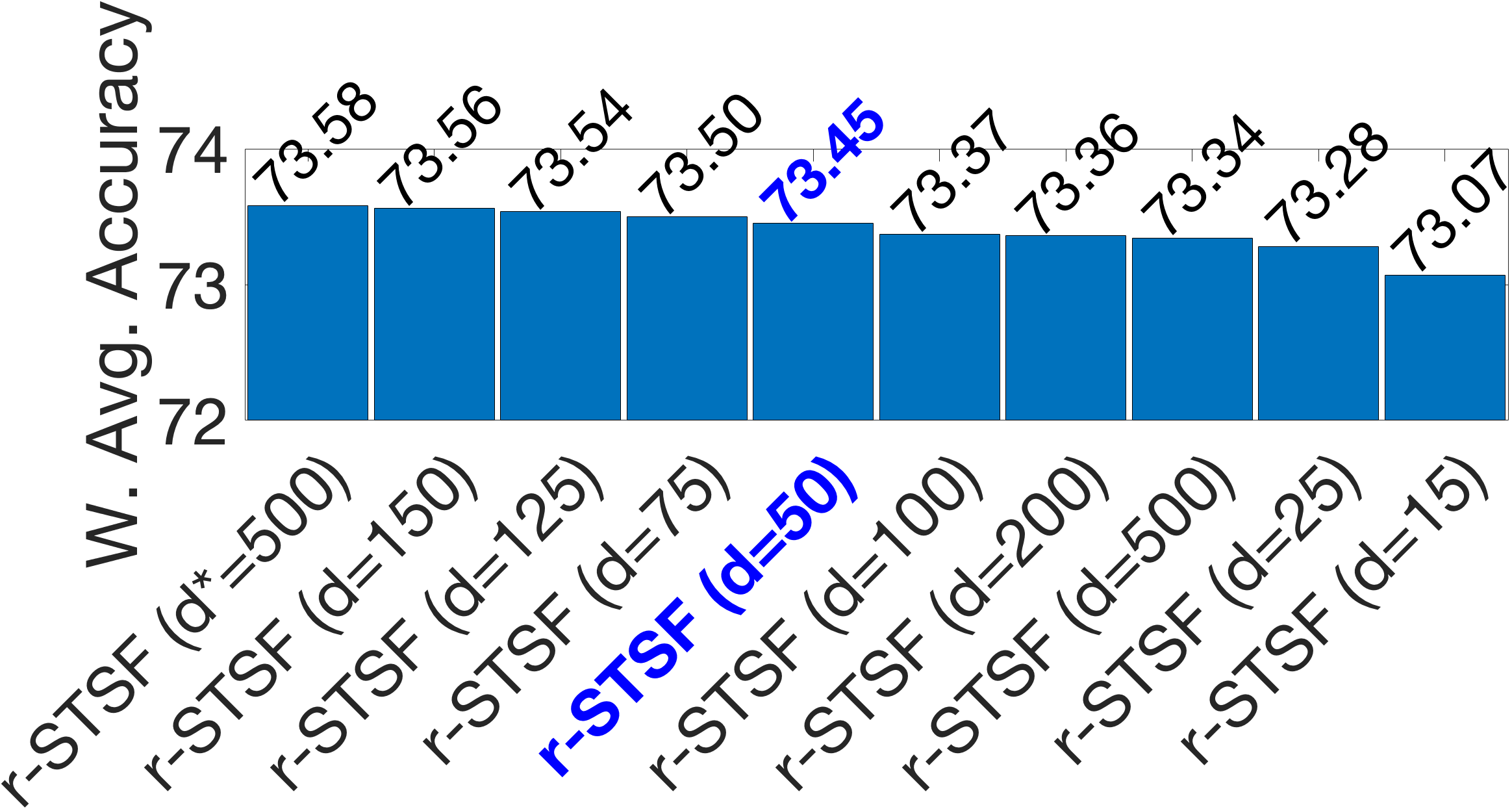}
}
\caption{\textcolor{black}{\textbf{(a)} Critical difference diagram of average ranks and \textbf{(b)} Weighted average accuracy of r-STSF with varying numbers of sets of candidate discriminatory interval features, i.e., $d$. After $d=50$, the average ranks are very close and statistically similar. According to the weighted average accuracy, r-STSF (d=50) is slightly more effective than r-STSF (d=500). This validates our decision of just computing a small number of sets of candidate intervals.
The special case of training the ensemble of trees as in STSF (i.e., one set per tree -- r-STSF (d$^{*}$=500)) is the most accurate setting, but it is an order of magnitude slower and not significantly different than r-STSF (d=50). This validates our decision to build each tree from features randomly selected from a larger superset}}
\label{fig:nCandDiscrIntvFeats}
\end{figure}

The extraction of $d$ sets of candidate discriminatory interval features may become expensive on large datasets with very long series. This parameter has a direct impact on the efficiency of r-STSF. The process to extract a single set of candidate intervals (i.e., $d=1$) is summarized in Algorithm~\ref{algo:candidate_interval_feats}. r-STSF's predecessor, STSF, extracts a set of intervals for each tree of the ensemble. For an ensemble with 500 trees, STSF extracts 500 sets of candidate discriminatory interval features. In r-STSF, we propose to extract 50 sets of interval features (i.e., $d=50$). The training process of r-STSF does not only differ from that of STSF in the number of extracted set of intervals, but in how such features are used to build each of the trees in the ensemble. As detailed in Section~\ref{subsec:rstsfVSstsf}, STSF uses each set of intervals to build each of the $r$ trees in the ensemble, whereas r-STSF merges all sets into a superset $\mathcal{F}$. Then, each of the $r$ trees in the ensemble is built by using a set of randomly selected interval features from $\mathcal{F}$. We study the classification accuracy of r-STSF when setting $d$ to $\{15, 25, 50, 75, 100, 125, 150, 200, 500\}$. We also include a special case, $d^{*}=500$, where each the 500 trees of the ensemble is build as in STSF (i.e., each set of intervals builds one tree at a time).
As shown in \figref{fig:nCandDiscrIntvFeats_ar}, after extracting 25 sets of candidate interval features ($d=25$), there is no significant difference in terms of classification accuracy. When increasing $d$ beyond 50, the average rank only drops marginally. Thus, for r-STSF, we set $d$ to 50 by default for efficiency considerations. Using $d=50$ instead of $d=500$ or $d^{*}=500$ makes r-STSF an order of magnitude faster in the feature extraction process without affecting its effectiveness. \figref{fig:nCandDiscIntvFeats_waa} supports this design choice, where r-STSF (d=50) is slightly more effective than r-STSF (d=500). Although r-STSF (d$^{*}$=500) has a slightly higher weighted average accuracy than r-STSF (d=50), the improvement in efficiency obtained with the latter compensates for this insignificant difference in accuracy.

\subsection{Time Series Representations}
\begin{figure}[]
\centering
\subfloat[\label{fig:tsrepr-comparison_ar}]{
	\includegraphics[scale=0.15]{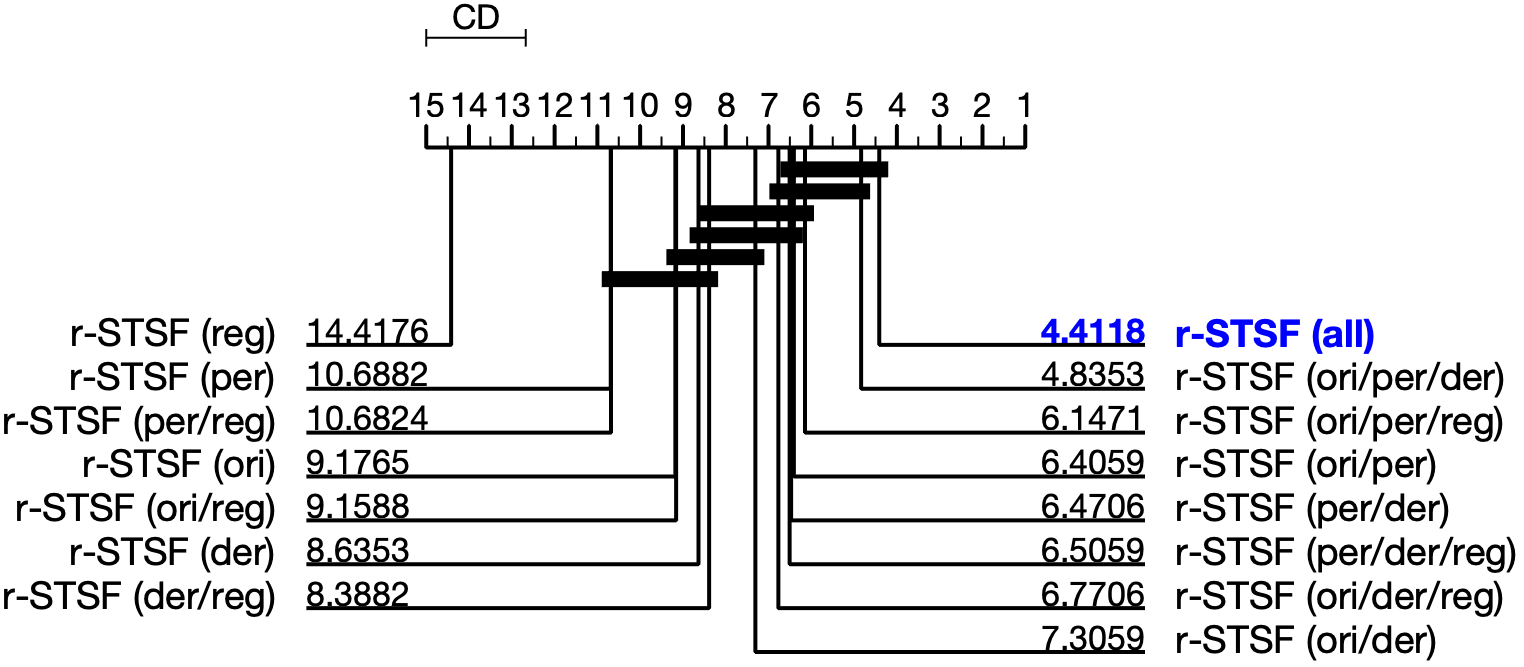}
}\\
\subfloat[\label{fig:tsrepr-comparison_waa}]{
	\includegraphics[scale=0.07]{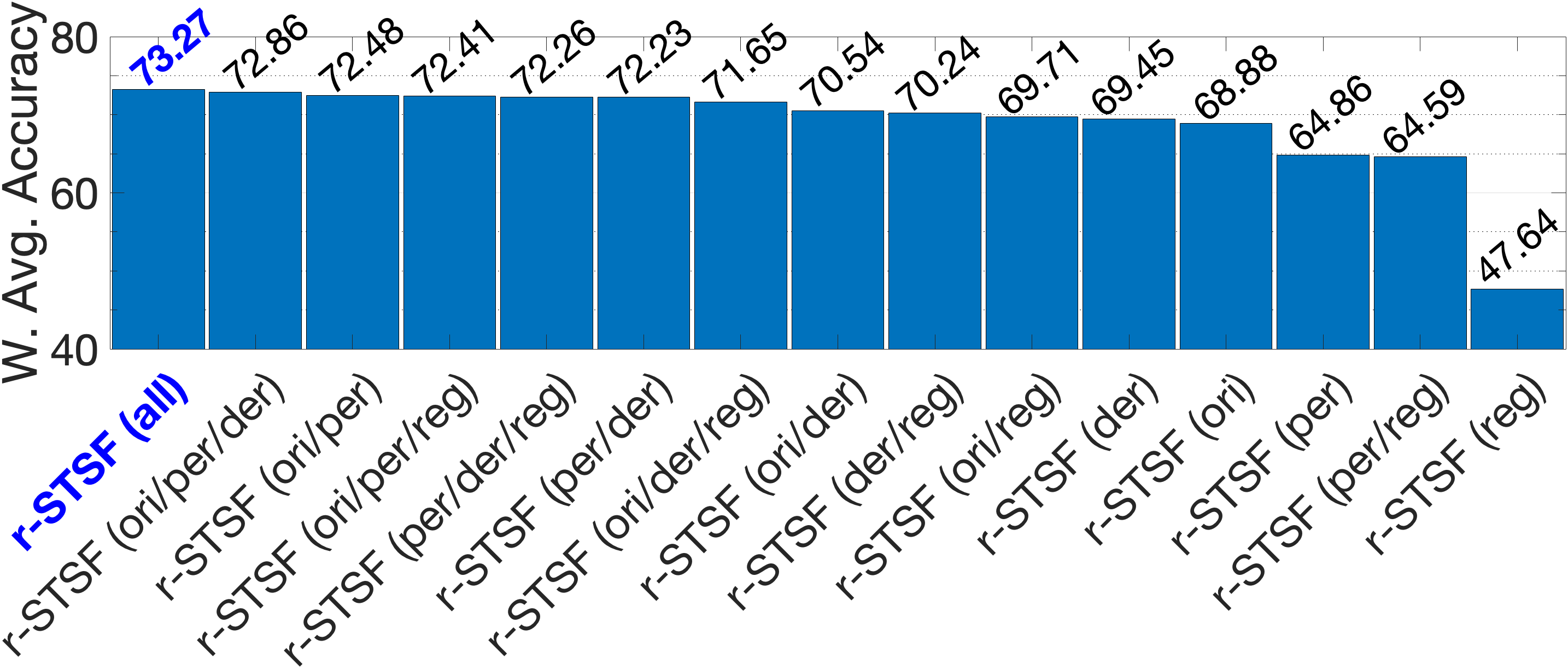}
}
\caption{\textcolor{black}{\textbf{(a)} Critical difference diagram of average ranks and \textbf{(b)} Weighted average accuracy of r-STSF with different combinations of time series representations: original (ori), periodogram (per), derivative (der), and autoregressive (reg). r-STSF is more effective when using all four time series representations, i.e., r-STSF (all) for both metrics}}
\label{fig:tsrepr-comparison}
\end{figure}

r-STSF uses four time series representations to extract discriminatory interval features, including the original (raw),  \textcolor{black}{i.e., $X_{O}$},  periodogram, \textcolor{black}{i.e., $X_{P}$}, derivative, \textcolor{black}{i.e., $X_{D}$}, and autoregressive representations,  \textcolor{black}{i.e., $X_{G}$}, of the training time series. As presented in~\figref{fig:tsrepr-comparison_ar} and ~\figref{fig:tsrepr-comparison_waa}, the best rank is and the highest weighted average accuracy are achieved by using all four time series representations. Individually, $X_{D}$ is the most effective (i.e., r-STSF (der)). Using only $X_{G}$ (i.e., r-STSF (reg)) makes r-STSF perform poorly (in terms of accuracy).

$X_{G}$ is relevant when past values have an effect on current values of a given time series. If a time series does not hold this premise, it is challenging to classify such series by just using this representation. Moreover, even for a series holding this premise it is necessary to estimate the correct lag order which allows to identify the lagged relationships within a series. If a large number of time series classification problems from $UCR_{85}$ cannot be modeled by an autoregressive process, then using just the autoregressive representation of the series is not enough to achieve accurate classifications. However, when comparing the classification accuracy of r-STSF (all) and r-STSF (ori/per/der) (~\figref{fig:tsrepr-comparison2}, Appendix~\ref{appendix:autoregressive-importance}) we find that the addition of $X_{G}$ improves the  classification accuracy in many datasets. In some datasets such as Car, Ham, OSULeaf, Phoneme,  ProximalPhalanxOutlineCorret, ScreenType, SonyAIBORobotSurface2, Wine, and Worms this improvement is between 2\% to 6\%.

$X_{D}$ does not require any assumptions about the data other than the presence of trends or changes in the series values which are very common in time series data.

\subsection{Aggregation Functions}
\begin{figure}[h]
\centering
\subfloat[\label{fig:aggfns-comparison_ar}]{
	\includegraphics[scale=0.15]{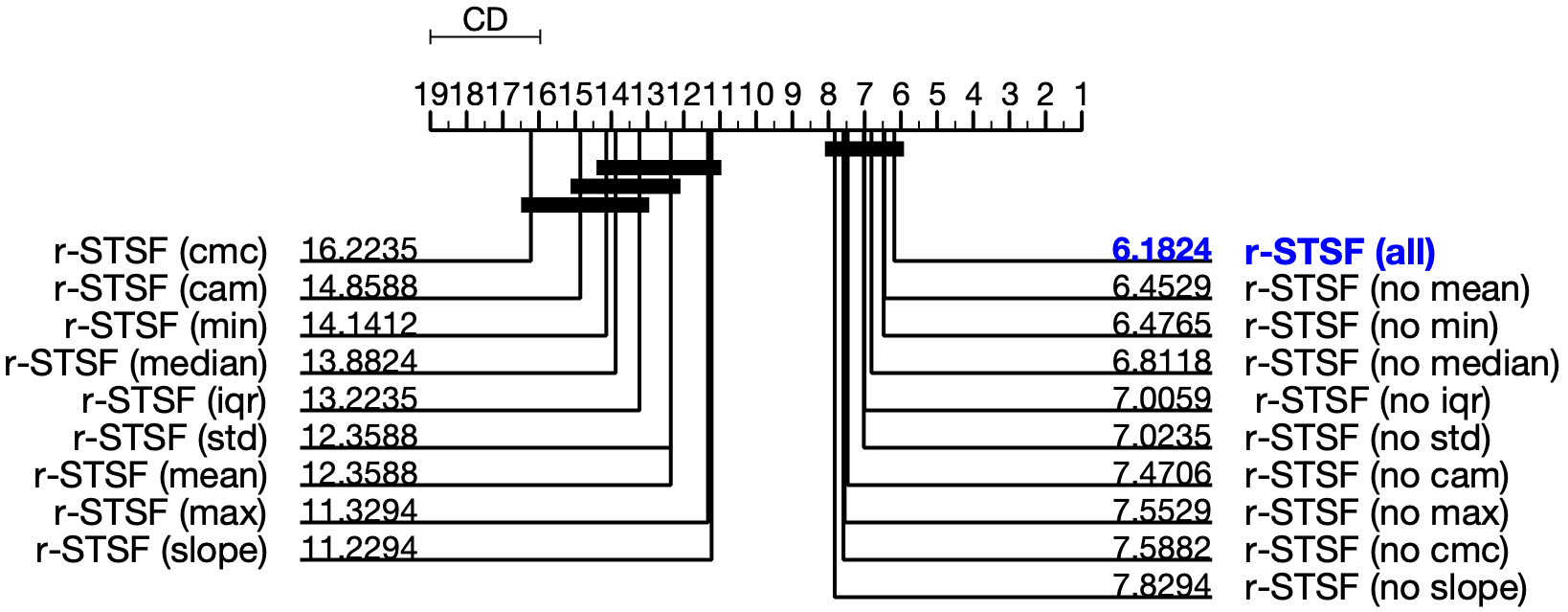}
}\\
\subfloat[\label{fig:aggfns-comparison_waa}]{
	\includegraphics[scale=0.07]{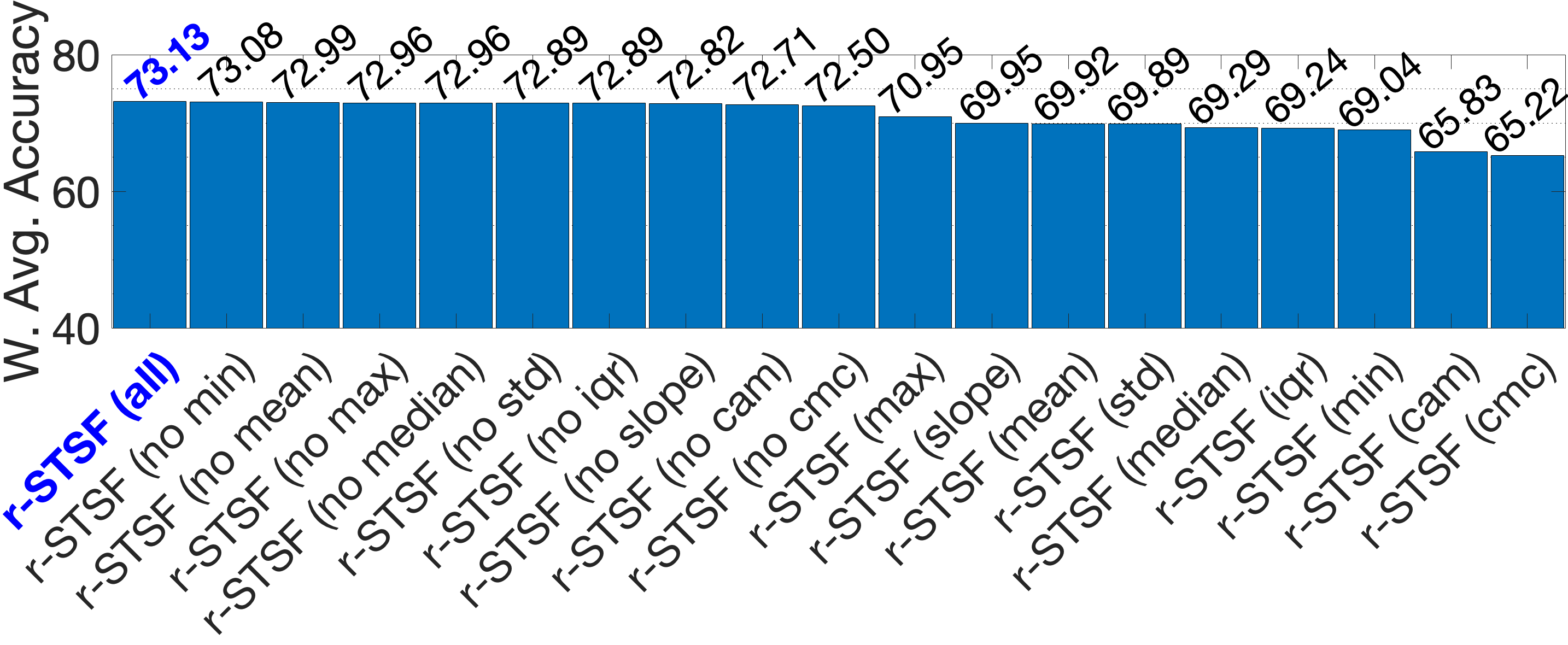}
}
\caption{\textcolor{black}{\textbf{(a)} Critical difference diagram of average ranks and \textbf{(b)} Weighted average accuracy of r-STSF with different combinations of aggregation functions (or statistics). r-STSF (\emph{statistic}) means using r-STSF just with the respective \emph{statistic}. r-STSF (no \emph{statistic}) means using all aggregation functions but \emph{statistic}. r-STSF (all) uses all the aggregation functions and is the most effective on both evaluation metrics.}}
\label{fig:aggfns-comparison}
\end{figure}

We evaluate the impact of the aggregation functions (aggregation statistics) on the classification accuracy of r-STSF, which by default uses nine statistics to compute the interval features (detailed in Section~\ref{subsec:intervalfeaturesextraction}). 
For each statistic, we compute r-STSF by (i) using that statistic and (ii) removing it from the set of (nine) aggregation functions. As shown in~\figref{fig:aggfns-comparison}, generic statistics such as the mean, median, and min contribute slightly to the effectiveness of r-STSF. The most important statistics are slope and max, which yield the individual best ranks and highest weighted average accuracies (i.e., r-STSF (slope) and r-STSF (max)). Their importance is also reflected when they are removed from the set of aggregation functions (i.e., r-STSF (no slope) and r-STSF (no max)), which lead to the largest drops in the average rank. Besides, although individually cmc and cam are the least important, if they are removed from the aggregation functions, r-STSF has a substantial drop in its average rank (similar to removing the max statistic) and in its weighted average accuracy. This suggests that cmc and cam are more effective when combined with other statistics. 

\begin{figure}[h]
\centering
\subfloat[\label{fig:aggfns-cmc-cam_ar}]{
	\includegraphics[scale=0.25]{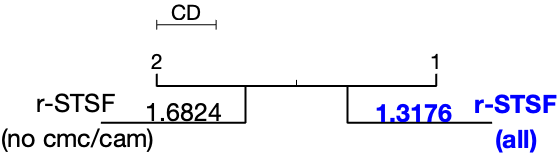}
} \hspace{5mm}
\subfloat[\label{fig:aggfns-cmc-cam_waa}]{
	\includegraphics[scale=0.07]{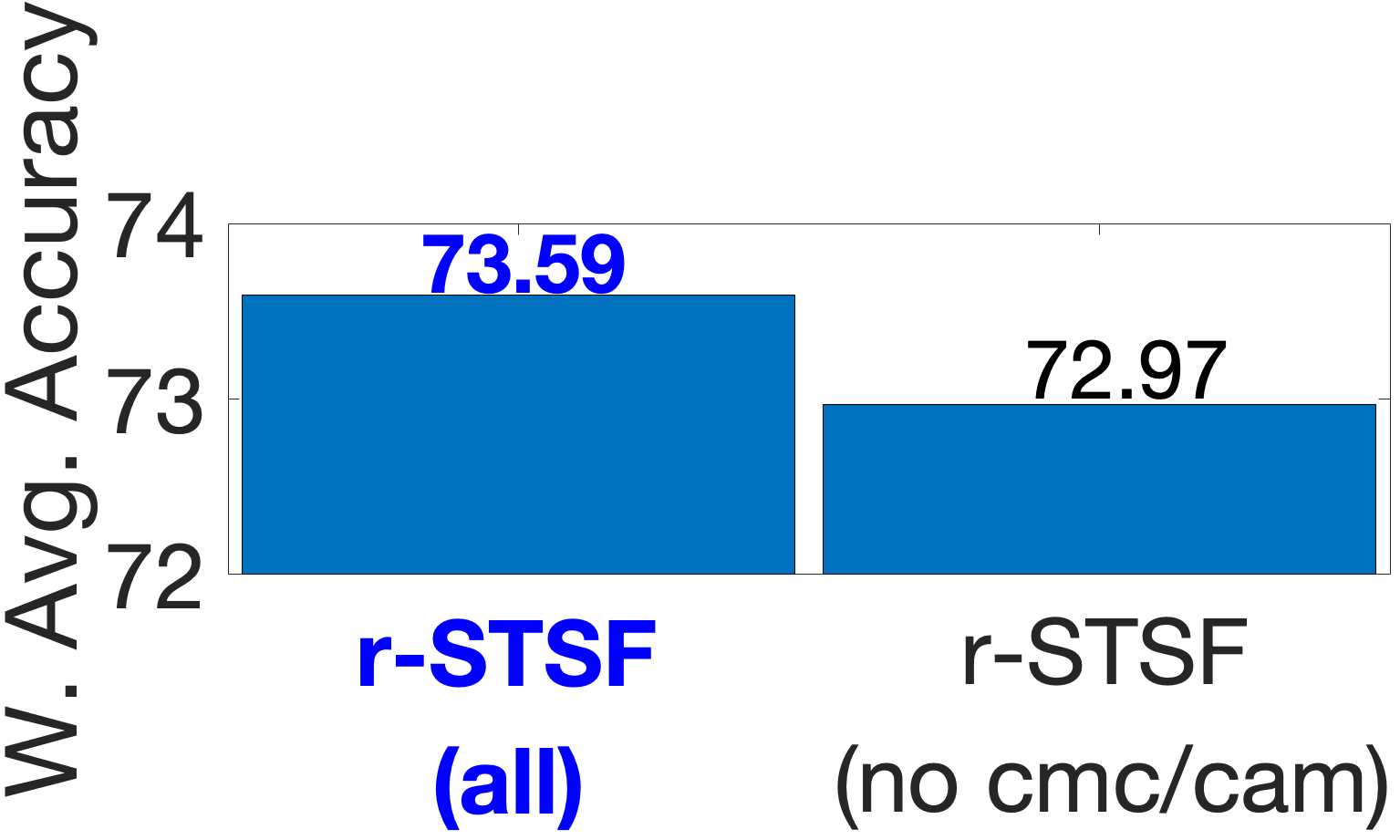}
}
\caption{\textcolor{black}{\textbf{(a)} Critical difference diagram of average ranks and \textbf{(b)} Weighted average accuracy of r-STSF with and without cmc and cam aggregation functions. r-STSF (no cmc/cam) means using r-STSF without cmc and cam statistics. r-STSF (all) uses all the aggregation functions and is the most effective on both evaluation metrics}}
\label{fig:aggfns-cmc-cam}
\end{figure}

As shown in~\figref{fig:aggfns-cmc-cam}, not including the cmc and cam aggregation functions makes r-STSF significantly less accurate when compared to the version of r-STSF which includes them into its set of statistics. Moreover, as shown in~\figref{fig:aggfns-cmc-cam-comparison}, Appendix~\ref{appendix:cmc-cam-importance}, the addition of cmc and cam statistics improves the classification accuracy in many datasets. In some datasets such as Beef, Ham, RefrigerationDevices, ToeSegmentation1, Wine, and WormsTwoClasses this improvement is between 2\% to 5\%.

\subsection{Impact of the Perturbation Scheme}  
\label{subsec:perturbation-scheme}
r-STSF uses a novel perturbation scheme to create an ensemble of  uncorrelated trees for higher   accuracy. Our scheme employs (i) random partitions when assessing the discriminatory quality of sub-series and (ii) randomized trees to build the ensemble of trees for classification. We evaluate the impact of these steps in this section. 

\begin{figure}[h]
\centering
\subfloat[\label{fig:midcutpointVSrandcutpoint-ar}]{
	\includegraphics[scale=0.15]{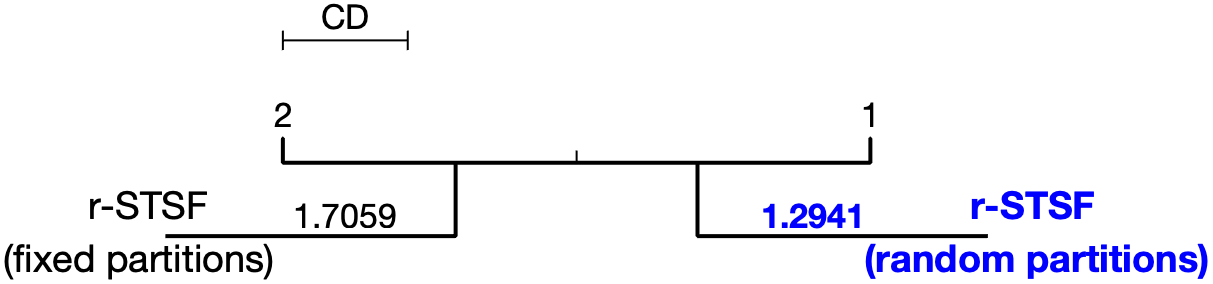}
}
\subfloat[\label{fig:midcutpointVSrandcutpoint-waa}]{

	\includegraphics[scale=0.060]{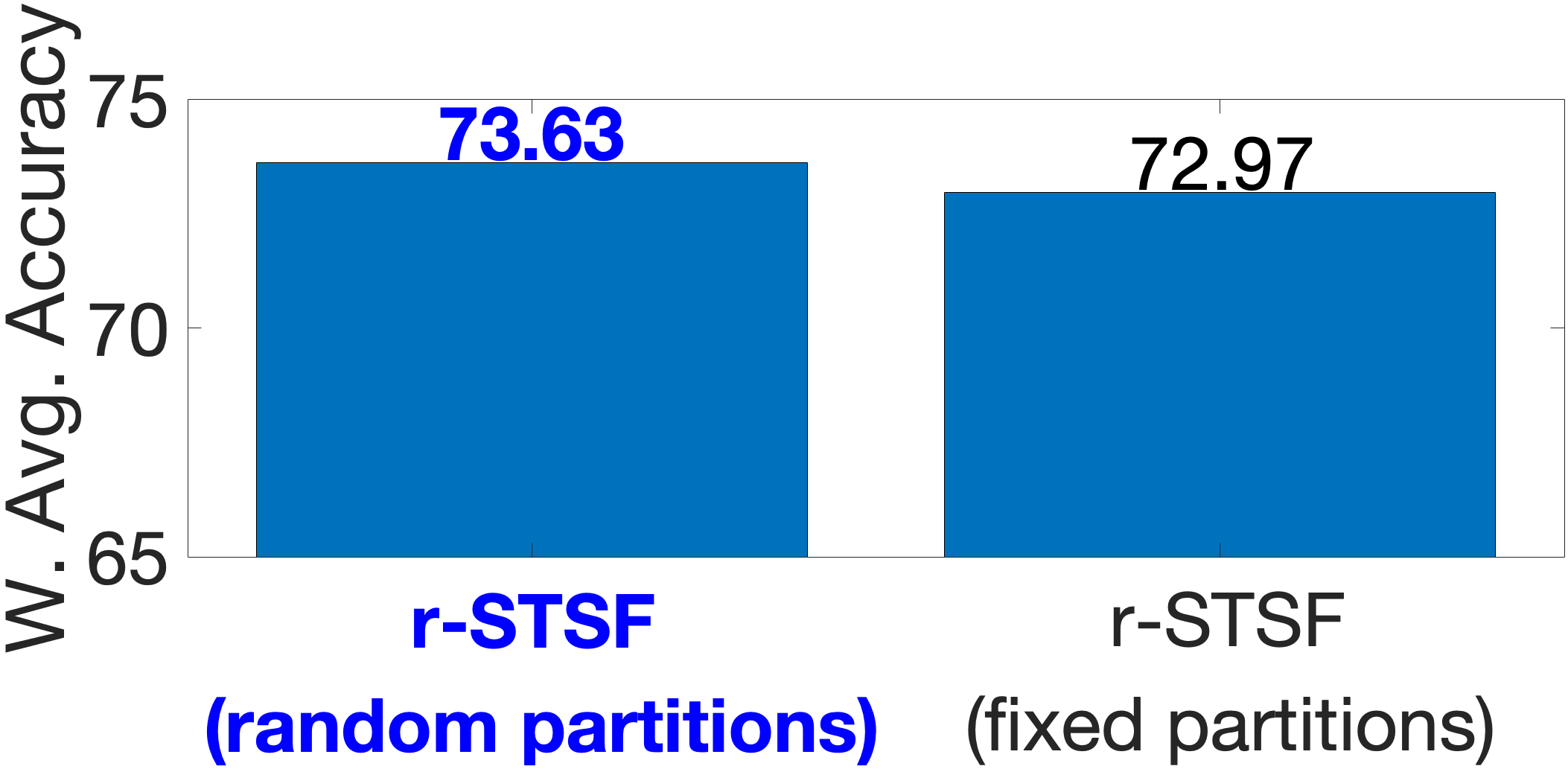}
}
\caption{\textcolor{black}{\textbf{(a)} Critical difference diagram of average ranks and \textbf{(b)} Weighted average accuracy of r-STSF with fixed and random partitions. r-STSF with random partitions is more effective (according to both evaluation metrics) and is significantly more accurate than r-STSF with fixed partitions}}
\label{fig:midcutpointVSrandcutpoint}
\end{figure}

\textbf{Random partitions:} r-STSF's predecessor, STSF, uses the middle-point of the sub-series as a cut point and builds fixed partitions when searching for candidate discriminatory interval features.
In Section~\ref{subsec:intervalfeaturesextraction}, we showed that such a technique may not be the best if the goal is to create an  ensemble of uncorrelated trees. In r-STSF, we rely on random partitions when assessing the discriminatory quality of sub-series, which increases the average accuracy of r-STSF by 1\% compared to using fixed partitions. Moreover, as shown in~\figref{fig:midcutpointVSrandcutpoint}, r-STSF with random partitions is significantly more accurate than r-STSF with fixed partitions.

\textbf{Randomized trees:} The Extra-trees (ET) algorithm builds an ensemble of randomized binary trees to decrease the variance of the ensemble (by creating uncorrelated trees) and thus achieves high classification accuracy. In randomized trees, the cut-point of each feature is randomly selected when looking for the feature that provides the best split. These trees are different from those in a random forest (RF), where for each feature only the best cut-point (i.e., split with the lowest entropy) is selected. r-STSF's predecessor, STSF, follows RF and builds non-randomized binary trees. For r-STSF, we create an ensemble of randomized binary trees. 
    
As detailed in~\tableref{table:rstsfVSstsf}, r-STSF first extracts a set of candidate discriminatory interval features, i.e., $\mathcal{F}$. Then, (at node level) it randomly selects $\sqrt{\mathcal{F}}$ interval features from $\mathcal{F}$ when looking for the best random split. In this section, we denote r-STSF as r-STSF (ET) and compare it against two variants of r-STSF which use non-randomized trees and are denoted as r-STSF (RF) and r-STSF (RF all), respectively. r-STSF (RF) selects $\sqrt{\mathcal{F}}$ interval features to split the tree nodes and builds an ensemble of non-randomized trees, whereas r-STSF (RF all) also builds non-randomized trees but uses all features from $\mathcal{F}$ for the splits. Thus, r-STSF (RF) builds ``more" randomized trees than r-STSF (RF all), but ``less" randomized than r-STSF (ET). We also include in our comparisons a variant of r-STSF, denoted as r-STSF (ET 1), that randomly selects a single interval feature from $\mathcal{F}$ to split each node and builds randomized trees. Hence, on r-STSF (ET 1), the feature and its cut-point are randomly selected. Trees built in this manner are known as \emph{totally randomized trees}~\citep{geurts2006extremely} and are ``more" randomized than those of r-STSF (ET). 

\begin{figure}[h]
\centering
\subfloat[\label{fig:RFvsET-ar}]{
\hspace{-10mm}
	\includegraphics[scale=0.34]{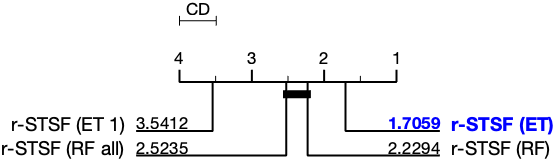}
}
\subfloat[\label{fig:RFvsET-waa}]{
\hspace{2mm}
	\includegraphics[scale=0.060]{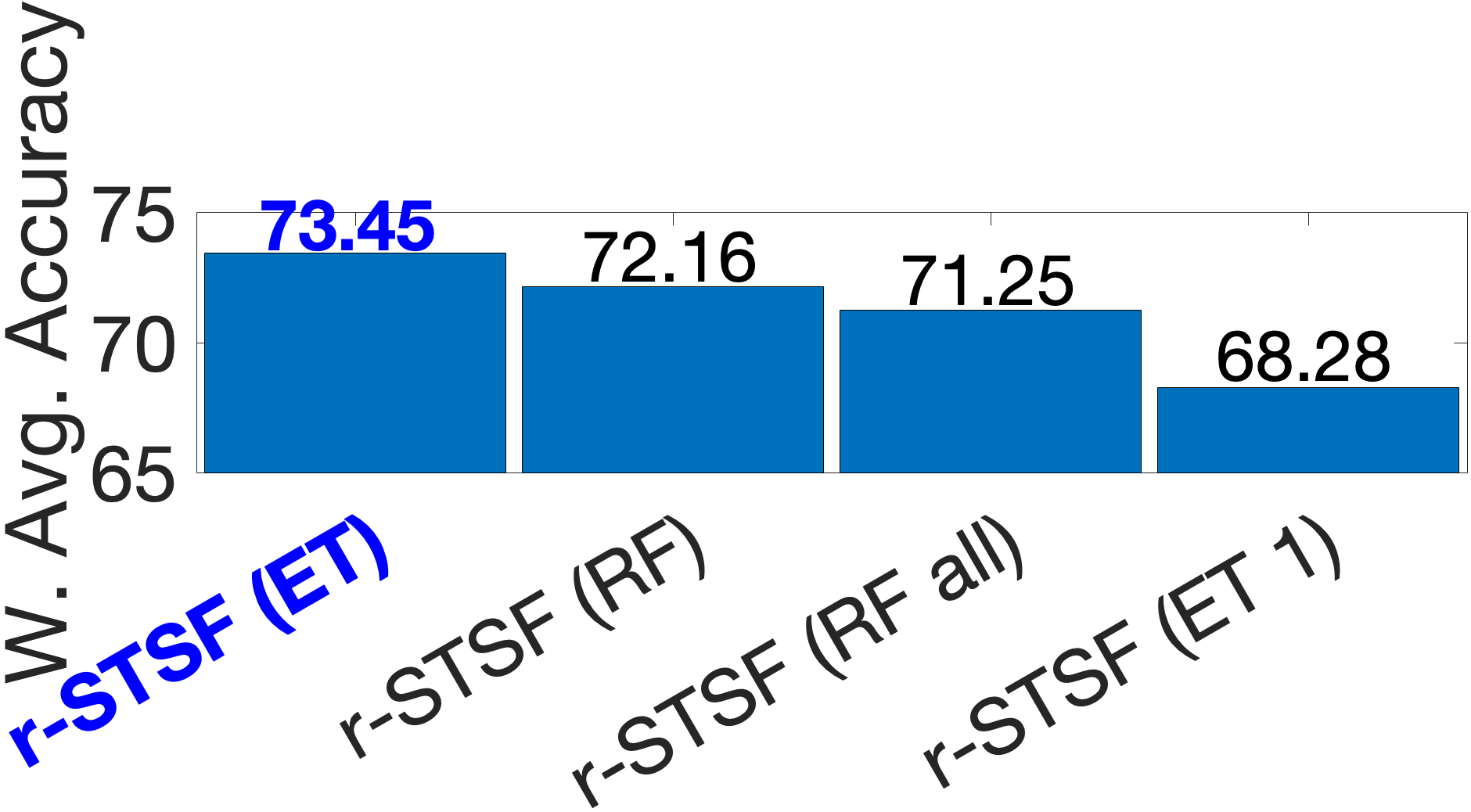}
}
\caption{\textcolor{black}{\textbf{(a)} Critical difference diagram of average ranks and \textbf{(b)} Weighted average accuracy of r-STSF at different levels of perturbation. r-STSF (ET 1) is the most extreme case where a single feature and cut-point are randomly selected to split the nodes. It follows r-STSF (ET) where the nodes are split according to the random cut-point (from a group of randomly selected features) that provides the best split. Next, r-STSF (RF) which similar to r-STSF (ET) selects a group of features, but looks for the best cut-point instead of selecting one at random. r-STSF (RF all) also looks for the best split, but among all the available features, i.e., not from a selected group. r-STSF (ET) is more effective (according to both evaluation metrics) and significantly more accurate than the other versions of r-STSF}}
\label{fig:RFvsET}
\end{figure}

As shown in~\figref{fig:RFvsET}, r-STSF (ET) achieves the highest weighted average accuracy and is significantly more accurate than the versions of r-STSF using non-randomized trees. Moreover, the poor performance of r-STSF (ET 1) suggests that there is a limit on the level of perturbation that can be applied when building uncorrelated trees. Trees that are ``too'' uncorrelated negatively impact the classification accuracy.

\color{black}

\section{Interpretable Classification}
\label{sec:roi}

To provide further insights to the classification decisions of r-STSF, we use the discriminatory intervals computed by it.
Each discriminatory interval has information on its interval boundary (i.e., starting and ending indices), the aggregation function (e.g., mean), and the time series representation (e.g., periodogram) that was used to generate the corresponding feature. To highlight the regions of the time series which are discriminatory, i.e., maximizing class separability, we count the number of times that each data point of a testing series is contained in the discriminatory intervals. We normalize such counts and use a heatmap to visualize the most interesting regions.

Most of the interpretable classifiers identify and highlight relevant time-stamps to allow for interpretations.
TSF and CIF capture discriminatory features embedded in sub-series, whose location in the time series (i.e., time-stamps) is highlighted for interpretability. However, discriminatory or relevant features are not always presented in the time series in their original form (i.e., time-stamped data), but can be found in additional representations/transformations of the time series~\citep{bagnall2012transformation}. For example, mtSAX-SEQL+LR~\citep{le2019interpretable} captures discriminatory features in symbolic representations of the time series. Given that symbol-based representations do not directly support interpretability, mtSAX-SEQL+LR maps the features back to their locations in the (original) series to enable interpretability.

\textcolor{black}{As discussed in Section~\ref{sec:approach},} r-STSF uses different \textit{time series representations} for classification. 
These representations allow for interpretations directly, i.e., it is not necessary to map their relevant features back to the original series. In this section, we present cases where the original time series \textcolor{black}{does not support} the identification of discriminatory features and show how our additional representations provide more discriminatory features while enabling  interpretability. More importantly, since r-STSF uses simple statistics to compute its features, \emph{our interpretations are not only limited to highlighting discriminatory regions, i.e., we can also  provide  explanations for the classifier's decision}. For example, if a discriminatory interval was extracted with the slope, it can be interpreted that sub-series from one class may exhibit a different trend from that of sub-series of another class.

\subsection{Computing Importance of Time Series Representations and Aggregation Functions}

Using different time series representations increases the chances to find discriminatory interval features. Depending on the time series dataset, some representations may be more important than the others. Similarly, depending on the time series representation, there are aggregation functions which may contribute more than the others when extracting discriminatory intervals.

Discriminatory interval features are located in the nodes of our tree-based ensemble, i.e., $\mathcal{F}^{*}$ in~\figref{fig:overview}. Each of these interval features keeps a record of the time series representation and aggregation function used for its computation. To identify which time series representation and which aggregation functions contribute the most over a given dataset, we estimate the importance of their corresponding features by computing the \emph{mean decrease impurity} (MDI)~\citep{louppe2013understanding} of such features.
At each tree node $\theta$, the interval feature $\boldsymbol{f}$ that achieves the largest impurity decrease is selected as the best split $S_{\theta}$. When using \emph{entropy} as the impurity measure, the information gain 
(IG) computes the impurity decrease (see Eq.~\eqref{eq:ig}).
Thus, the importance of an interval feature $\boldsymbol{f}$ is estimated by adding up the weighted impurity decrease $\rho_{\theta}  \hbox{IG}(\boldsymbol{f},\theta)$ of every node $\theta$ where $\boldsymbol{f}$ is used (to split that node), averaged over all $r$ trees in the forest:

\begin{equation}
    \label{eq:mdi}
    \hbox{MDI}(\boldsymbol{f}) =
    \frac{1}{r}
    \sum\limits_{\intercal}
    \sum\limits_{\theta \in  \intercal : S_{\theta} = \boldsymbol{f} }
    \rho_{\theta}  \hbox{IG}(\boldsymbol{f},\theta)
\end{equation}

where $\intercal$ is a tree classifier and $\rho_{\theta}$ is the probability of reaching node $\theta$. Probability $\rho_{\theta}$ is calculated as the number of time series instances that reach the node, divided by the total number of time series instances $n$.
The higher the MDI, the more important the interval feature is. Time series representations and aggregation functions used to compute \emph{important} interval features are also considered to be more important. To estimate the importance of a representation (or aggregation function), we average the importance of their corresponding features. For example, to estimate the importance of the periodogram representation, we compute the average of the MDI (i.e., importance) of all the features extracted  from this representation. After estimating the importance of the remaining representations (i.e., original, derivative and autoregressive), the importance values are normalized. A similar process is used to estimate the importance of each aggregation function.

\begin{figure}[h]
\centering
\subfloat[\label{fig:IPD_reprImportance}]{
	\includegraphics[scale=0.075]{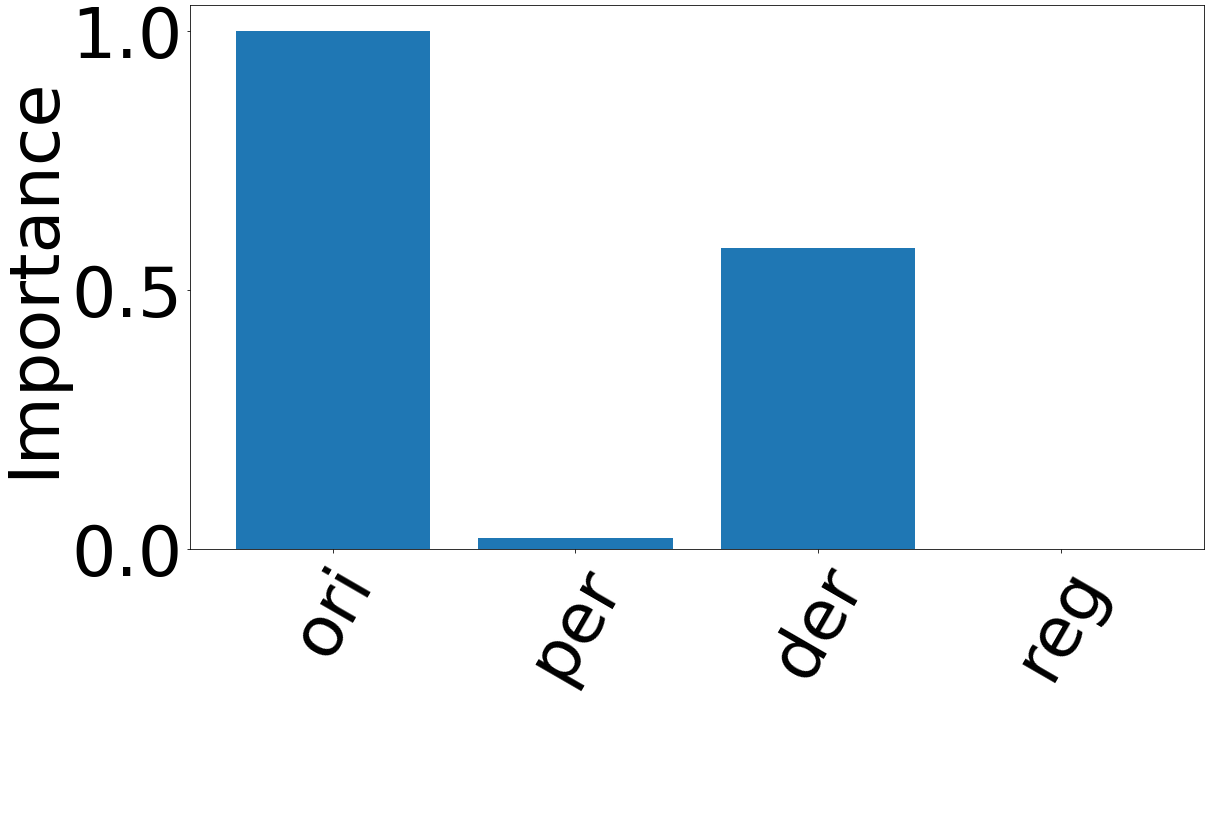}
}\hspace{10mm}
\subfloat[\label{fig:IPD_statsImportance}]{
	\includegraphics[scale=0.075]{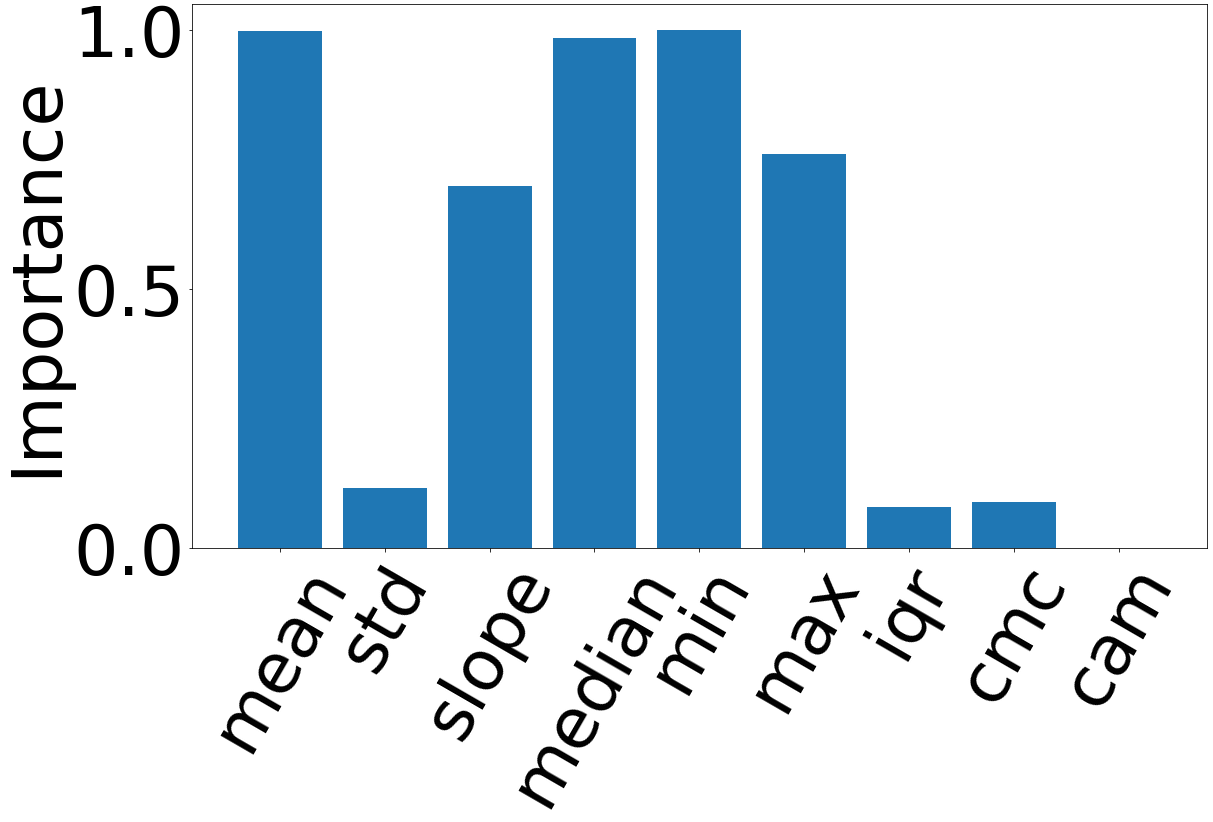}
}
\caption{\textbf{(a)} Importance of time series representations and \textbf{(b)} importance of aggregation functions when classifying the ItalyPowerDemand dataset with r-STSF. The original (ori) representation is the most important. The mean and min aggregation functions are the most important when extracting discriminatory intervals from the original representation of the dataset.}
\label{fig:IPD_repr_stats_importance}
\end{figure}

\subsection{Interpreting Classifications with the Original Time Series}

r-STSF uses different time series representations for classification. When the time series contain intervals with discriminatory time-stamped data (e.g., class A series have values above threshold $T$, whereas class B series have values below $T$), the original representation of the series is enough to achieve an accurate classification.
As a case study, we use the ItalyPowerDemand dataset~\citep{UCRArchive}, which consists of series of daily household energy consumption. When classifying ItalyPowerDemand with r-STSF, the original time series representation is the most important  (\figref{fig:IPD_reprImportance}). Besides, as shown in~\figref{fig:IPD_statsImportance}, in this representation, the mean and min aggregation functions contribute the most to the classification accuracy. We select the min aggregation to continue with our case (the mean shows similar results and is omitted for conciseness).

\begin{figure}[h]
\centering
\subfloat[\label{fig:IPD_interpretability_minOri}]{
	\includegraphics[scale=0.075]{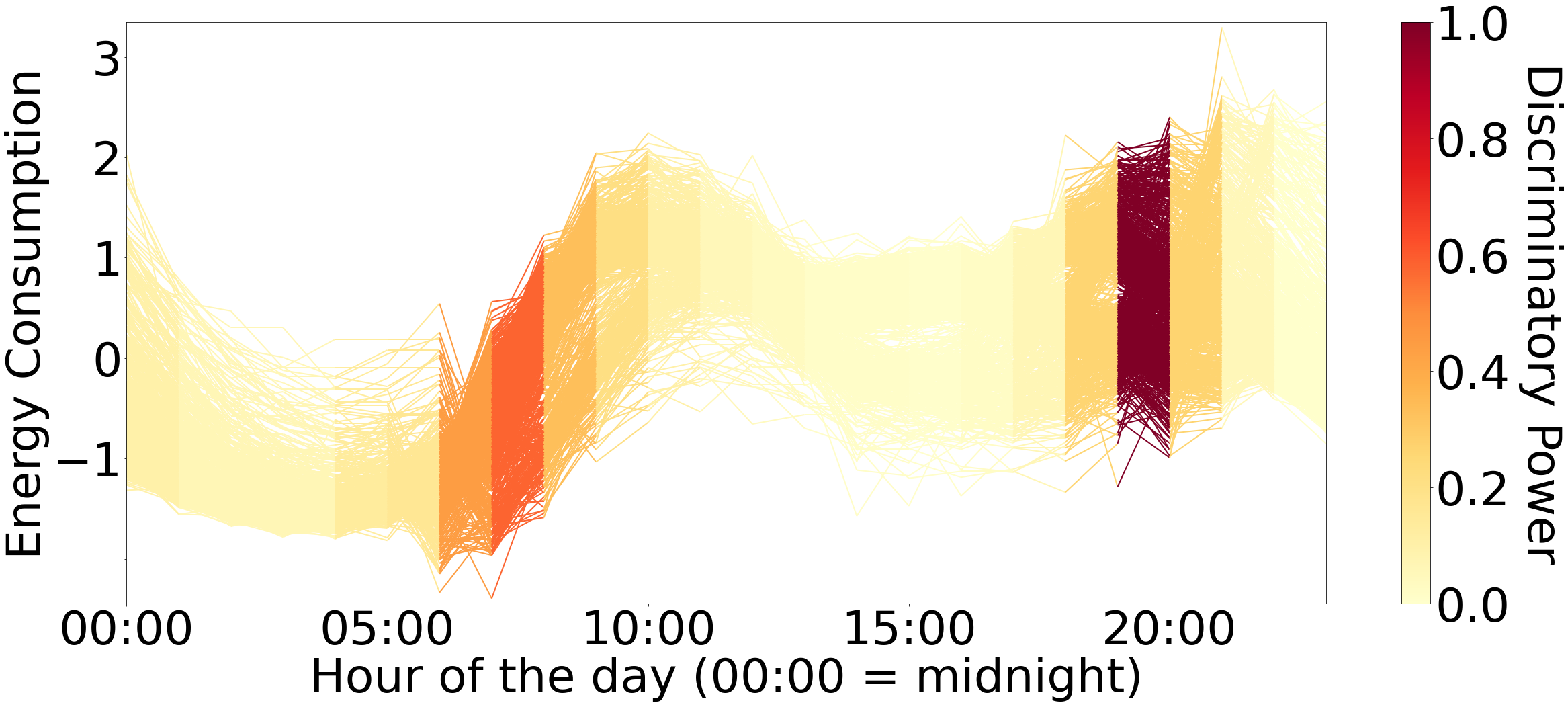}
}\hspace{1mm}
\subfloat[\label{fig:IPD_Ori_singleplot}]{
	\includegraphics[scale=0.075]{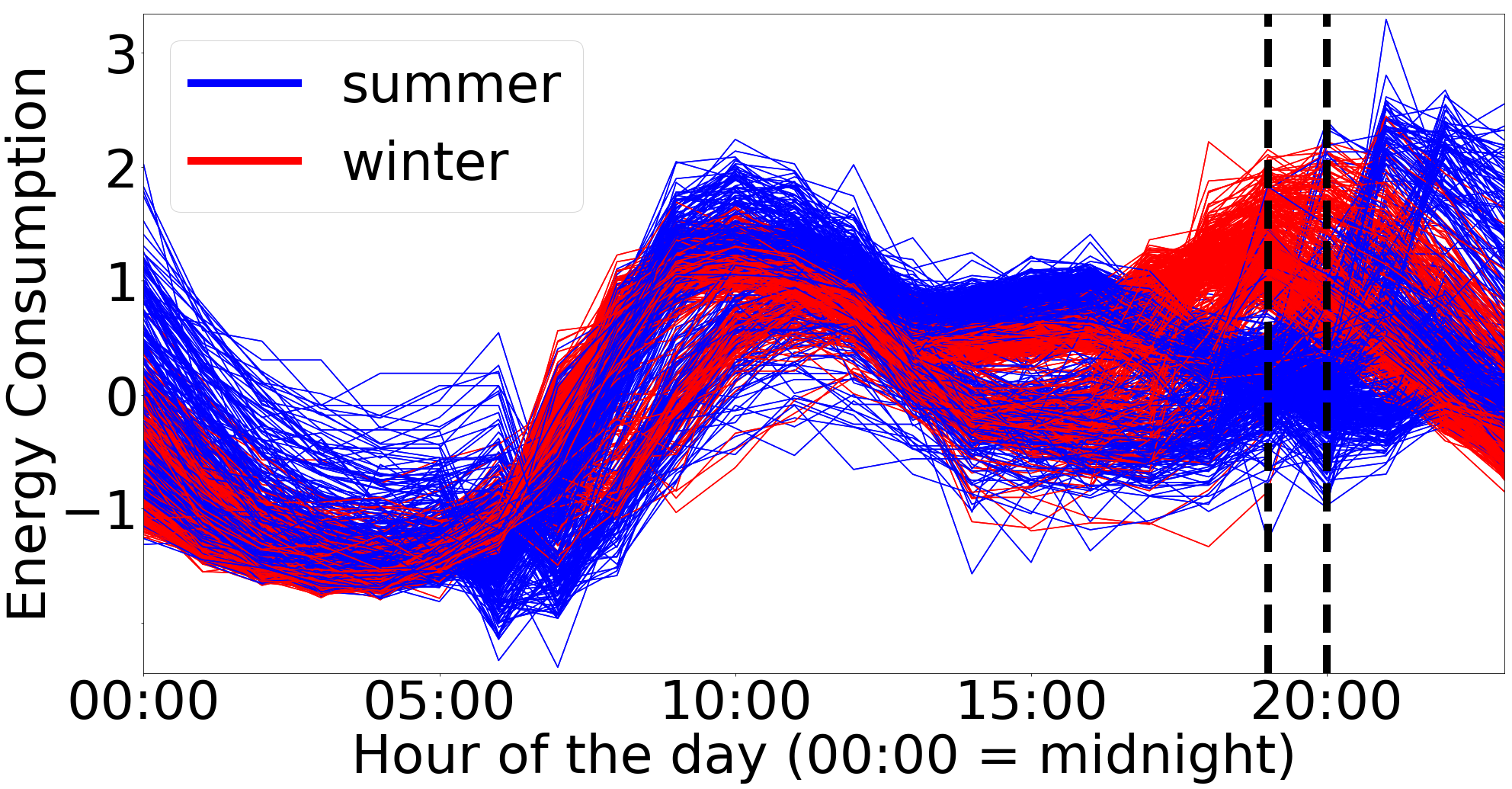}
}
\caption{Original representations of all time series in the ItalyPowerDemand dataset. \textbf{(a)}~Location (in time) of discriminatory intervals according to the min aggregation function. \textbf{(b)} The two types of series can be differentiated according to their minimum energy consumption between 19:00 and  20:00. Summer days are in blue color; winter days in red color}

\label{fig:IPD_interpr_Ori_singleplot}
\end{figure}

The most discriminatory interval (according to the min aggregation function) is located between 19:00 and  20:00 (\figref{fig:IPD_interpretability_minOri}). As shown in~\figref{fig:IPD_Ori_singleplot},  between 19:00 and 20:00 (highlighted between the pair of dashed black lines),  the majority of winter series (red color) are above the summer series (blue color).  This interval shows that, in winter, by 20:00, residents reach a peak in the energy consumption of their households appliances, whereas in summer, at that hour of the day, the energy consumption is smaller, \textcolor{black}{which can be explained by residents' tendency to spend earlier times indoors in winter due to shorter days and cooler temperatures.}

\subsection{Interpreting Classifications with the Periodogram Representation}

\begin{figure}[h]
    \centering
    \includegraphics[scale=0.075]{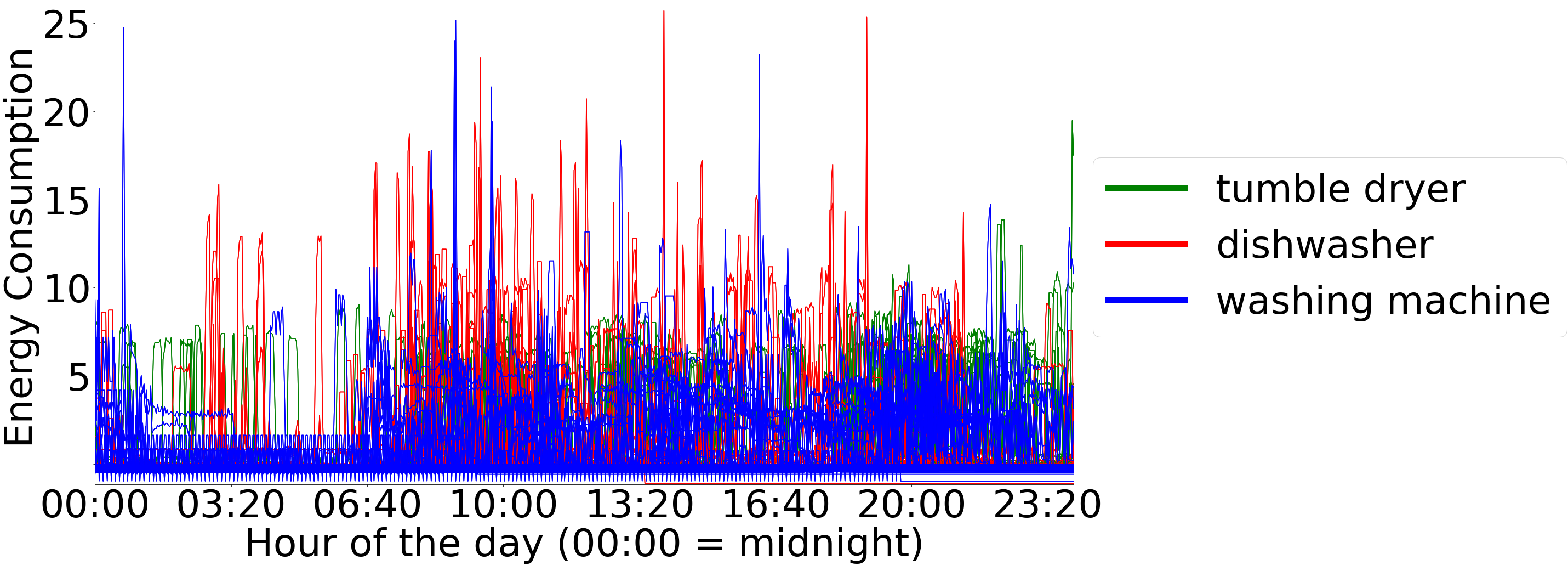}
    \caption{LargeKitchenAppliances (original) time series. It is difficult to identify discriminatory intervals in the original representation of the series. The few cases of high energy consumption for the dishwasher (red color) around 03:20 are outliers and hence are not discriminatory for r-STSF. Most of dishwasher series have (in that hour) small energy consumption values, similar to those of washing machine (blue color) and tumble dryer (green color) series.}
    \label{fig:LKA_originalTS}
\end{figure}

In datasets such as LargeKitchenAppliances~\citep{UCRArchive}  which contains energy consumption readings from three home appliances (dishwasher, tumble dryer, and washing machine), the variability in the time of using each appliance makes it difficult to identify discriminatory intervals in the original time series (\figref{fig:LKA_originalTS}). 

\begin{figure}[h]
\centering
\subfloat[\label{fig:LKA_reprImportance}]{
	\includegraphics[scale=0.075]{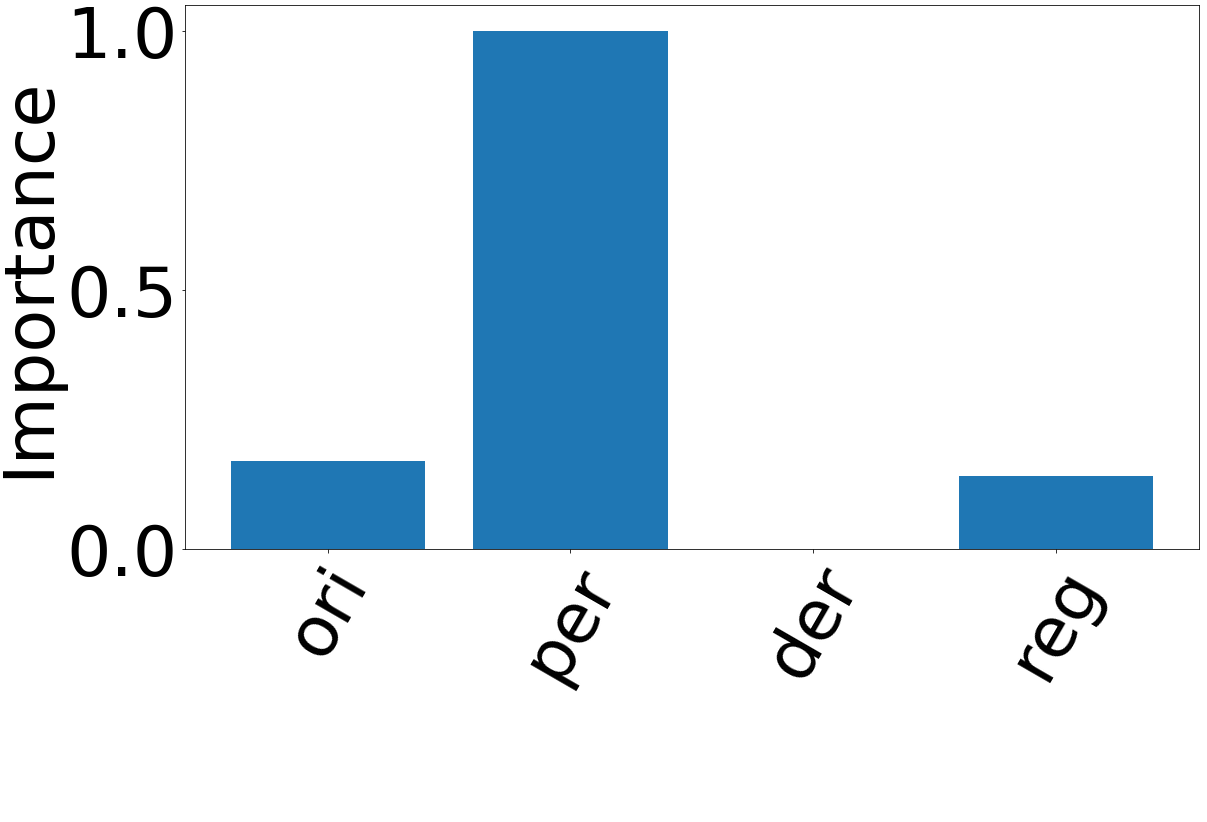}
}\hspace{10mm}
\subfloat[\label{fig:LKA_statsImportance}]{
	\includegraphics[scale=0.075]{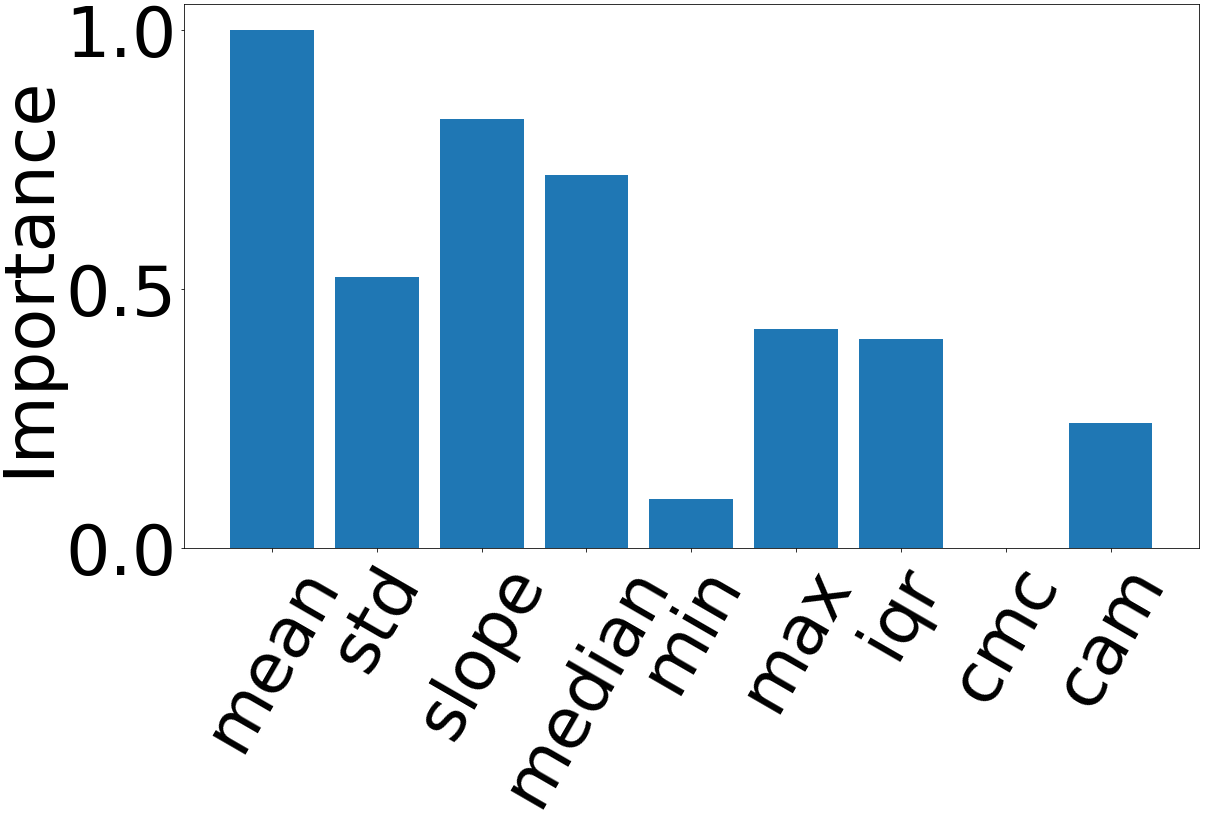}
}
\caption{\textbf{(a)} Importance of time series representations and \textbf{(b)} importance of aggregation functions when classifying the LargeKitchenAppliances dataset with r-STSF. The periodogram (per) representation is the most important representation when classifying this dataset. The mean aggregation function is the most important when extracting discriminatory intervals from the periodogram representation of LargeKitchenAppliances series}
\label{fig:LKA_repr_stats_importance}
\end{figure}

As shown in ~\figref{fig:LKA_reprImportance}, the original representation has a low importance in the classification of the  LargeKitchenAppliances dataset whereas the periodogram representation is much more important. Moreover, mean is the aggregation function that contributes the most to the classification accuracy when extracting features from the periodogram representation (\figref{fig:LKA_statsImportance}).

\begin{figure}[h]
\centering
\subfloat[\label{fig:LKA_interpretability_meanPer}]{
	\includegraphics[scale=0.075]{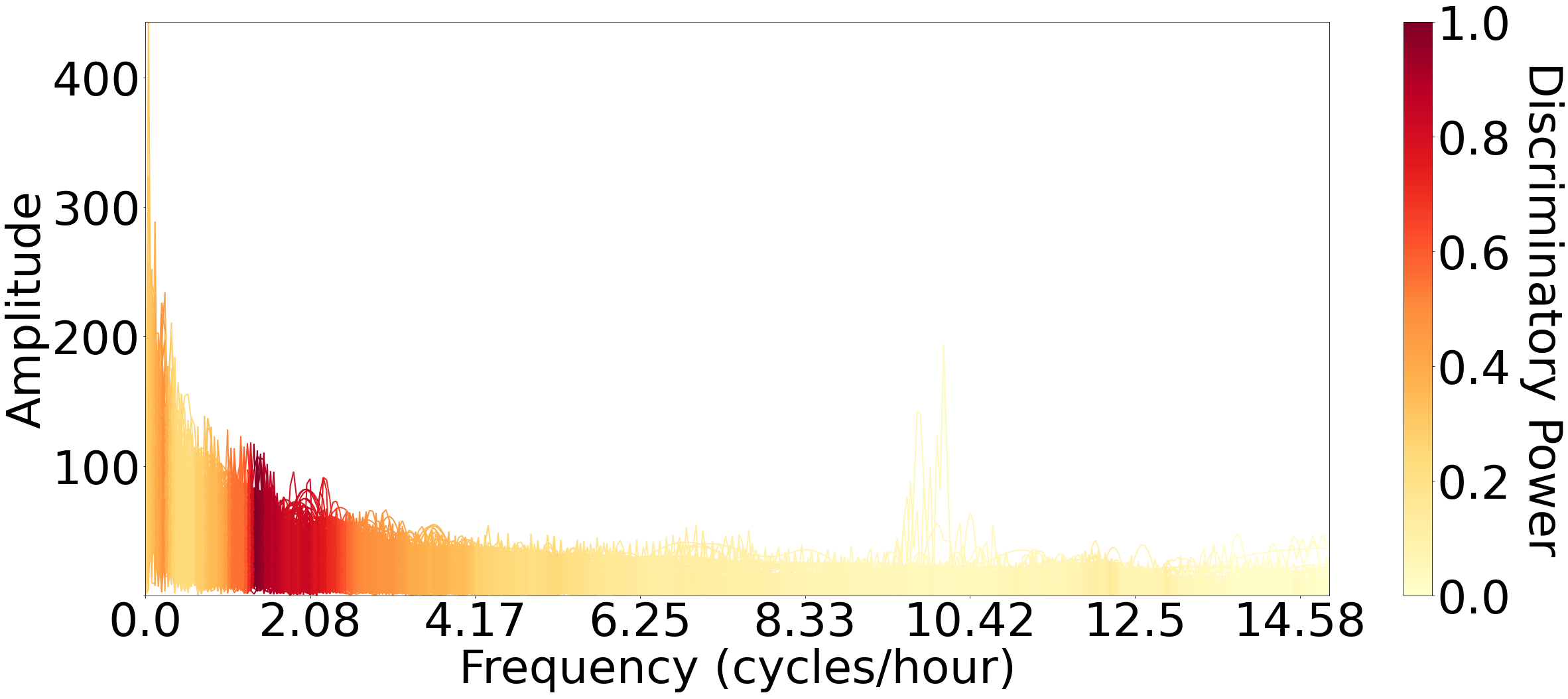}
}
\subfloat[\label{fig:LKA_Per_singleplot}]{
	\includegraphics[scale=0.075]{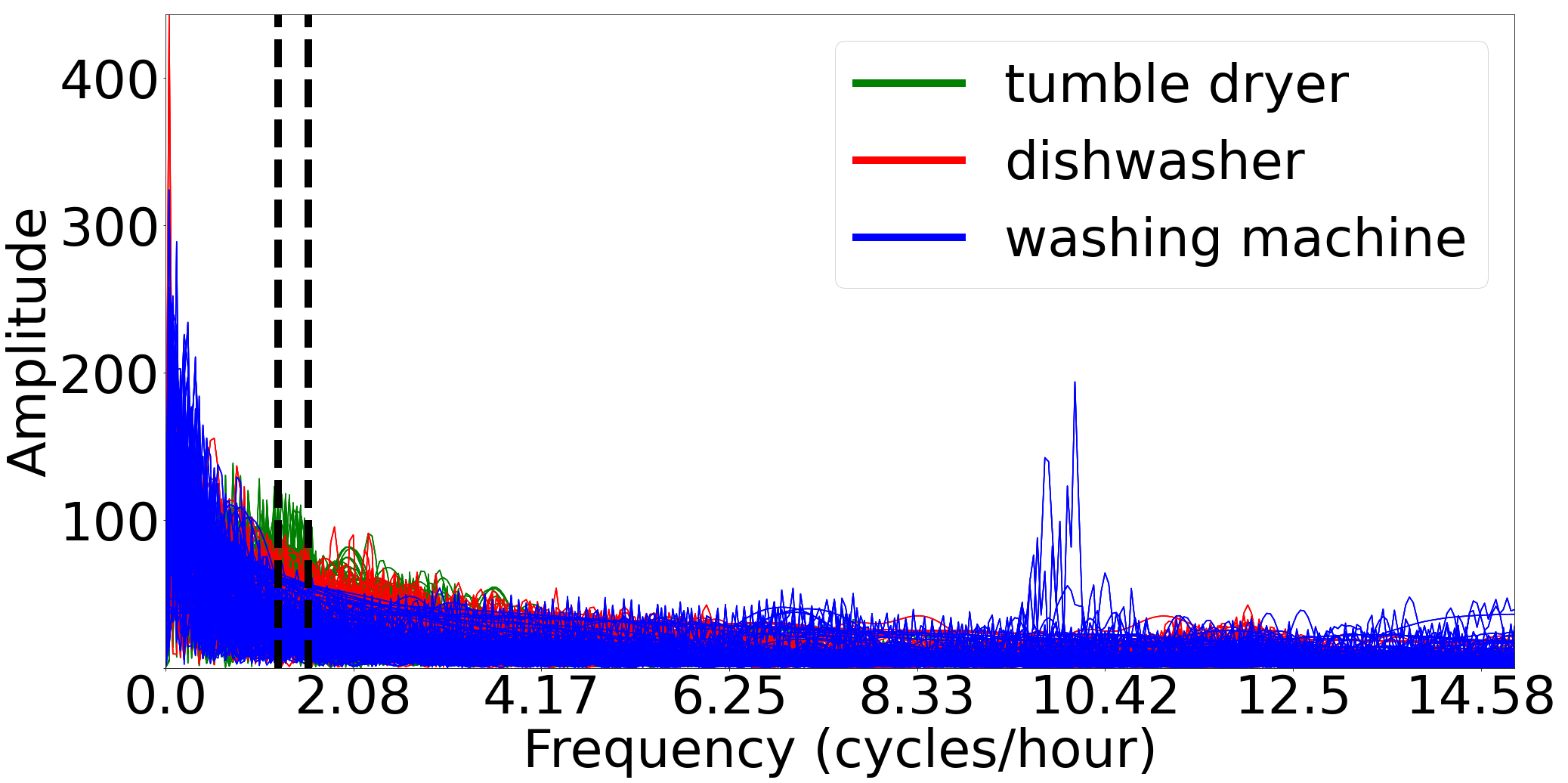}
}
\caption{Periodogram representations of all time series in LargeKitchenAppliances dataset. \textbf{(a)} Location (in the frequency-domain) of discriminatory intervals according to the mean aggregation function. \textbf{(b)} The three types of appliances can be differentiated according to their amplitude around the frequency of 1 cycle/hour. Tumble dryer series are in green color, dishwasher series in red color and washing machine series in blue color}
\label{fig:LKA_interpr_Per_singleplot}
\end{figure}

The periodogram representation shows how the signal's amplitude is distributed over a range of frequencies. The frequencies with higher amplitude reveal repeated patterns at such frequencies. As shown in~\figref{fig:LKA_interpretability_meanPer}, r-STSF uses the periodogram representation of LargeKitchenAppliances series to identify discriminatory intervals (using the mean aggregation function) located approximately at the frequency of 1 cycle/hour. From \figref{fig:LKA_Per_singleplot}, we see that, around this frequency (highlighted between the pair of dashed black lines),  there is a difference in the amplitude of the three types of appliances.
The energy consumption signal of each appliance belongs to one operation cycle. 
The discriminatory frequency of 1 cycle/hour reveals that all three appliances have the most different energy consumption around the one-hour mark on average while running these appliances. The higher amplitude of this frequency for the tumble dryer than for the other two appliances suggests that householders use the tumble dryer more often during the day (e.g., one hour in the morning and one hour in the evening) or that it takes more than one hour to complete the drying process (when overloaded, the tumble dryer usually requires two hours). Similarly, this discriminatory frequency also suggests that householders use the dishwasher more often than the washing machine.

\subsection{Interpreting Classifications with the First-order Derivative Representation}

\begin{figure}[h]
    \centering
    \includegraphics[scale=0.075]{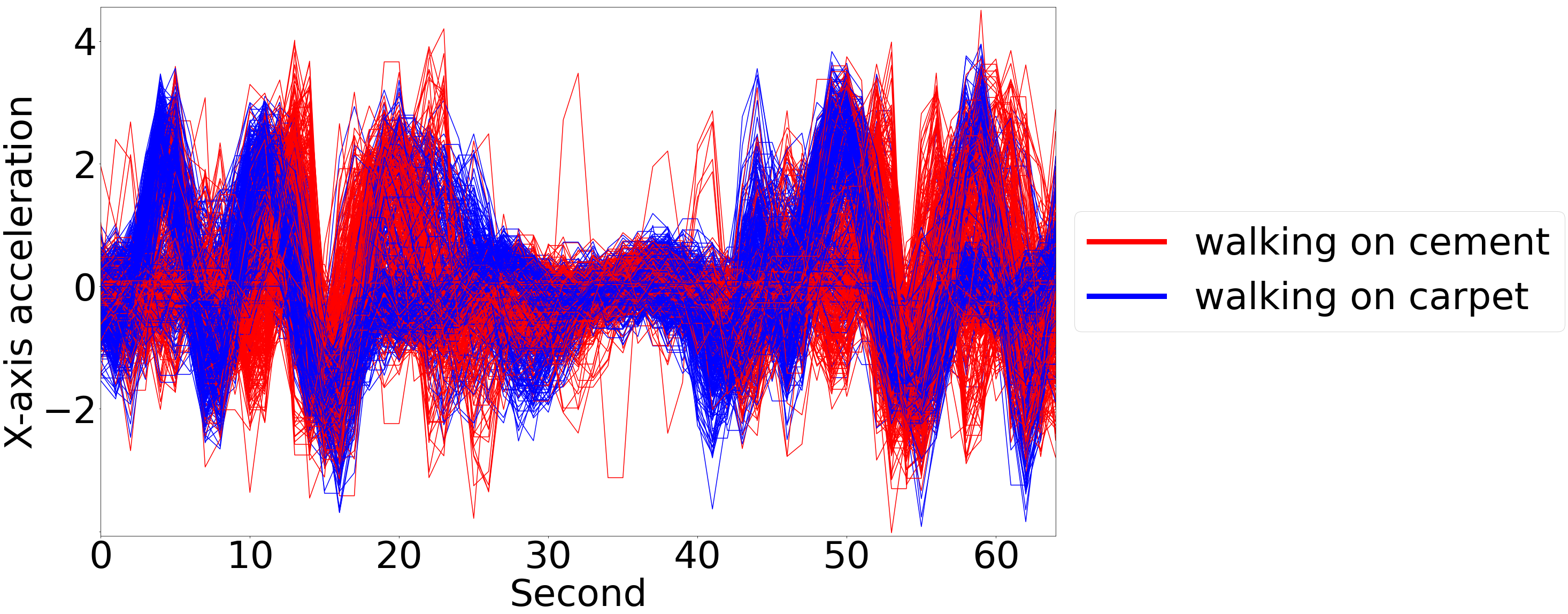}
    \caption{SonyAIBORobotSurface2 (original) time series. Potential discriminatory
    intervals are located in second 10 and in second 50. At both locations, walking on cement (red color) and walking on carpet (blue color) signals are out-of-phase}
    % \textcolor{red}{10 and 50.}}
    \label{fig:SONY_originalTS}
\end{figure}

\begin{figure}[h]
\centering
\subfloat[\label{fig:SONY_reprImportance}]{
	\includegraphics[scale=0.075]{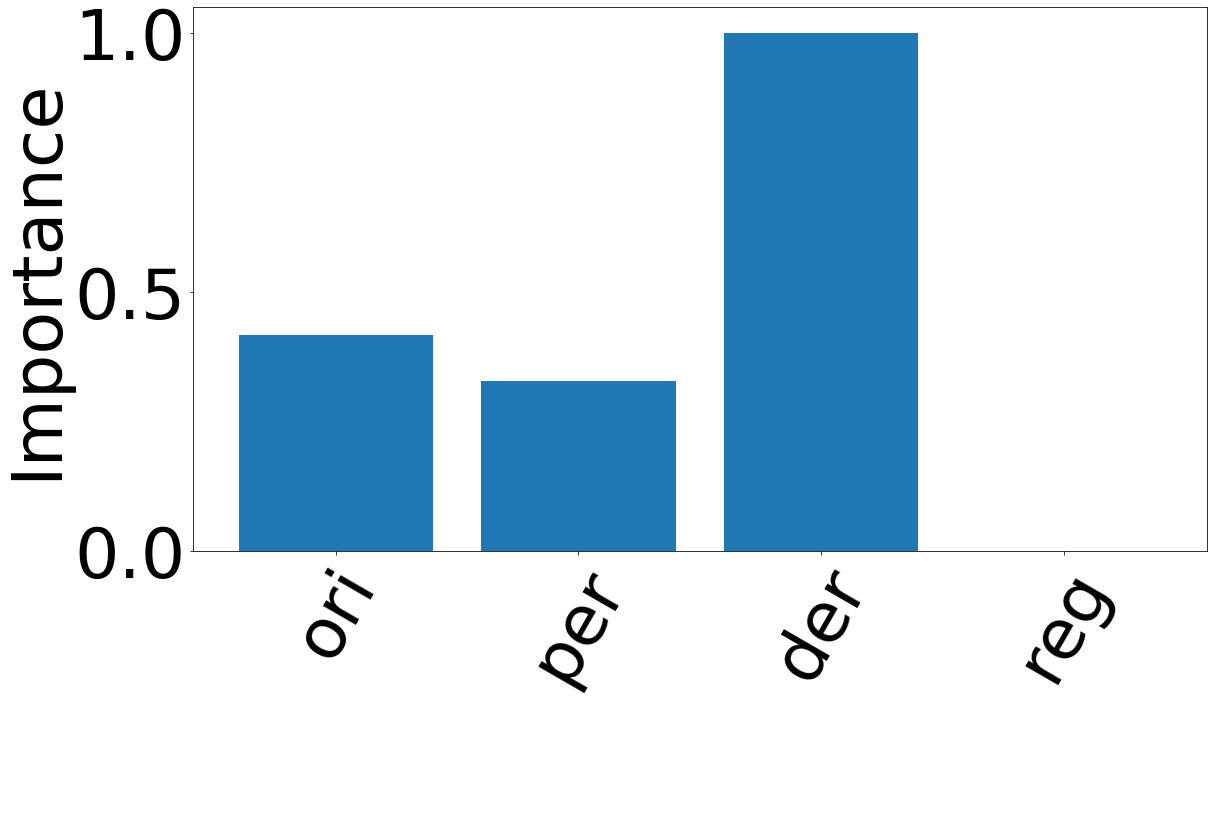}
}\hspace{10mm}
\subfloat[\label{fig:SONY_statsImportance}]{
	\includegraphics[scale=0.075]{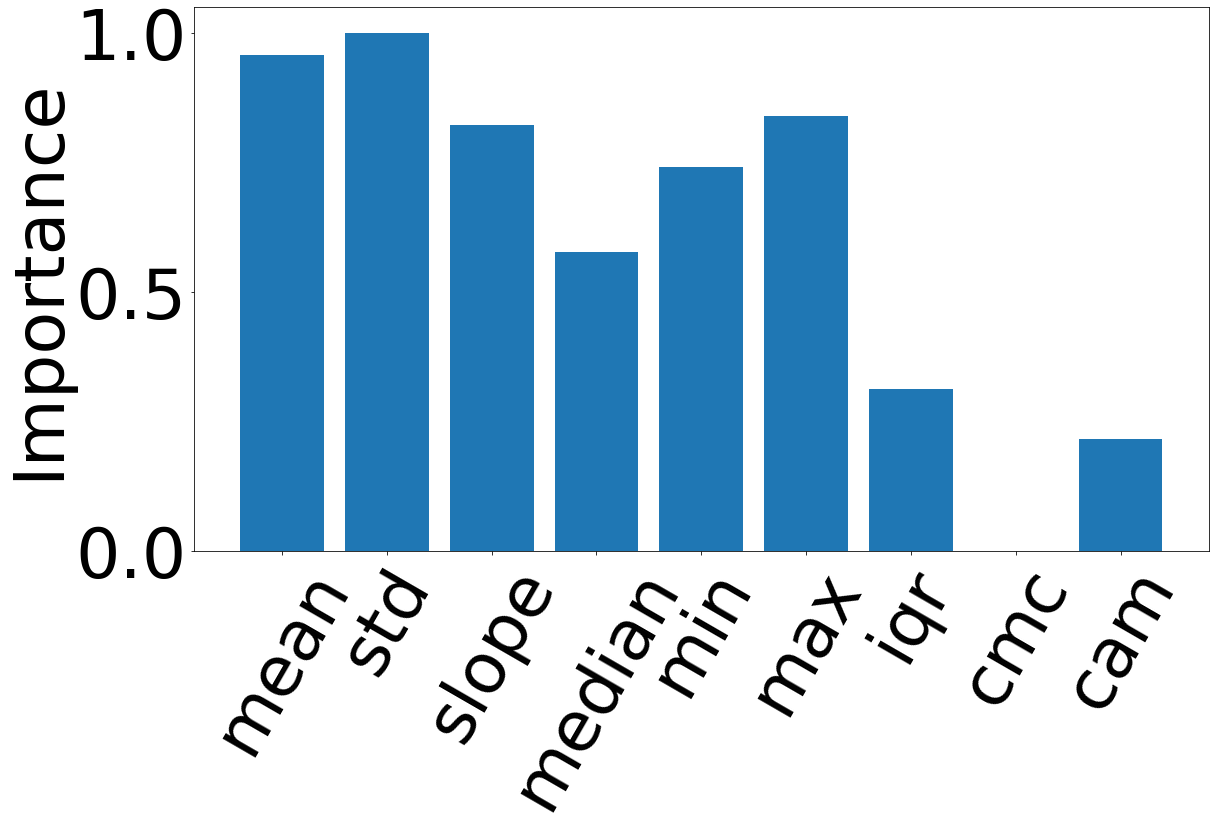}
}
\caption{\textbf{(a)} Importance of time series representations and \textbf{(b)} importance of aggregation functions when classifying the SonyAIBORobotSurface2 dataset with r-STSF. The derivative (der) representation is the most important representation, and std is the most important aggregation function when extracting discriminatory intervals from the derivative representation}
\label{fig:LKA_repr_stats_importance}
\end{figure}

To demonstrate how r-STSF allows for interpretable classifications in the derivative representation, we discuss the case of SonyAIBORobotSurface2 dataset~\citep{UCRArchive} in this section. This dataset contains readings from an X-axis accelerometer attached to a quadruped Sony AIBO dog robot while walking on two different surfaces: cement and carpet. Each time series has a length of 65 data points (we assume 65 seconds of walking). The classification task is to detect the surface being walked on. Although an inspection to the original series of this dataset may lead to potential discriminatory intervals (\figref{fig:SONY_originalTS}), the derivative representation has a much more important role when classifying the series (\figref{fig:SONY_reprImportance}).

The derivative representation is useful when time series can be differentiated by their trends or changes in the series values. The derivative representation is the first-order difference of a time series, and hence, every data point of this representation can be mapped back to a pair of original data points (i.e., interval) of the original representation. As stated by~\cite{kertesz2014exploring}: ``When a robot walks on a rigid surface, it produces vertical body oscillations while soft surfaces absorb these anomalies". Hence, walking on cement increases the variability (or rate of change) of the X-axis acceleration readings. The derivative representation is thus more useful to classify the SonyAIBORobotSurface2 series. Similarly,  the std aggregation function, which measures the spread of the values, is also important to detect discriminatory interval in this representation (\figref{fig:SONY_statsImportance}).

\begin{figure}[h]
\centering
\subfloat[\label{fig:SONY_interpretability_stdDer}]{
	\includegraphics[scale=0.075]{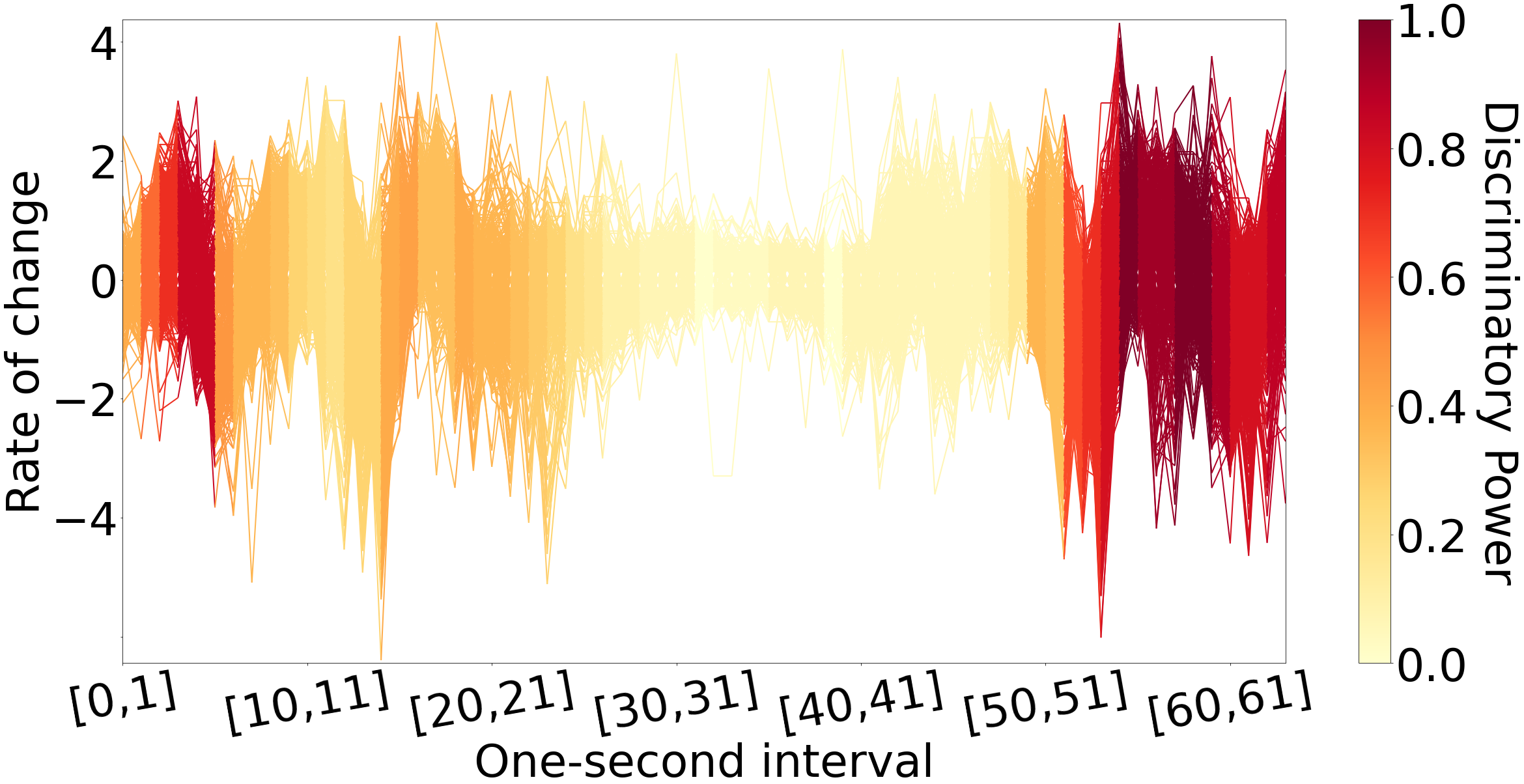}
}
\subfloat[\label{fig:SONY_Der_singleplot}]{
	\includegraphics[scale=0.075]{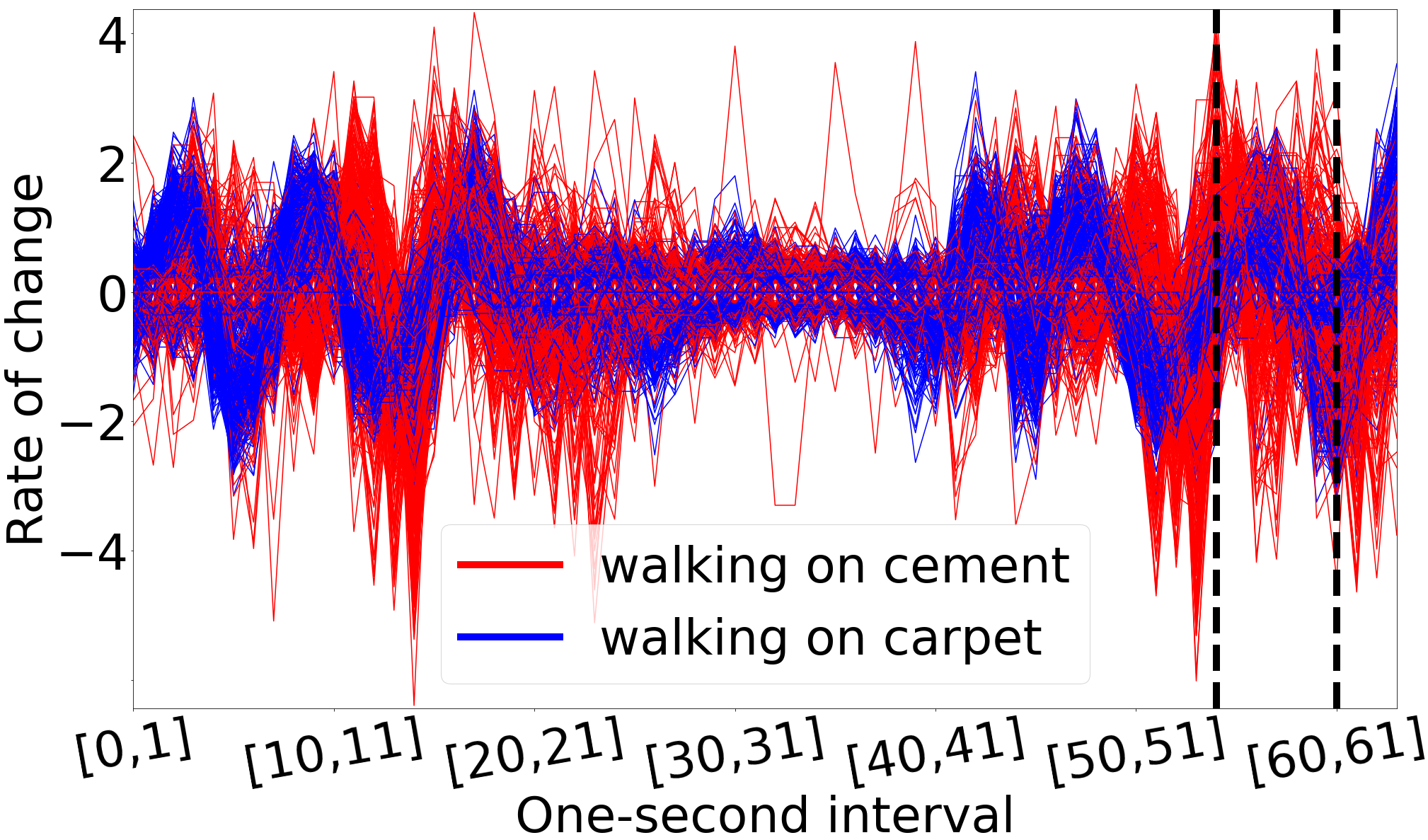}
}
\caption{Derivative representations of all time series in the SonyAIBORobotSurface2 dataset. \textbf{(a)} Location of discriminatory intervals according to  the std aggregation function. \textbf{(b)} The two types of surfaces can be differentiated for the high variability of the AIBO robot acceleration when walking on cement in the time interval between second 55 and second 60 of the original time series. Walking on cement series are in red color; walking on carpet series in blue color}
\label{fig:SONY_interpr_Der_singleplot}
\end{figure}

As shown in~\figref{fig:SONY_interpretability_stdDer}, std detects discriminatory intervals at the beginning (first 5 seconds of walking) and at the end (last 10 seconds of walking) in the derivative representations. A higher difference in the variability of the acceleration values between walking on cement and on carpet can be found approximately in the time interval between seconds 55 to 60 (highlighted between the pair of dashed black lines in ~\figref{fig:SONY_Der_singleplot}). 
We have no further information of the data collection process for the  SonyAIBORobotSurface2 dataset, which limits our explanations for the higher variability (of walking on cement) at the beginning and the end of the walking sessions. One interpretation could be that due to the inertia (i.e., tendency of a body to resist a change in motion or rest), the acceleration varies when the robot starts its movement and also when it is about the stop.

\subsection{Interpreting Classifications with the Autoregressive Representation}

\begin{figure}[h]
    \centering
    \includegraphics[scale=0.075]{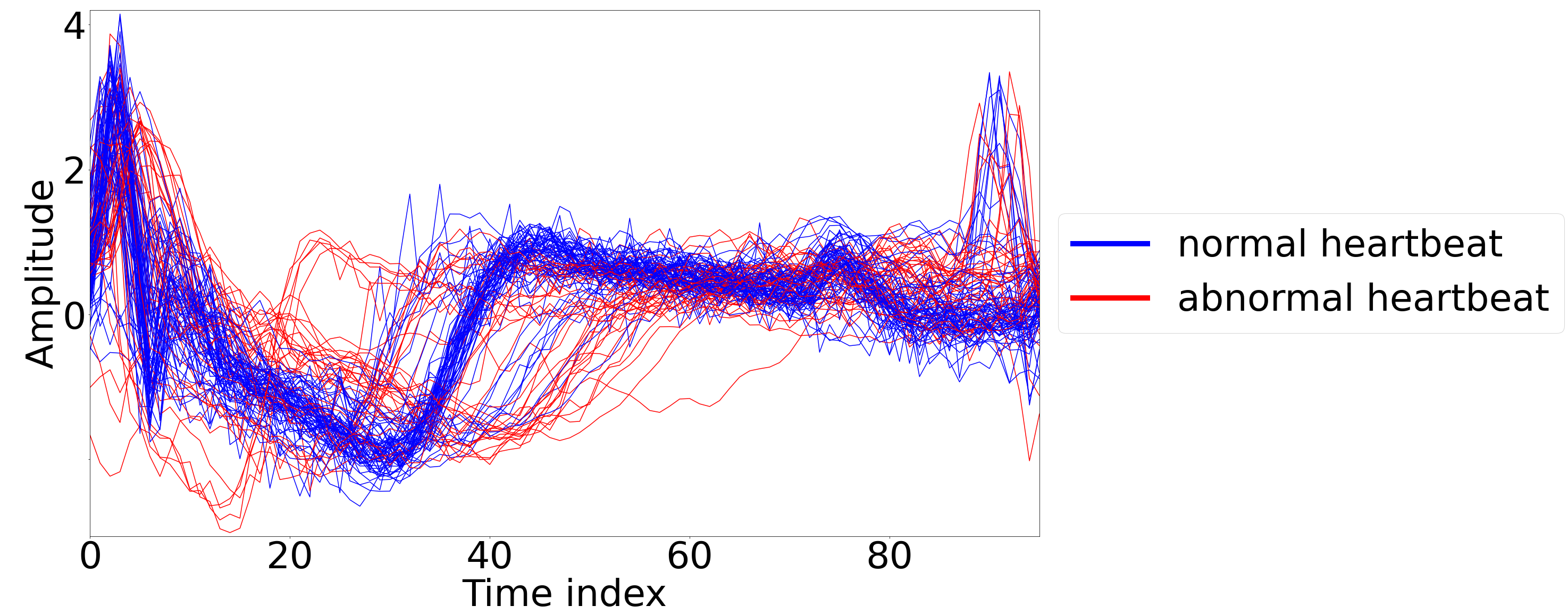}
    \caption{ECG200 (original) time series. There is not a specific interval where normal (blue color) and abnormal (red color) series are easily differentiable}
    \label{fig:ECG_originalTS}
\end{figure}

We use the ECG200 dataset~\citep{UCRArchive, olszewski2001generalized} to show how r-STSF uses the autoregressive representation for classification and how to interpret such classification. Each series in ECG200 contains the measurements recorded by one electrode during one heartbeat. The heartbeats are labeled as normal and abnormal. The original representation of ECG200 series (\figref{fig:ECG_originalTS}) may not be the most informative when aiming to capture discriminatory intervals. Normal and abnormal series follow a similar pattern in time, which makes it difficult to identify discriminatory intervals.

\begin{figure}[h]
\centering
\subfloat[\label{fig:ECG_reprImportance}]{
	\includegraphics[scale=0.075]{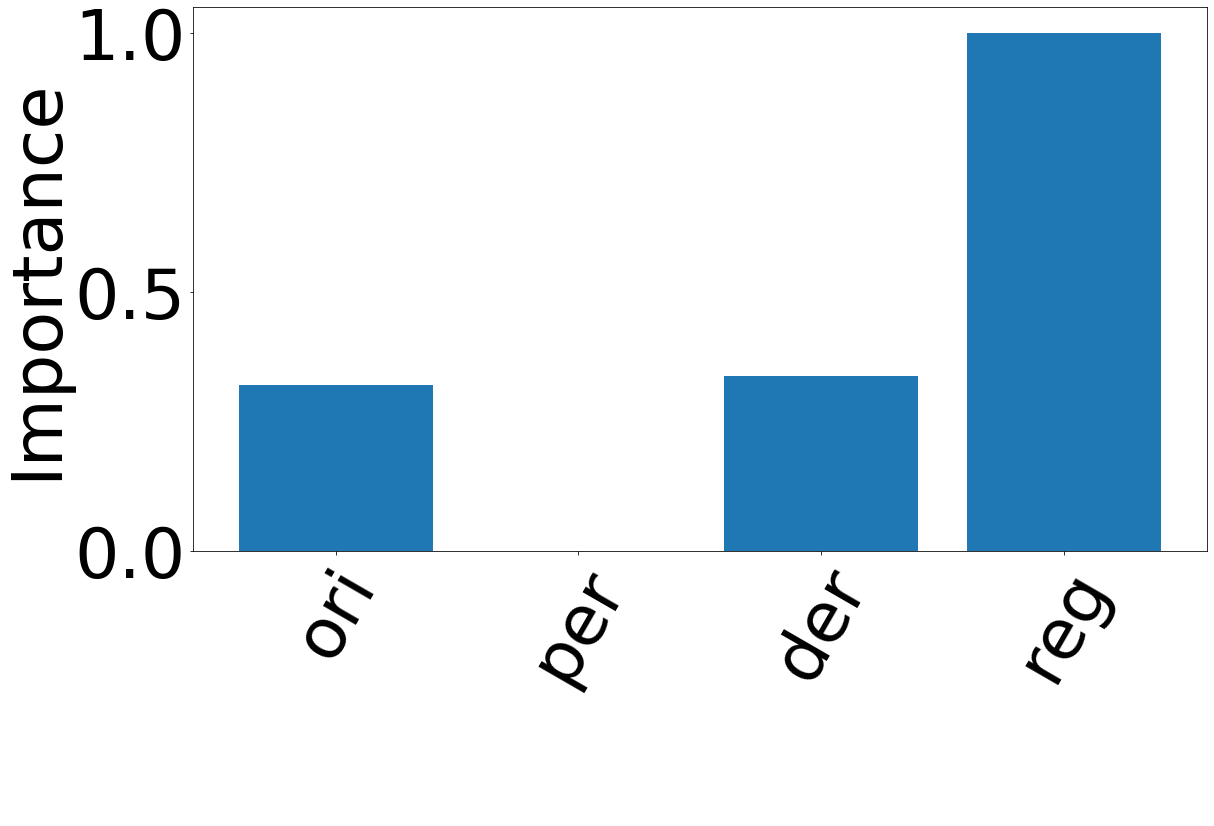}
}\hspace{10mm}
\subfloat[\label{fig:ECG_statsImportance}]{
	\includegraphics[scale=0.075]{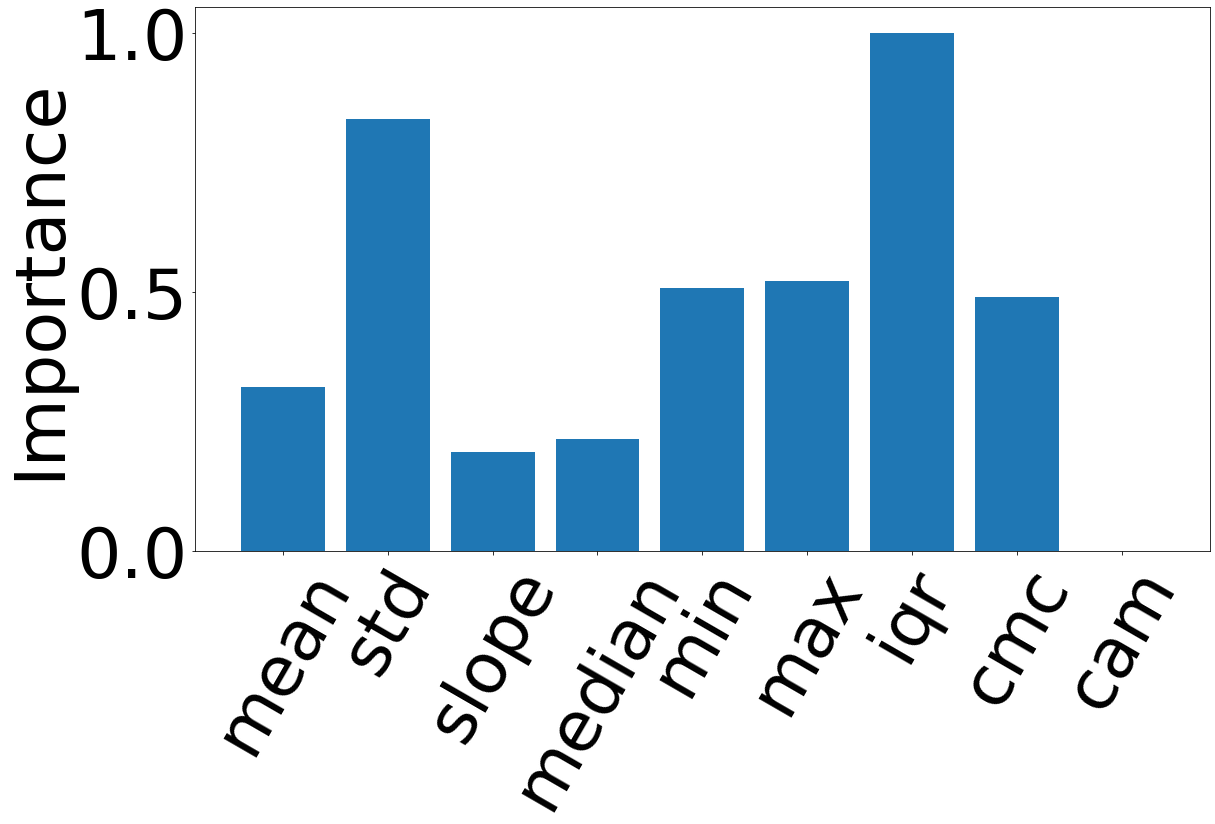}
}
\caption{\textbf{(a)} Importance of time series representations and \textbf{(b)} importance of aggregation functions when classifying the ECG200 dataset with r-STSF. The autoregressive (reg) representation is the most important representation. The iqr aggregation function is the most important aggregation function when extracting discriminatory intervals from the derivative representation}
\label{fig:ECG_repr_stats_importance}
\end{figure}

As shown in~\figref{fig:ECG_reprImportance}, the autoregressive representation is identified as the most important representation by r-STSF. Moreover, std and iqr, which capture dispersion or variability of the data points, extract the most discriminatory intervals in the autoregressive representation (\figref{fig:ECG_statsImportance}).
An autoregressive (AR) model holds the premise that past values have an effect on current values. For example, in an AR process with lag order of 1 (denoted by ``AR(1)''), the current value is predicted based on the immediate preceding value using a linear model. 
The coefficients of such predictors are the AR coefficients, which can be used as features for time series classification~\citep{lines2018time}. r-STSF does not use the AR coefficients directly as features, but it uses them as another representation of the original time series, i.e., AR representation. Thus, r-STSF extracts discriminatory interval features from the AR representation of the time series.

\begin{figure}[h]
\centering
\subfloat[\label{fig:ECG_interpretability_iqrReg}]{
	\includegraphics[scale=0.075]{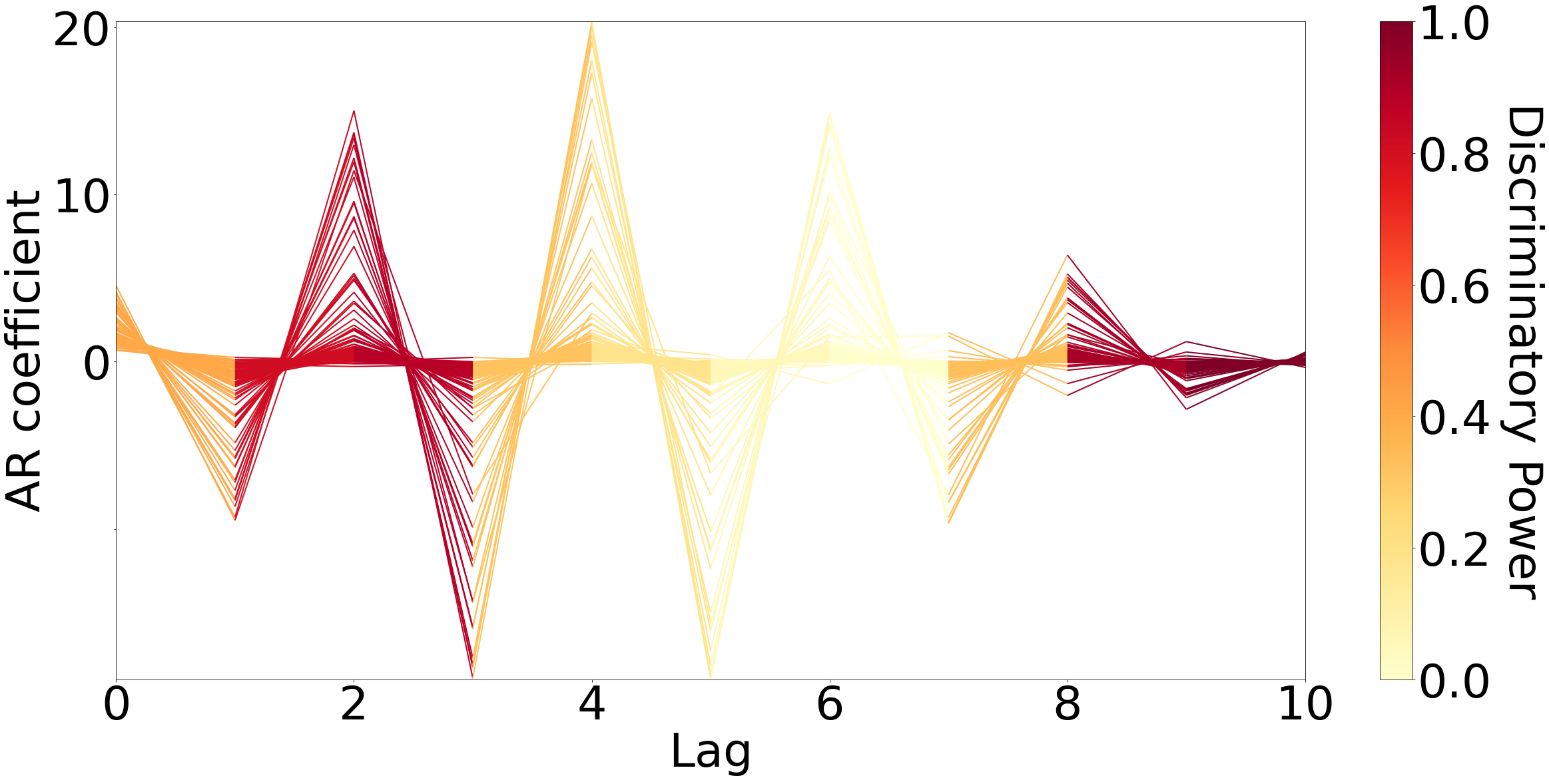}
}
\subfloat[\label{fig:ECG_Reg_singleplot}]{
	\includegraphics[scale=0.075]{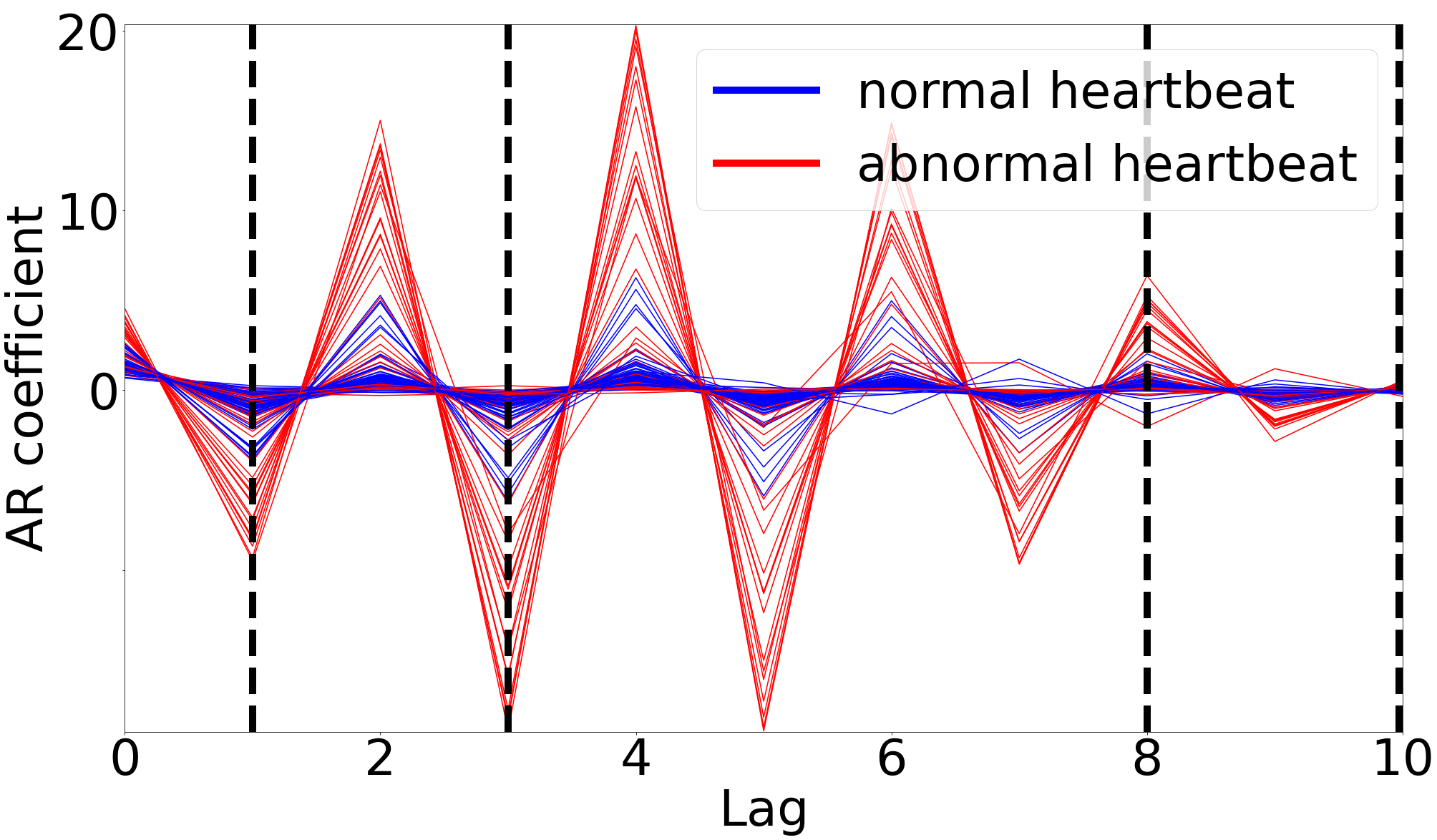}
}
\caption{Autoregressive representations of all time series in the ECG200 dataset. \textbf{(a)} Location of discriminatory intervals according to the iqr aggregation function. \textbf{(b)} The variability of the AR coefficients of the abnormal heartbeats (red color) compared to those of the normal heartbeats (blue color) is higher between lags 1 and 3 and between lags 8 and 10}
\label{fig:ECG_interpr_Reg_singleplot}
\end{figure}

As shown in~\figref{fig:ECG_interpretability_iqrReg}, the most discriminatory interval features are located between lags 1 to 3 and also between lags 8 to 10. The variability of the AR coefficients of series from abnormal heartbeats is much higher than the variability of AR coefficients from series of normal heartbeats between these lags. r-STSF uses this difference in the variability to differentiate between normal and abnormal heartbeats.

\begin{figure}[htb]
  \centering
    \includegraphics[scale=0.25]{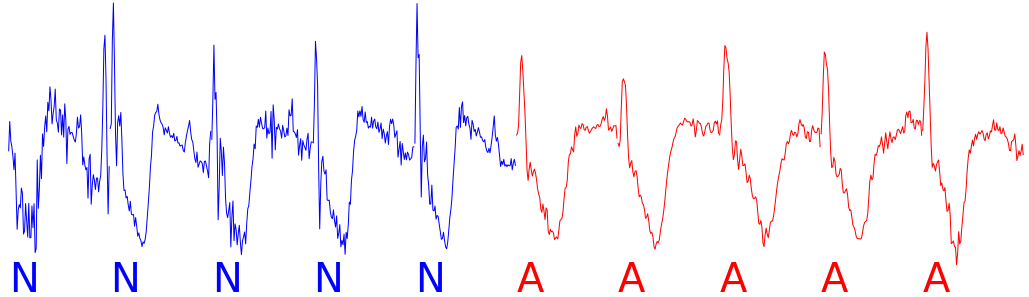}
    \caption{Example of ten ECG200 series. N: normal heartbeats (blue color), A: abnormal heartbeats (red color)}
\label{fig:ECG200dataset}
\end{figure}

The AR coefficients of abnormal heartbeats are usually larger than those of normal heartbeats (\figref{fig:ECG_interpr_Reg_singleplot}). AR coefficients do not provide specific information on the relationship of the variables, i.e., AR coefficients cannot tell to which extent current and past value are correlated. Nonetheless, a high AR coefficient implies that past values have some effect on current values whereas a low AR coefficient implies a small or no effect. Hence, in the ECG200 dataset, past values from abnormal series have some effect in current values. In the contrary, in normal heartbeats there is a small or no effect of past values in current ones. As shown in~\figref{fig:ECG200dataset} normal heartbeats are more irregular or noisy than abnormal heartbeats. Therefore, for the case of normal heartbeats it is difficult to establish an effect of past values on current ones.

Further, from \figref{fig:ECG_interpr_Reg_singleplot}, we can infer that the effect of past 2, 3, 8, and 9 values on current ones is much higher in abnormal heartbeats than in normal heartbeats. Although for abnormal heartbeats there are also high AR coefficients at lag 4 and 5; the iqr aggregation function considers such values as extreme or outliers, thus the interval between such lags is not considered as discriminatory.

\color{black}

\section{Conclusions and Future Work} 

We proposed r-STSF, a highly efficient interval-based algorithm for time series classification. r-STSF is the only TSC method that achieves SOTA classification accuracy and allows for interpretable classifications. To achieve competitive classification accuracies, r-STSF builds an ensemble of randomized trees for classification. It uses four time series representations, nine aggregation functions, and a supervised search strategy combined with a feature ranking metric when searching for highly discriminatory sets of interval features. The discriminatory interval features enable interpretable classification results. r-STSF not only allows for interpretations in the original (time-stamped) time series, but also in the periodogram, derivative, and autoregressive representation of the time series. Extensive experiments on real-world datasets validate the accuracy and efficiency of our proposed method -- r-STSF is as accurate as SOTA TSC methods but orders of magnitude faster, enabling it to classify large datasets with long series. 
Further, r-STSF  outperforms SOTA TSC methods in terms of weighted average accuracy, demonstrating its robustness in  classify more complex datasets. 

While randomized trees have shown to improve the classification accuracy (when compared to non-randomized trees) for a large number of datasets, they are less likely to identify relevant features in datasets with a small number of relevant features as they might miss those. As future work, we plan to make r-STSF adaptive. When a dataset is expected to have a high percentage of relevant features -- estimated with techniques such as \emph{permutation importance}~\citep{louppe2013understanding} -- r-STSF trains an ensemble of randomized trees for classification. Otherwise, r-STSF uses non-randomized trees. Moreover, the intrinsic multivariate nature of time series signals (e.g., 3-axis accelerometer) leads us to extend r-STSF towards multivariate or multidimensional scenarios. Other interpretable methods such as CIF extend to multivariate scenarios by searching for discriminatory interval features in different dimensions of a series. However, extracting interval features per dimension (i) could potentially miss discriminatory features only found when data from every dimension is combined, e.g., an interval from series of the $x$, $y$, and $z$-axis of the accelerometer may not be discriminatory when assessed by separate, but it is discriminatory when data from each axis is combined, and (ii) may hinder the interpretability when classifying datasets with a high number of dimensions due to such interpretability has to be done per dimension (by highlighting discriminatory regions in the series from each dimension). We plan to transform the individual (per dimension) time series into a single (unified) representation from where to extract discriminatory interval features. More importantly, this single representation will allow r-STSF to provide ``readable" interpretable classifications.

\begin{acknowledgements}
This research was partially supported under Australian Research Council's Discovery Projects funding scheme (project number DP170102472).
\end{acknowledgements}

% BibTeX users please use one of

% \bibliographystyle{spmpsci}      % mathematics and physical sciences
% \bibliographystyle{spphys}       % APS-like style for physics
\bibliography{main}   % name your BibTeX data base
\bibliographystyle{spbasic}      % basic style, author-year citations

% Non-BibTeX users please use
% \begin{thebibliography}{}
% %
% % and use \bibitem to create references. Consult the Instructions
% % for authors for reference list style.
% %
% \bibitem{RefJ}
% % Format for Journal Reference
% Author, Article title, Journal, Volume, page numbers (year)
% % Format for books
% \bibitem{RefB}
% Author, Book title, page numbers. Publisher, place (year)
% % etc
% \end{thebibliography}
\clearpage
% \appendixpagename
% \appendixtitleon
% \appendixtitletocon
\begin{appendices}
\section{Results on 85 UCR benchmark time series datasets}
\label{appendix:85dsets_appendix}

\setlength{\tabcolsep}{1pt}
\small
\begin{longtable}[c]{lcccccccccc}
\caption{\textcolor{black}{Average classification accuracy over 10 runs of r-STSF for each dataset of $UCR_{85}$. Other methods are as reported in their papers or accompanying websites}}
\label{table:comparison}\\
\hline
\textbf{Datasets} &
  \textbf{STSF} &
  \textbf{TSF} &
  \textbf{BOSS} &
  \textbf{ResNet} &
  \textbf{HCOTE} &
  \textbf{PF} &
  \textbf{CHIEF} &
  \textbf{ITime} &
  \textbf{ROCKET} &
  \textbf{r-STSF} \\ \hline
\endfirsthead
\multicolumn{11}{c}%
{{\bfseries Table \thetable\ continued from previous page}} \\
\hline
\textbf{Datasets} &
  \textbf{STSF} &
  \textbf{TSF} &
  \textbf{BOSS} &
  \textbf{ResNet} &
  \textbf{HCOTE} &
  \textbf{PF} &
  \textbf{CHIEF} &
  \textbf{ITime} &
  \textbf{ROCKET} &
  \textbf{\textcolor{black}{r-STSF}} \\ \hline
\endhead
\hline
\endfoot
\endlastfoot
\textbf{\underline{Adiac}} &
  82.79 &
  76.32 &
  76.47 &
  82.90 &
  81.07 &
  73.40 &
  79.80 &
  82.97 &
  78.47 &
  \textbf{83.55} \\
\textbf{ArrowHead} &
  67.49 &
  72.57 &
  83.43 &
  84.50 &
  86.29 &
  \textbf{87.54} &
  83.27 &
  84.69 &
  80.51 &
  74.05 \\
\textbf{\underline{Beef}} &
  84.00 &
  86.33 &
  80.00 &
  75.30 &
  93.33 &
  72.00 &
  70.61 &
  68.67 &
  83.33 &
  \textbf{87.00} \\
\textbf{BeetleFly} &
  94.00 &
  75.00 &
  90.00 &
  85.00 &
  \textbf{95.00} &
  87.50 &
  91.36 &
  80.00 &
  90.00 &
  92.00 \\
\textbf{BirdChicken} &
  90.00 &
  80.00 &
  \textbf{95.00} &
  88.50 &
  85.00 &
  86.50 &
  90.91 &
  \textbf{95.00} &
  90.00 &
  90.50 \\
\textbf{Car} &
  81.50 &
  76.67 &
  83.33 &
  \textbf{92.50} &
  86.67 &
  84.67 &
  85.45 &
  89.00 &
  89.17 &
  87.50 \\
\textbf{\underline{CBF}} &
  97.90 &
  97.27 &
  99.78 &
  99.50 &
  99.89 &
  99.33 &
  99.79 &
  99.78 &
  \textbf{99.99} &
  99.17 \\
\textbf{\underline{ChlCon}} &
  78.04 &
  74.93 &
  66.09 &
  84.40 &
  71.20 &
  63.39 &
  71.67 &
  \textbf{87.28} &
  81.30 &
  77.82 \\
\textbf{\underline{CinCECGTorso}} &
  98.49 &
  95.07 &
  88.70 &
  82.60 &
  99.64 &
  93.43 &
  98.32 &
  84.22 &
  83.49 &
  \textbf{99.93} \\
\textbf{\underline{Coffee}} &
  \textbf{100.00} &
  \textbf{100.00} &
  \textbf{100.00} &
  \textbf{100.00} &
  \textbf{100.00} &
  \textbf{100.00} &
  \textbf{100.00} &
  \textbf{100.00} &
  \textbf{100.00} &
  \textbf{100.00} \\
\textbf{Computers} &
  75.60 &
  72.00 &
  75.60 &
  \textbf{81.50} &
  76.00 &
  64.44 &
  70.51 &
  78.56 &
  76.00 &
  73.20 \\
\textbf{\underline{CricketX}} &
  68.33 &
  64.97 &
  73.59 &
  79.10 &
  82.31 &
  80.21 &
  81.38 &
  \textbf{84.05} &
  82.23 &
  74.51 \\
\textbf{\underline{CricketY}} &
  74.77 &
  70.87 &
  75.38 &
  80.30 &
  84.87 &
  79.38 &
  80.19 &
  83.90 &
  \textbf{85.03} &
  77.21 \\
\textbf{\underline{CricketZ}} &
  72.18 &
  66.62 &
  74.62 &
  81.20 &
  83.08 &
  80.10 &
  83.40 &
  84.92 &
  \textbf{85.77} &
  77.23 \\
\textbf{\underline{DiaSizeRed}} &
  96.63 &
  94.80 &
  93.14 &
  30.10 &
  94.12 &
  96.57 &
  \textbf{97.30} &
  93.46 &
  97.03 &
  92.52 \\
\textbf{DisPhaOutAgeGro} &
  72.81 &
  74.82 &
  74.82 &
  71.70 &
  \textbf{76.26} &
  73.09 &
  74.62 &
  73.38 &
  75.47 &
  73.02 \\
\textbf{DisPhaOutCor} &
  78.84 &
  77.17 &
  72.83 &
  77.10 &
  77.17 &
  \textbf{79.28} &
  78.23 &
  76.81 &
  76.78 &
  78.11 \\
\textbf{DisPhaTW} &
  68.27 &
  66.91 &
  67.63 &
  66.50 &
  68.35 &
  65.97 &
  67.04 &
  66.47 &
  \textbf{71.87} &
  68.06 \\
\textbf{Earthquakes} &
  \textbf{76.91} &
  74.82 &
  74.82 &
  71.20 &
  74.82 &
  75.40 &
  74.82 &
  74.24 &
  74.82 &
  75.25 \\
\textbf{\underline{ECG200}} &
  88.00 &
  85.50 &
  87.00 &
  87.40 &
  85.00 &
  \textbf{90.90} &
  86.18 &
  91.80 &
  90.60 &
  89.70 \\
\textbf{ECG5000} &
  94.21 &
  93.89 &
  94.13 &
  93.40 &
  94.62 &
  93.65 &
  94.54 &
  93.93 &
  \textbf{94.70} &
  94.39 \\
\textbf{\underline{ECGFiveDays}} &
  97.77 &
  93.69 &
  \textbf{100.00} &
  97.50 &
  \textbf{100.00} &
  84.92 &
  \textbf{100.00} &
  \textbf{100.00} &
  \textbf{100.00} &
  99.47 \\
\textbf{ElectricDevices} &
  74.06 &
  69.25 &
  \textbf{79.92} &
  72.90 &
  77.03 &
  70.60 &
  75.53 &
  70.86 &
  73.05 &
  74.03 \\
\textbf{\underline{FaceAll}} &
  78.85 &
  76.83 &
  78.17 &
  83.90 &
  80.30 &
  89.38 &
  84.14 &
  80.08 &
  \textbf{94.75} &
  92.64 \\
\textbf{\underline{FaceFour}} &
  97.73 &
  98.41 &
  \textbf{100.00} &
  95.50 &
  95.45 &
  97.39 &
  \textbf{100.00} &
  95.68 &
  97.50 &
  98.86 \\
\textbf{\underline{FacesUCR}} &
  88.59 &
  90.01 &
  95.71 &
  95.50 &
  96.29 &
  94.59 &
  \textbf{96.63} &
  96.40 &
  96.16 &
  89.51 \\
\textbf{\underline{FiftyWords}} &
  77.05 &
  73.28 &
  70.55 &
  72.70 &
  80.88 &
  83.14 &
  \textbf{84.50} &
  80.66 &
  83.05 &
  76.99 \\
\textbf{\underline{Fish}} &
  90.34 &
  85.26 &
  98.86 &
  97.90 &
  98.86 &
  93.49 &
  \textbf{99.43} &
  97.60 &
  97.89 &
  92.91 \\
\textbf{FordA} &
  96.30 &
  81.52 &
  92.95 &
  92.00 &
  96.44 &
  85.46 &
  94.10 &
  95.73 &
  94.49 &
  \textbf{97.68} \\
\textbf{FordB} &
  79.42 &
  68.77 &
  82.00 &
  \textbf{91.30} &
  82.35 &
  71.49 &
  82.96 &
  84.89 &
  80.63 &
  83.01 \\
\textbf{\underline{GunPoint}} &
  92.00 &
  95.07 &
  \textbf{100.00} &
  99.10 &
  \textbf{100.00} &
  99.73 &
  \textbf{100.00} &
  \textbf{100.00} &
  \textbf{100.00} &
  97.00 \\
\textbf{Ham} &
  73.81 &
  74.29 &
  66.67 &
  75.70 &
  66.67 &
  66.00 &
  71.52 &
  70.48 &
  72.57 &
  \textbf{76.95} \\
\textbf{HandOutlines} &
  92.03 &
  91.89 &
  91.10 &
  91.10 &
  93.24 &
  92.14 &
  93.22 &
  \textbf{94.65} &
  94.16 &
  91.57 \\
\textbf{\underline{Haptics}} &
  50.75 &
  43.57 &
  46.10 &
  51.90 &
  51.95 &
  44.45 &
  51.68 &
  \textbf{54.87} &
  52.50 &
  51.56 \\
\textbf{Herring} &
  62.97 &
  60.94 &
  54.69 &
  61.90 &
  \textbf{68.75} &
  57.97 &
  58.81 &
  66.56 &
  68.59 &
  60.47 \\
\textbf{\underline{InlineSkate}} &
  55.47 &
  32.24 &
  51.64 &
  37.30 &
  50.00 &
  54.18 &
  52.69 &
  48.51 &
  45.82 &
  \textbf{66.75} \\
\textbf{InsWinSou} &
  66.56 &
  63.28 &
  52.32 &
  50.70 &
  65.51 &
  61.87 &
  64.29 &
  63.04 &
  65.66 &
  \textbf{66.78} \\
\textbf{\underline{ItaPowDem}} &
  97.06 &
  97.00 &
  90.86 &
  96.30 &
  96.31 &
  96.71 &
  97.06 &
  96.42 &
  96.91 &
  \textbf{97.31} \\
\textbf{LarKitApp} &
  79.39 &
  57.07 &
  76.53 &
  90.00 &
  86.40 &
  78.19 &
  80.68 &
  \textbf{90.03} &
  90.00 &
  80.64 \\
\textbf{\underline{Lightning2}} &
  \textbf{86.89} &
  72.46 &
  83.61 &
  77.00 &
  81.97 &
  86.56 &
  74.81 &
  78.69 &
  76.39 &
  76.72 \\
\textbf{\underline{Lightning7}} &
  76.99 &
  74.11 &
  68.49 &
  \textbf{84.50} &
  73.97 &
  82.19 &
  76.34 &
  80.27 &
  82.19 &
  76.85 \\
\textbf{\underline{Mallat}} &
  96.88 &
  96.46 &
  93.82 &
  97.20 &
  96.20 &
  95.76 &
  \textbf{97.50} &
  94.06 &
  95.60 &
  96.57 \\
\textbf{Meat} &
  93.17 &
  93.33 &
  90.00 &
  \textbf{96.80} &
  93.33 &
  93.33 &
  88.79 &
  93.33 &
  94.50 &
  95.00 \\
\textbf{\underline{MedicalImages}} &
  78.59 &
  78.00 &
  71.84 &
  77.00 &
  77.76 &
  75.82 &
  79.58 &
  78.66 &
  79.75 &
  \textbf{81.67} \\
\textbf{MidPhaOutAgeGro} &
  56.82 &
  57.79 &
  54.55 &
  56.90 &
  \textbf{59.74} &
  56.23 &
  58.32 &
  52.34 &
  59.55 &
  59.35 \\
\textbf{MidPhaOutCor} &
  82.27 &
  82.82 &
  78.01 &
  80.90 &
  83.16 &
  83.64 &
  \textbf{85.35} &
  81.65 &
  84.12 &
  83.61 \\
\textbf{MiddlePhalanxTW} &
  58.90 &
  56.49 &
  54.55 &
  48.40 &
  57.14 &
  52.92 &
  55.02 &
  50.78 &
  55.58 &
  \textbf{59.67} \\
\textbf{\underline{MoteStrain}} &
  92.36 &
  88.63 &
  87.86 &
  92.80 &
  93.29 &
  90.24 &
  \textbf{94.75} &
  88.61 &
  91.42 &
  94.48 \\
\textbf{\underline{NonInvFetECGTho1}} &
  93.27 &
  89.97 &
  83.82 &
  94.50 &
  93.03 &
  90.66 &
  91.13 &
  \textbf{95.62} &
  95.14 &
  93.64 \\
\textbf{\underline{NonInvFetECGTho2}} &
  94.06 &
  91.13 &
  90.08 &
  94.60 &
  94.45 &
  93.99 &
  94.50 &
  95.79 &
  \textbf{96.88} &
  94.60 \\
\textbf{\underline{OliveOil}} &
  \textbf{93.33} &
  90.67 &
  86.67 &
  83.00 &
  90.00 &
  86.67 &
  88.79 &
  82.00 &
  92.67 &
  90.00 \\
\textbf{\underline{OSULeaf}} &
  79.83 &
  58.39 &
  95.45 &
  97.90 &
  97.93 &
  82.73 &
  \textbf{99.14} &
  92.48 &
  93.80 &
  84.88 \\
\textbf{PhaOutCor} &
  83.17 &
  80.30 &
  77.16 &
  83.90 &
  80.65 &
  82.35 &
  \textbf{84.50} &
  83.75 &
  83.00 &
  84.06 \\
\textbf{Phoneme} &
  32.52 &
  21.20 &
  26.48 &
  33.40 &
  38.24 &
  32.01 &
  36.91 &
  32.81 &
  27.96 &
  \textbf{39.76} \\
\textbf{Plane} &
  \textbf{100.00} &
  \textbf{100.00} &
  \textbf{100.00} &
  \textbf{100.00} &
  \textbf{100.00} &
  \textbf{100.00} &
  \textbf{100.00} &
  \textbf{100.00} &
  \textbf{100.00} &
  \textbf{100.00} \\
\textbf{ProPhaOutAgeGro} &
  84.44 &
  84.88 &
  83.41 &
  85.30 &
  \textbf{85.85} &
  84.63 &
  84.97 &
  84.49 &
  85.51 &
  85.41 \\
\textbf{ProPhaOutCor} &
  90.52 &
  82.82 &
  84.88 &
  \textbf{92.10} &
  87.97 &
  87.32 &
  88.82 &
  91.75 &
  89.90 &
  92.06 \\
\textbf{ProPhaTW} &
  76.49 &
  81.46 &
  80.00 &
  78.00 &
  81.46 &
  77.90 &
  \textbf{81.86} &
  78.15 &
  81.61 &
  79.12 \\
\textbf{RefDev} &
  58.03 &
  58.93 &
  49.87 &
  52.50 &
  55.73 &
  53.23 &
  55.83 &
  52.27 &
  53.47 &
  \textbf{59.04} \\
\textbf{ScreenType} &
  53.33 &
  45.60 &
  46.40 &
  \textbf{62.20} &
  58.93 &
  45.52 &
  50.81 &
  57.97 &
  48.56 &
  54.61 \\
\textbf{ShapeletSim} &
  98.33 &
  47.78 &
  \textbf{100.00} &
  77.90 &
  \textbf{100.00} &
  77.61 &
  \textbf{100.00} &
  91.67 &
  \textbf{100.00} &
  97.89 \\
\textbf{ShapesAll} &
  85.22 &
  79.17 &
  90.83 &
  92.10 &
  90.50 &
  88.58 &
  \textbf{93.00} &
  91.83 &
  90.82 &
  86.11 \\
\textbf{SmaKitApp} &
  83.44 &
  81.07 &
  72.53 &
  78.60 &
  \textbf{85.33} &
  74.43 &
  82.21 &
  75.57 &
  82.13 &
  82.35 \\
\textbf{\underline{SonAIBORobSur1}} &
  90.67 &
  75.64 &
  63.23 &
  \textbf{95.80} &
  76.54 &
  84.58 &
  82.64 &
  86.39 &
  92.41 &
  89.58 \\
\textbf{\underline{SonAIBORobSur2}} &
  83.25 &
  81.86 &
  85.94 &
  \textbf{97.80} &
  92.76 &
  89.63 &
  92.48 &
  94.61 &
  91.64 &
  87.40 \\
\textbf{\underline{StarLightCurves}} &
  97.84 &
  96.40 &
  97.78 &
  97.20 &
  98.15 &
  98.13 &
  \textbf{98.24} &
  97.78 &
  98.11 &
  97.94 \\
\textbf{Strawberry} &
  96.38 &
  96.49 &
  97.57 &
  98.10 &
  97.03 &
  96.84 &
  96.63 &
  \textbf{98.27} &
  98.19 &
  96.84 \\
\textbf{\underline{SwedishLeaf}} &
  94.29 &
  89.57 &
  92.16 &
  95.60 &
  95.36 &
  94.66 &
  96.55 &
  96.35 &
  \textbf{96.59} &
  95.54 \\
\textbf{\underline{Symbols}} &
  88.39 &
  88.56 &
  96.68 &
  90.60 &
  97.39 &
  96.16 &
  97.66 &
  \textbf{98.03} &
  97.46 &
  97.24 \\
\textbf{\underline{SynthCon}} &
  99.03 &
  97.57 &
  96.67 &
  \textbf{100.00} &
  99.67 &
  99.53 &
  99.79 &
  99.60 &
  99.70 &
  99.00 \\
\textbf{ToeSeg1} &
  84.43 &
  74.12 &
  93.86 &
  96.30 &
  \textbf{98.25} &
  92.46 &
  96.53 &
  96.14 &
  97.02 &
  84.74 \\
\textbf{ToeSeg2} &
  88.46 &
  81.54 &
  \textbf{96.15} &
  90.60 &
  95.38 &
  86.23 &
  95.38 &
  94.31 &
  92.62 &
  87.93 \\
\textbf{\underline{Trace}} &
  99.00 &
  97.80 &
  \textbf{100.00} &
  \textbf{100.00} &
  \textbf{100.00} &
  \textbf{100.00} &
  \textbf{100.00} &
  \textbf{100.00} &
  \textbf{100.00} &
  \textbf{100.00} \\
\textbf{\underline{TwoLeadECG}} &
  98.72 &
  90.39 &
  98.07 &
  \textbf{100.00} &
  99.65 &
  98.86 &
  99.46 &
  99.67 &
  99.91 &
  98.44 \\
\textbf{\underline{TwoPatterns}} &
  99.77 &
  94.67 &
  99.30 &
  \textbf{100.00} &
  \textbf{100.00} &
  99.96 &
  \textbf{100.00} &
  \textbf{100.00} &
  \textbf{100.00} &
  99.69 \\
\textbf{UWavGesLibAll} &
  95.48 &
  95.73 &
  93.89 &
  86.00 &
  96.85 &
  97.23 &
  96.89 &
  94.37 &
  \textbf{97.57} &
  95.59 \\
\textbf{\underline{UWavGesLibX}} &
  81.10 &
  78.96 &
  76.21 &
  78.00 &
  83.98 &
  82.86 &
  84.11 &
  81.38 &
  \textbf{85.46} &
  82.88 \\
\textbf{\underline{UWavGesLibY}} &
  74.16 &
  71.15 &
  68.51 &
  67.00 &
  76.55 &
  76.15 &
  77.23 &
  75.46 &
  \textbf{77.29} &
  75.74 \\
\textbf{\underline{UWavGesLibZ}} &
  75.86 &
  73.58 &
  69.49 &
  75.00 &
  78.31 &
  76.40 &
  78.44 &
  75.02 &
  \textbf{79.17} &
  76.81 \\
\textbf{\underline{Wafer}} &
  \textbf{99.98} &
  99.50 &
  99.48 &
  99.90 &
  99.94 &
  99.55 &
  99.91 &
  99.85 &
  99.83 &
  99.97 \\
\textbf{Wine} &
  66.85 &
  62.96 &
  74.07 &
  74.40 &
  77.78 &
  56.85 &
  \textbf{89.06} &
  65.93 &
  80.74 &
  77.78 \\
\textbf{\underline{WordSynonyms}} &
  63.64 &
  62.43 &
  63.79 &
  62.20 &
  73.82 &
  77.87 &
  \textbf{78.74} &
  73.23 &
  75.52 &
  65.39 \\
\textbf{Worms} &
  76.75 &
  61.04 &
  55.84 &
  79.10 &
  55.84 &
  71.82 &
  \textbf{80.17} &
  76.88 &
  72.73 &
  79.22 \\
\textbf{WormsTwoClass} &
  79.09 &
  62.34 &
  \textbf{83.12} &
  74.70 &
  77.92 &
  78.44 &
  81.58 &
  78.18 &
  79.87 &
  80.52 \\
\textbf{\underline{Yoga}} &
  82.80 &
  84.14 &
  \textbf{91.83} &
  87.00 &
  91.77 &
  87.86 &
  83.47 &
  89.06 &
  90.85 &
  85.59 \\
\textbf{} &
  &
  &
  &
  &
  &
  &
  &
  &
  &
  \\
  \\ \hline
\end{longtable}

\normalsize

\clearpage

\section{Comparison of r-STSF with and without the autoregressive representation.}
\label{appendix:autoregressive-importance}
\begin{figure}[h!]
    \centering
    \includegraphics[scale=0.24, angle=90]{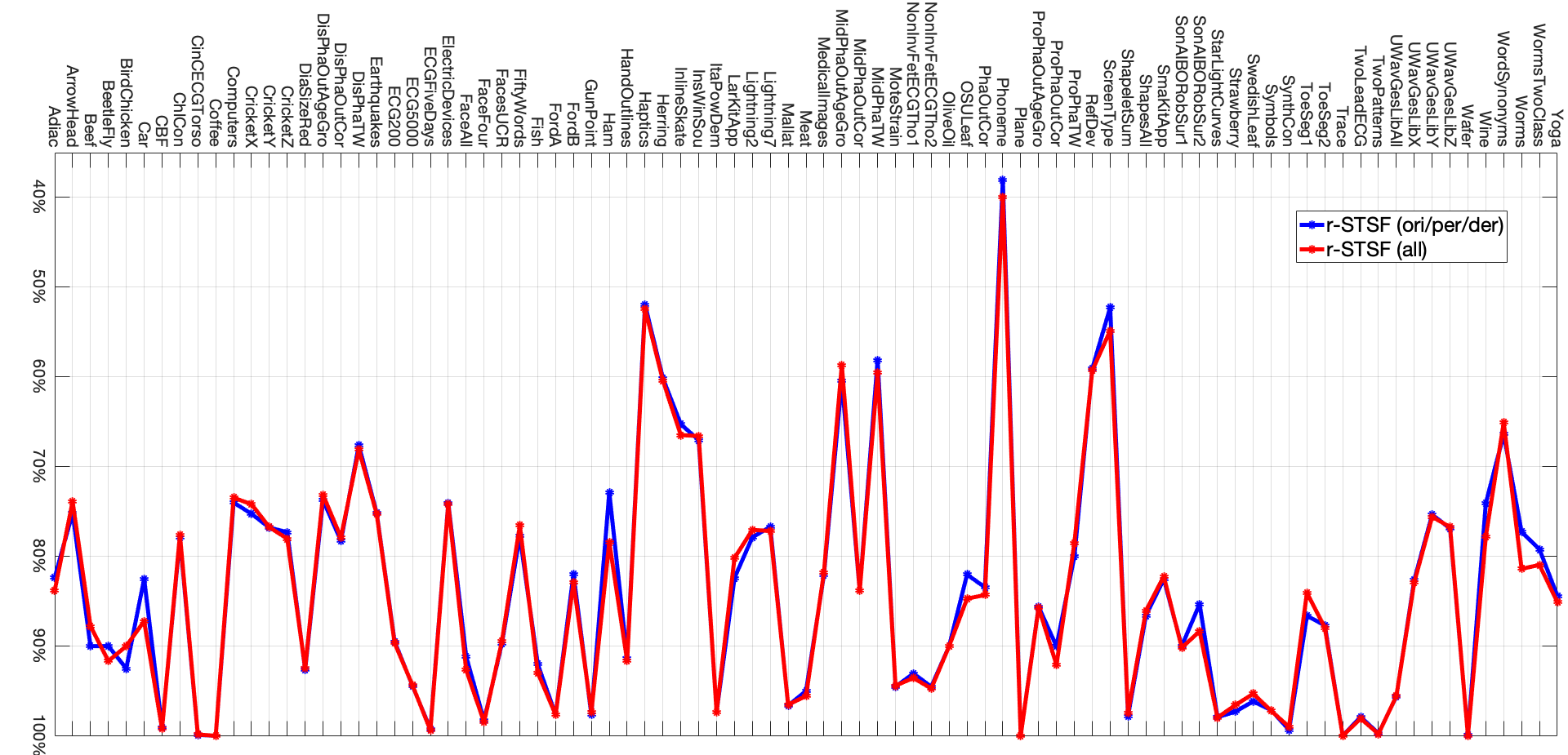}
    \caption{\textcolor{black}{Comparison of average accuracy  (x-axis) of r-STSF when using three time series representations (r-STSF (ori/per/der)) and when using four time series representations (i.e., including autoregressive representation - r-STSF (all))}} 
    \label{fig:tsrepr-comparison2}
\end{figure}

\clearpage

\section{Comparison of r-STSF with and without the \emph{counts of mean-crossings} (cmc) and \emph{counts of values above the mean} (cam) aggregation functions.}
\label{appendix:cmc-cam-importance}
\begin{figure}[h!]
    \centering
    \includegraphics[scale=0.15, angle=90]{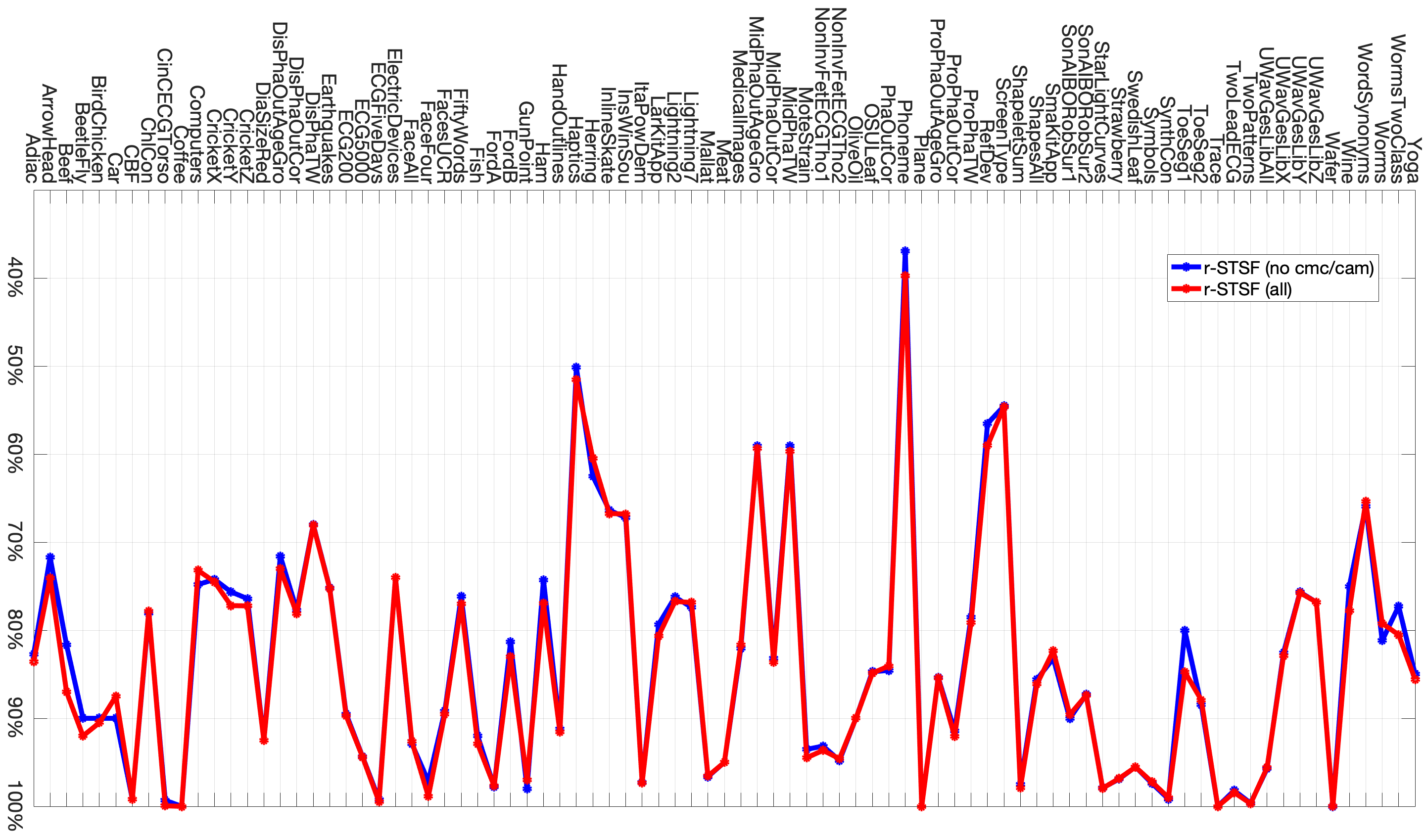}
    \caption{\textcolor{black}{Comparison of average accuracy  (x-axis) of r-STSF when using all nine aggregation functions (r-STSF (all)) and when using seven aggregation functions (i.e., not including cmc and cam aggregation functions - r-STSF (no cmc/cam))}} 
    \label{fig:aggfns-cmc-cam-comparison}
\end{figure}

\end{appendices}

\end{document}